%% file: manuscript.tex
\documentclass[acmsmall,screen]{acmart}

\AtBeginDocument{%
  \providecommand\BibTeX{{%
    \normalfont B\kern-0.5em{\scshape i\kern-0.25em b}\kern-0.8em\TeX}}}

\setcopyright{acmcopyright}
\acmJournal{TIST}
\acmYear{2022} \acmVolume{1} \acmNumber{1} \acmArticle{1} \acmMonth{1} \acmPrice{15.00}\acmDOI{10.1145/3510032}

\usepackage{color}
\usepackage{bbding}
\usepackage{microtype}
\usepackage{graphicx}
\usepackage{grffile}
\usepackage{subfigure}
\usepackage{booktabs}
\usepackage{algorithm}
\usepackage{algorithmic,eqparbox,array}
\usepackage{diagbox}
\usepackage{multirow}
\usepackage{mathrsfs}
\usepackage{amsmath}
\usepackage{makecell}

\usepackage[nolist,nohyperlinks]{acronym}
\acrodef{AI}{Artificial Intelligence}
\acrodef{GRNN}{Generative Regression Neural Network}
\acrodef{CNN}{Convolutional Neural Network}
\acrodef{FCNN}{Fully-Connected Neural Network}
\acrodef{FC}{Fully-Connected}
\acrodef{DL}{Deep Learning}
\acrodef{ANN}{Artificial Neural Network}
\acrodef{HMM}{Hidden Markov Model}
\acrodef{SVM}{Support Vector Machine}
\acrodef{FL}{Federated Learning}
\acrodef{MPC}{Secure Multi-Party Computation}
\acrodef{GAN}{Generative Adversarial Network}
\acrodef{DLG}{Deep Leakage from Gradients}
\acrodef{iDLG}{Improved \ac{DLG}}
\acrodef{IG}{Inverting Gradient}
\acrodef{LFW}{Labeled Faces in the Wild}
\acrodef{FedSGD}{Federated Stochastic Gradient Descent}
\acrodef{FedAvg}{Federated Averaging}
\acrodef{ML}{Machine Learning}
\acrodef{GLU}{Gated Linear Unit}
\acrodef{MSE}{Mean Square Error}
\acrodef{WD}{Wasserstein Distance}
\acrodef{CD}{Cosine Distance}
\acrodef{TVLoss}{Total Variation Loss}
\acrodef{EM}{Earth Mover’s}
\acrodef{PSNR}{Peak Signal-to-Noise Ratio}
\acrodef{SSIM}{Structural Similarity}
\acrodef{HE}{Homomorphic Encryption}
\acrodef{DP}{Differential Privacy}
\acrodef{TEE}{Trusted Execution Environment}
\acrodef{Non-IID}{Non-Independent and Non-Identically Distributed}
\acrodef{IID}{Independent and Identically Distributed}
\acrodef{SGD}{Stochastic Gradient Descent}
\acrodef{GDPR}{\emph{General Data Protection Regulation}}
\acrodef{EU}{\emph{European Union}}
\acrodef{EEA}{\emph{European Economic Area}}

\begin{document}

\title{GRNN: Generative Regression Neural Network - A Data Leakage Attack for Federated Learning}

\author{Hanchi Ren}
\email{845231@swansea.ac.uk}
\orcid{0000-0003-1310-8777}

\author{Jingjing Deng}
\email{j.deng@swansea.ac.uk}
\orcid{0000-0001-9274-651X}

\author{Xianghua Xie}
\email{x.xie@swansea.ac.uk}
\orcid{0000-0002-2701-8660}

\affiliation{%
  \institution{Computer Vision and Machine Learning Group (\url{csvision.swansea.ac.uk}), Department of Computer Science, Swansea University}
  \city{Swansea}
  \country{United Kingdom}
}
\renewcommand{\shortauthors}{H. Ren, J. Deng and X. Xie}

\begin{abstract}
  Data privacy has become an increasingly important issue in Machine Learning (ML), where many approaches have been developed to tackle this challenge, \emph{e.g.} cryptography (Homomorphic Encryption (HE), Differential Privacy (DP), \emph{etc.}) and collaborative training (Secure Multi-Party Computation (MPC), Distributed Learning and Federated Learning (FL)). These techniques have a particular focus on data encryption or secure local computation. They transfer the intermediate information to the third party to compute the final result. Gradient exchanging is commonly considered to be a secure way of training a robust model collaboratively in Deep Learning (DL). However, recent researches have demonstrated that sensitive information can be recovered from the shared gradient. Generative Adversarial Network (GAN), in particular, has shown to be effective in recovering such information. However, GAN based techniques require additional information, such as class labels which are generally unavailable for privacy-preserved learning. In this paper, we show that, in the FL system, image-based privacy data can be easily recovered in full from the shared gradient only via our proposed Generative Regression Neural Network (GRNN). We formulate the attack to be a regression problem and optimize two branches of the generative model by minimizing the distance between gradients. We evaluate our method on several image classification tasks. The results illustrate that our proposed GRNN outperforms state-of-the-art methods with better stability, stronger robustness, and higher accuracy. It also has no convergence requirement to the global FL model. Moreover, we demonstrate information leakage using face re-identification. Some defense strategies are also discussed in this work.
\end{abstract}

\ccsdesc[500]{Computing methodologies~Machine learning}
\ccsdesc[500]{Computing methodologies~Computer vision}
\ccsdesc[500]{Security and privacy~Database and storage security}

\keywords{Federated Learning, Data Privacy, Gradient Leakage Attack, Image Generation}

\maketitle

\section{Introduction}
More often than not, the success of \ac{DL}~\cite{he2016deep,krizhevsky2017imagenet} relies on the availability of a large quantity of data. A centralized learning scheme with cloud-based distributed computing system is commonly used to speed up the training and scale up to larger datasets~\cite{li2014scaling,xing2015petuum,moritz2015sparknet,iandola2016firecaffe,lin2017deep}. However, due to data protection and privacy requirements such systems are generally infeasible as they require centralized data for training. For instance, data owners (\emph{e.g.} hospital, finance company, or government agent) are not willing or not able to share private data with algorithm and computing platform providers, which causes the so-called ``Data Islands'' issue. Therefore, decentralized training approaches with data privacy protection are more attractive.

\ac{FL}~\cite{mcmahan2017communication} was proposed to jointly train a model without directly accessing the private training data. Instead of sharing data, the aggregation server shares a global model and requires the individual data owners to expose the gradient information computed privately only. The gradient is generally considered to be secure to share. However, recent studies found that the gradient-sharing scheme is in fact not privacy-preserved~\cite{fredrikson2015model,shokri2017membership,hitaj2017deep,melis2019exploiting}. For example, the presence of a specific property of the training dataset can be identified given the gradient information~\cite{melis2019exploiting}. Hitaj \emph{et al.}~\cite{hitaj2017deep} showed the feasibility of generating images that are similar to training images using \ac{GAN} given any known class label~\cite{goodfellow2014generative}. \ac{DLG} based models~\cite{zhu2019deep,zhao2020idlg} were proposed to approximate the leaked gradient via learning the input while fixing the model weights. However, they are usually unstable and sensitive to the size of the training batch and the resolution of input image. Furthermore, Geiping \emph{et al.}~\cite{geiping2020inverting} discussed the theoretical aspect of inverting gradient to its corresponding training data and used magnitude-invariant cosine similarity loss function in proposed \ac{IG}, which is capable of recovering high-resolution images (\emph{i.e.} 224*224) with a large number of training batch (\emph{i.e.} $\#Batch=100$). However, we find that the success rate is relatively low.

\ac{GAN} is a generative model proposed by Goodfellow \emph{et al.}~\cite{goodfellow2014generative} for image generation. Inspired by the \ac{GAN} and \ac{DLG} model, we introduce a gradient guided image generation strategy that properly addresses the stability and data quality issues of \ac{DLG} based methods. The proposed \ac{GRNN}, a novel data leakage attack method is capable of recover private training image up to a resolution of 256*256 and a batch size of 256. The method is particularly suitable for \ac{FL} as the local gradient and global model are readily available in the system setting. \ac{GRNN} consists of two branches for generating fake training data and corresponding label. It is trained in an end-to-end fashion by approximating the fake gradient that is calculated by the generated data and label to the true gradient given the global model. \ac{MSE}, \ac{WD} and \ac{TVLoss} are used jointly to evaluated divergence between true and fake gradients. We empirically evaluate the performance of our method on several image classification tasks, and comprehensively compared against the state-of-the-art. The experimental results confirm that the proposed method is much more stable and capable of producing image with better quality when a large batch size and resolution are used. The contributions of this work are five-folds:

\begin{itemize}
    \item We propose a novel method of data leakage attack for \ac{FL}, which is capable of recovering private training image up to a resolution of 256*256, a batch size of 256 as well as the corresponding labels from the shared gradient.
    \item We conduct comprehensive evaluation, where both qualitative and quantitative results are presented to prove the effectiveness of \ac{GRNN}. We also compare the proposed method against the state-of-the-art, which shows that \ac{GRNN} is superior in terms of success rate of attack, the fidelity of recovered data and the accuracy of label inference. In addition, our method is much more stable than others with respect to the size of the training batch and input resolution.
    \item We conduct a face re-identification experiment that shows using the image generated by the proposed \ac{GRNN} can achieve higher \emph{Top-1}, \emph{Top-3} and \emph{Top-5} accuracies compared to the state-of-the-art. 
    \item We discuss the potential defense strategies and quantitatively evaluate the effectiveness of our method against noise addition defense strategy. 
    \item The implementation of the method is publicly available to ensure its reproducibility.
\end{itemize}

The rest of the paper is organized as follows: An overview of related works on data leakage attack are presented in Section~\ref{sec:rw}. The proposed method is described in Section~\ref{sec:pm}. The details of experimental results, discussions, and potential defense strategies are provided in Section~\ref{sec:er}. Section~\ref{sec:cc} concludes the paper.

\section{Related Work}
\label{sec:rw}
Given a trained victim model and a target label, Hitaj \emph{et al.}~\cite{hitaj2017deep} proposed a \ac{GAN} based data recovery method that can generate a set of new data having close distribution to the training dataset. First, each participant trains a local model on its own dataset for several iterations to achieve an accuracy above the pre-set threshold which is used as a discriminator in the next stage. To train the generator, the weights of the discriminator are firstly fixed. Then, given a specified class, the generator is learned via producing an image that maximizes the classification confidence of the discriminator. The generated image has no explicit correspondence to the training data, and the method is sensitive to the variance of training data~\cite{melis2019exploiting}.

\ac{DLG} proposed by Zhu \emph{et al.}~\cite{zhu2019deep} is one of the first works investigating the training data leakage in collaborative learning. It formulates the image recovery as a regression problem as follows: A batch of randomly initialized ``dummy'' images and labels are used to compute ``dummy'' gradient via a forward-backward pass on the global model. The true local gradient is readily available in the training system. \ac{DLG} updates the ``dummy'' images and labels via minimizing the \ac{MSE} distance between the ``dummy'' gradient and the true gradient. It can achieve exact pixel-wise data recovering without additional information other than the shared global model and local gradient. Zhao \emph{et al.}~\cite{zhao2020idlg} introduced \ac{iDLG} that addresses the divergence and inconsistent label inference issues of \ac{DLG}. They found that the derivative value corresponding to the ground-truth label drops in the range of [-1, 0], and other cases lie in the range of [0, 1]. It is then feasible to identify the correct label in a such na\"ive way. In addition to the low accuracy of label inference, \ac{DLG} based methods often fail to recover the image from the gradient when the variance of the data is large, which is very common for the dataset that has a large number of classes. \ac{IG}~\cite{geiping2020inverting} improved the stability of \ac{DLG} and \ac{iDLG} by introducing magnitude-invariant cosine similarity measurement for loss function, namely \ac{CD}. It aims to find images that pursue similar prediction changes from the classification model rather than those that can generate the close values with the shared gradient. It shows recognizable results of recovering high-resolution images (\emph{i.e.} 224*224) with a large number of training batch (\emph{i.e.} $\#Batch=100$). Compared to the closely related works~\cite{zhu2019deep,zhao2020idlg,geiping2020inverting}, our proposed method takes a generative approach having higher stability for recovering high-resolution images (\emph{i.e.} up to 256*256) with a large batch size (\emph{i.e.} $\#Batch=256$). Table \ref{tab:compareMethod} presents the key differences between \ac{DLG}, \ac{iDLG} \ac{IG} and proposed \ac{GRNN}.

\begin{table}
    \centering
    \caption{Comparison of \ac{GRNN} with the closely related works.}
    \label{tab:compareMethod}
    \begin{tabular}{c|c|c|c|c}
         \hline
         \textbf{Method} & \textbf{Recovery Mode} & \textbf{\#Batch} & \textbf{Resolution} & \textbf{Loss Function} \\
         \hline
         \ac{DLG}\cite{zhu2019deep} & Discriminative & Small, 8 & Low 64*64 & \ac{MSE}  \\
         \ac{iDLG}\cite{zhao2020idlg} & Discriminative & Small, 1 only & Low 64*64 & \ac{MSE}  \\
         \ac{IG}\cite{geiping2020inverting} & Discriminative & Medium, 100 & High 224*224 & \ac{CD} \& \ac{TVLoss}  \\
         \hline
         \textbf{\ac{GRNN}} & Generative & Large, 256 & High 256*256 & \ac{MSE} \& \ac{WD} \& \ac{TVLoss} \\
         \hline
    \end{tabular}
\end{table}

Cryptography based privacy-preserving methods, such as secure aggregation protocol~\cite{bonawitz2016practical}, \ac{HE}~\cite{armknecht2015guide,kwabena2019mscryptonet}, \ac{DP}~\cite{dwork2006differential,dwork2014algorithmic,bu2020deep} have been developed for privacy-preserving learning, for instance,  linear regression model~\cite{gascon2016secure,mohassel2017secureml}, decision trees~\cite{aslett2015encrypted,bost2015machine}, deep neural networks~\cite{hesamifard2017cryptodl,hesamifard2019deep,kwabena2019mscryptonet}. However, the cryptography operations are computationally expensive, and the consistency of visual patterns of images is generally not guaranteed in encrypted data format, which usually leads to a learning model with poor generalization ability. Gradient encryption is an alternative defense strategy~\cite{hardy2017private,aono2017privacy,hao2019towards}, where \ac{HE} is used to encrypt the gradient shared between server and clients, therefore, the aggregation can be done directly on the encrypted gradient. In Section~\ref{sec:ds}, we discuss the potential defense strategies and quantitatively evaluate the effectiveness of proposed \ac{GRNN} against noise addition.

\section{Proposed Method}
\label{sec:pm}
There are three major challenges that have not been well addressed by existing data recovery methods as follows: model stability, the feasibility of recovering data from large batch size, and fidelity with high-resolution. Both \cite{zhu2019deep,zhao2020idlg,geiping2020inverting} treat the data recovery as a high-dimensional fitting problem driven by the gradient regression objective. The state-of-the-art models are fairly sensitive to the data initialization, easily fail to capture the individual image characteristic when a large number of gradients are aggregated (\emph{i.e.} \ac{FedAvg}), and hardly maintain the natural image structures in detail when resolution is increased. In this paper, we propose \ac{GRNN} model which formulates the data recovery task as a data generation problem that is guided by the gradient information. We introduce a \ac{GAN} model as the image data generator and a simply \ac{FC} layer as a label data generator, where the so-called fake gradient can then be computed given the shared global model. By jointly optimizing both two generators to approximate the true gradient, \ac{GRNN} can well align the latent space of \ac{GAN} with the gradient space of the shared global model, furthermore, generate the training data with high fidelity stably. Therefore, \ac{GRNN} is able to recover the data from the local gradient shared between client and server in a \ac{FL} setting, which can also be used for malicious purposes, such as stealing private data from a client. In this section, we will first introduce a neural network model based \ac{FL} system in Section \ref{sec:fl}, then present the proposed \ac{GRNN} method in Section \ref{sec:grnn}.

\begin{figure}
\centering
    \begin{center}
        \includegraphics[width=0.95\linewidth]{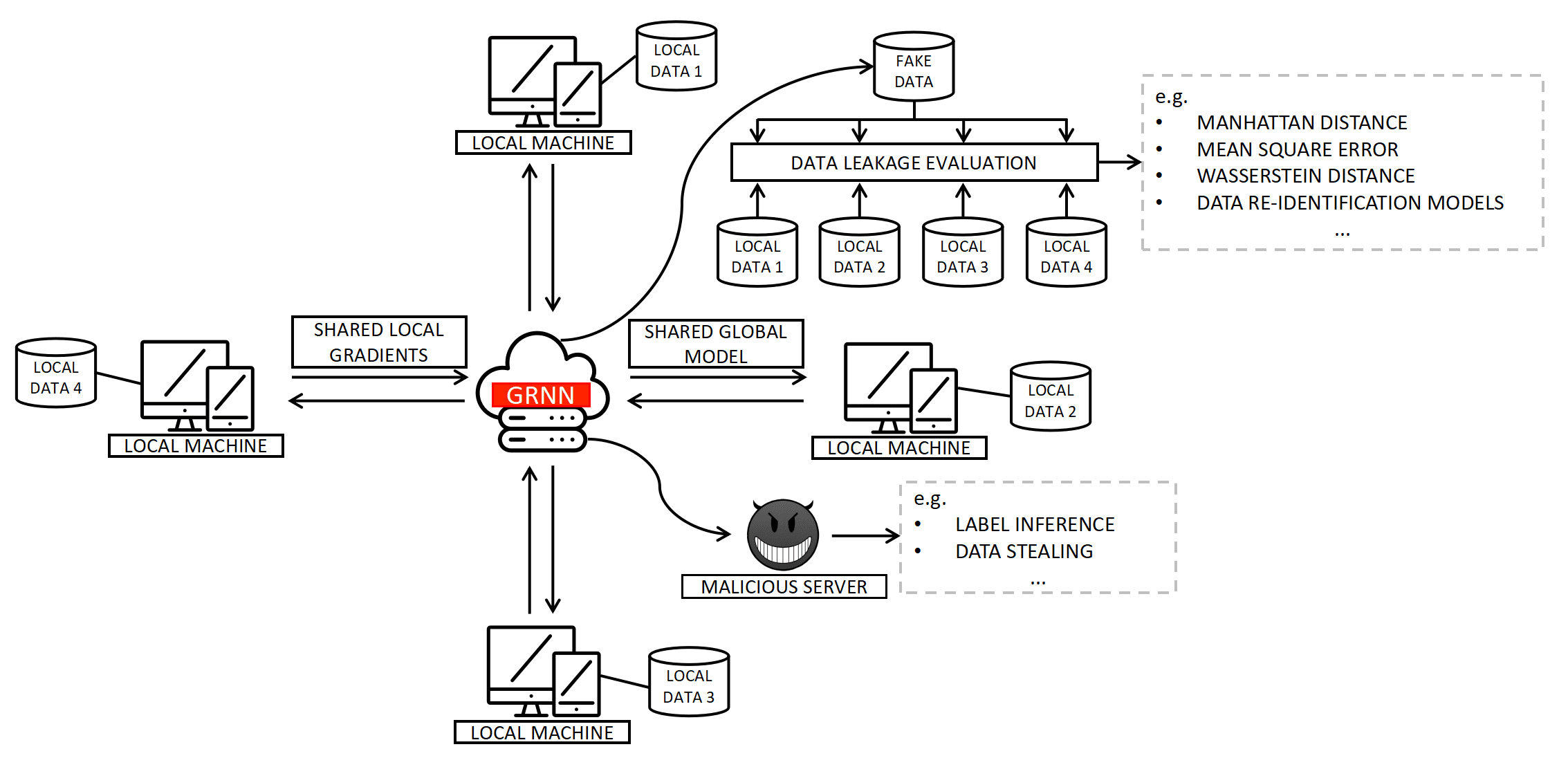}
    \end{center}
\caption{Illustration of how \ac{GRNN} as a malicious server is deployed in a \ac{FL} system.}
\label{fig:GRNN.in.FL}
\end{figure}

\subsection{Federated Learning}
\label{sec:fl}
\ac{FL} is a distributed collaborative training scheme that consists of multiple clients and one parameter server. The gradients calculated restrictively on client nodes are aggregated on the server node using fusion functions~\cite{shokri2015privacy,mcmahan2017communication}. We use the classic \ac{FedAvg} method to train deep neural network over multiple parties which runs a number of steps of \ac{SGD} in parallel on client node and then averages the resulting model updates via a central server periodically. The global model is merged by taking the average of gradients from local models according to $\omega\prime = \sum_{i}^N \frac {1}{N} \omega_i \label{FedAvg}$, where $\omega\prime$ and $\omega_{i}$ are the gradients of global model and the $i_{th}$ local model, and $N$ is the total number of the clients. 

We learn a \ac{CNN} based image classification model, whereas \ac{FedAvg} is applicable to any finite-sum objective. Formally, at iteration $t$, the $i_{th}$ ($i \in \{1,2,...,C\}$) client computes the \ac{CNN} model $\theta_t$ and the local gradient $g^i_t$ based on its local training data $(x^i_t, y^i_t)$ (see Equ.~\ref{equ:LocalGradients}). $\mathcal{F}(\bullet)$, $\theta_t$ and $\mathcal{L}(\bullet)$ are the global learning model, network parameters at iteration $t$ and loss function respectively. The local gradient $g^i_t$ is calculated using typical \ac{SGD} at client node independently. The server aggregates the local gradient $g^i_t$ and then updates the global model weights $\theta_{t+1}$, as shown in Equ.~\ref{equ:fed} where $C$ is the number of clients.

\begin{equation}
g^i_t = \frac{\partial \mathcal{L}(\mathcal{F}(<x^i_t, y^i_t>, \theta_t))}{\partial \theta_t} \label{equ:LocalGradients}
\end{equation}
\begin{equation}
\theta_{t+1}=\theta_t - \frac {1}{C} \sum_{i=1}^C g_i
\label{equ:fed}
\end{equation}


\subsection{GRNN: Data Leakage Attack}
\label{sec:grnn}
\begin{figure}
\centering
    \begin{center}
        \includegraphics[width=0.98\linewidth]{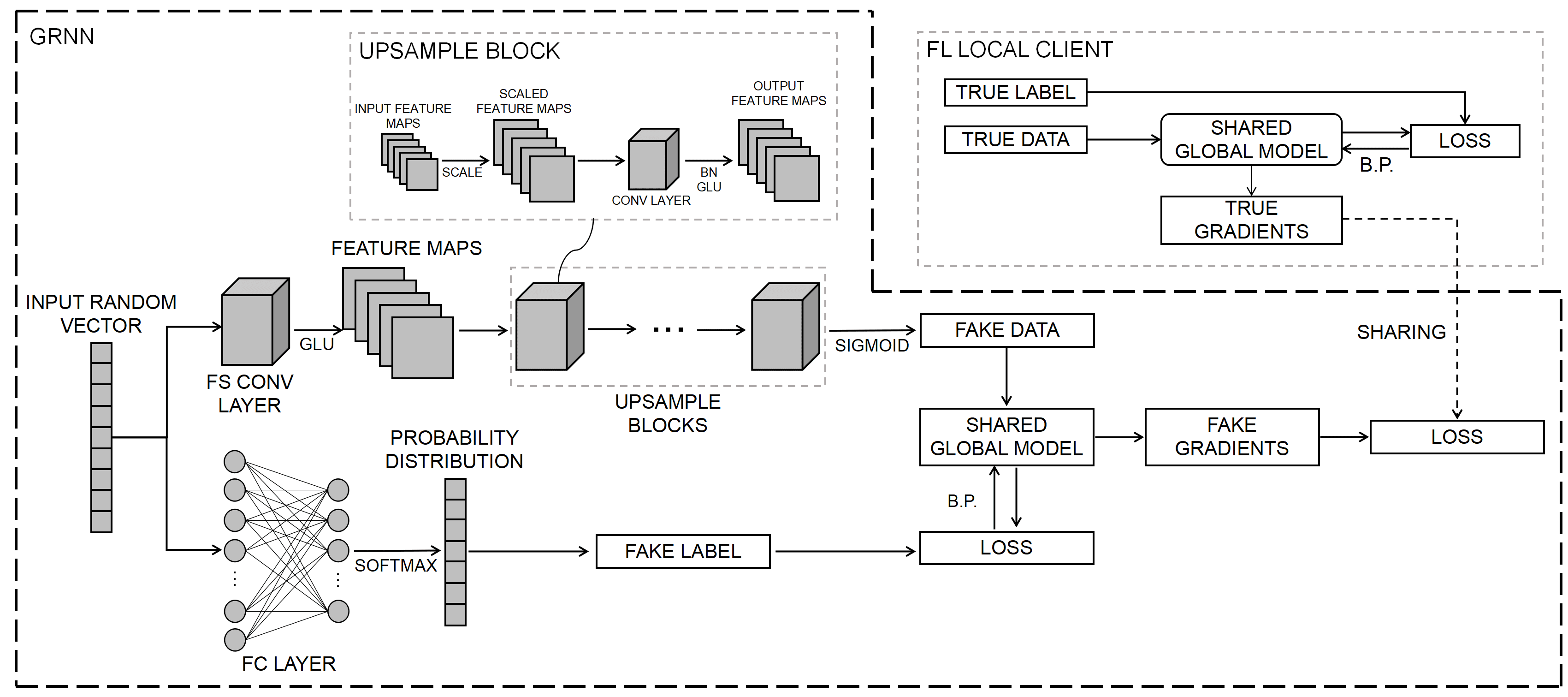}
    \end{center}
\caption{Details of the proposed \ac{GRNN} where the top branch is for generating the fake image and the bottom branch is for inferring the label. ``FC LAYER'' is fully-connected layer. ``FS CONV LAYER'' is fractionally-strided convolutional layer. For details of upsampling block, please refer to Table~\ref{tab:UpsampleBlock}.}
\label{fig:GRNN.Details}
\end{figure}

The proposed \ac{GRNN} is deployed at the central server as illustrated in Fig.~\ref{fig:GRNN.in.FL}. The \ac{GRNN} has access to the neural network architecture and parameters of the global model. The only information that can be obtained between server and client is the gradient calculated based on the current global model. To compute the gradient of the model, the private data, corresponding label, and the global model at current iteration are required at the local client. We hypothesize that the shared gradient contains the information of private data distribution and space partitioning given the global model. The architecture of \ac{GRNN} shown in Fig.~\ref{fig:GRNN.Details} has two branches with the same input that is sampled from a common latent space. We consider that each point within this latent space represents a pair of image data and its corresponding label. The objective of the \ac{GRNN} is to separate this tied representation in the latent space and recover the local image data and corresponding label by approximating the shared gradient that is accessible at the server node.

The top branch is used to recover image data, namely fake-data generator, which consists of a generative network model. We followed the design principle of iWGAN \cite{gulrajani2017improved} where the image is generated from a coarse scale to a fine-scale gradually. The concept is consistent with a reasonable intuition, where the image is drawn from sketch then the details are gradually added. The input random vector is first fed into a fractionally-strided convolutional layer~\cite{zeiler2010deconvolutional,zeiler2011adaptive} to produce a set of feature maps with a resolution of 4*4, which then goes through several upsampling blocks that gradually increase the spatial resolution. The number of the upsampling blocks is determined by the resolution of the target image. For example, if the target image resolution is 32*32, then 3 upsampling blocks ($4\rightarrow8\rightarrow16\rightarrow32$) are used. We also investigate the large resolution of the input image in Section~\ref{sec:dir}. In upsampling block, nearest-neighbor interpolation is used to recover the spatial resolution of feature maps from the previous layer. It then passes through a standard convolutional sub-block for rectifying the detail feature representation, which contains a convolutional layer, a batch normalization layer, and a \ac{GLU}~\cite{dauphin2017language} activation layer. 
Empirically, we found \ac{GLU} is far more stable than \emph{ReLU} and can learn faster than \emph{Sigmoid}. Therefore, we adopted \ac{GLU} as the activation function for our model while the learning strategy using \ac{GAN} driven by a regression objective is the key novelty. The proposed method also works with other activation functions, such as \emph{ReLU}. The details of the upsampling block are listed in Table~\ref{tab:UpsampleBlock}. At the end of the top branch, a data sample that has the same dimension as the training input of the \ac{FL} system is generated.

The bottom branch is used to recover label data, namely fake-label generator which contains a \ac{FC} layer followed by a softmax layer for classification. It takes a randomly sampled vector from latent space as input and outputs its corresponding fake label. We assume the elements in the input vector are independent of each other and subject to a standard Gaussian distribution. The label set in \ac{GRNN} is identical to the one used in the \ac{FL} system. Formally, the fake image $(\hat{x}^j_t$ and fake label $\hat{y}^j_t)$ data generation can be formulated as in Equ.~\ref{equ:Generative}, where $\hat{\theta}$ and $v_t$ are trainable parameters of \ac{GRNN} and input random vector is sampled from a unit Gaussian distribution.
\begin{equation}
(\hat{x}^j_t, \hat{y}^j_t) = \mathcal{G}(v_t | \hat{\theta}_t) \label{equ:Generative}
\end{equation}

\begin{table}
\setlength{\tabcolsep}{18pt}
\begin{center}
\caption{Construction of upsampling block.}
\label{tab:UpsampleBlock}
\begin{tabular}{c|c}
\hline
\textbf{Layer Name} & \textbf{Setting} \\
\hline
Upsampling Layer & \makecell[l]{scale factor: 2\\ mode: nearest} \\
\hline
Convolutional Layer & \makecell[l]{kernel size: 3\\ stride: 1\\ padding: 1} \\
\hline
Batch Normalization Layer & - \\
\hline
Gated Liner Unit & - \\
\hline
\end{tabular}
\end{center}
\end{table}

Given a pair of fake image and label that are generated as described above, a fake gradient on the current global model can be obtained via feeding them as training input and performing one iteration of \ac{SGD} according to Equ.~\ref{equ:LocalGradients}. The objective of \ac{GRNN} is to approximate the true gradient, therefore, the whole model can be trained by minimizing the distance between the fake gradient $\hat{g}^j_t$ and shared true gradient $g^j_t$, \emph{i.e.} the most commonly used loss \ac{MSE} is adopted here:

\begin{equation}
\arg\min \limits_{\hat{\theta}} ||g^j_t-\hat{g}^j_t||^2 \nonumber 
\Longrightarrow \arg\min \limits_{\hat{\theta}} ||\frac{\partial \mathcal{L}(\mathcal{F}(<x^j_t, y^j_t>, \theta_t))}{\partial \theta_t} - \frac{\partial \mathcal{L}(\mathcal{F}(<\hat{x}^j_t, \hat{y}^j_t>, \theta_t))}{\partial \theta_t}||^2 \nonumber
\label{equ:MSE}
\end{equation}

Note that the gradient of the model is a vector, the length of which is equal to the number of trainable parameters. In addition to measuring discrepancy between the true and fake gradients based on Euclidean distance, we also introduce the \ac{WD}~\cite{arjovsky2017wasserstein} loss to minimize the geometric difference between two gradient vectors and \ac{TVLoss}~\cite{rudin1992nonlinear} to impose the smoothness constrain on generated fake image data. Therefore, the loss function for \ac{GRNN}, namely $\hat{\mathcal{L}}(\bullet)$ is formulated as: 
\begin{equation}
    \hat{\mathcal{L}}(g, \hat{g}, \hat{x}) = MSE(g, \hat{g}) + WD(g, \hat{g}) + \alpha \cdot TVLoss(\hat{x})
    \label{equ:LossFunction}
\end{equation}
where we weight the \ac{MSE} loss and \ac{WD} equally and $\alpha$ is the weighting parameter for smoothness regularization. Both branches of the proposed \ac{GRNN} are parameterized using a neural network that is completely differentiable and can be jointly trained in an end-to-end fashion. The complete training procedure for \ac{GRNN} is described in Algorithm~\ref{alg:InferenceAttack}. 

\begin{algorithm}
    \caption{GRNN: Data Leakage Attack}
    \begin{algorithmic}[1]
    \STATE $g^j_t \leftarrow \partial \mathcal{L}(\mathcal{F}(<x^j_t, y^j_t>, \theta_t)) / \partial \theta_t$; \COMMENT{Produce true gradient on local client.}
    \STATE $v_t \leftarrow$ Sampling from $\mathcal{N}(0, 1)$; \COMMENT{initialize random vector inputs for \ac{GRNN}.}
    \FOR{each iteration $i \in [1,2,...,I]$}
        \STATE $(\hat{x}^j_{t,i}, \hat{y}^j_{t,i}) \leftarrow \mathcal{G}(v_t | \hat{\theta}_i)$; \COMMENT{Generate fake images and labels.}
        \STATE $\hat{g}^j_{t,i} \leftarrow {\partial \mathcal{L}(\mathcal{F}(<\hat{x}^j_{t,i}, \hat{y}^j_{t,i}>, \theta_t))}/{\partial \theta_t}$; \COMMENT{Calculate fake gradient on shared global model.}
        \STATE $\mathcal{D}_i \leftarrow \hat{\mathcal{L}}(g^j_t, \hat{g}^j_{t,i}, \hat{x}^j_{t,i})$; \COMMENT{Calculate \ac{GRNN} loss between true gradient and fake gradient.}
        \STATE $\hat{\theta}_{i+1} \leftarrow {\hat{\theta}_i - \eta (\partial \mathcal{D}_i} / {\partial \hat{\theta}_i)}$; \COMMENT{Update \ac{GRNN} model.}
    \ENDFOR
    \STATE \textbf{return} $(\hat{x}^j_{t, I}, \hat{y}^j_{t,I})$; \COMMENT{Return generated fake images and labels.}
    \end{algorithmic}
    \label{alg:InferenceAttack}
\end{algorithm}

\section{Experiment and Discussion}
\label{sec:er}

\subsection{Dataset and Experimental Setting}
A number of experiments on typical computer vision tasks including digit recognition, image classification, and face recognition were conducted to evaluate our proposed method. We used four public benchmarks, MNIST~\cite{lecun1998mnist}, CIFAR-100~\cite{krizhevsky2009learning}, \ac{LFW}~\cite{huang2008labeled} and VGG-Face~\cite{parkhi2015deep} for those tasks. There are 7000 gray-scale handwritten digit images of 28*28 resolution in the MNIST dataset. CIFAR-100 consists of 60000 color images of size 32*32 with 100 categories. \ac{LFW} is a human face dataset that has 13233 images from 5749 different people. We treated each individual as one class, hence, there are 5749 classes in total. VGG-Face consists of over 2.6 million human face images and 2622 identities. 

\emph{LeNet}~\cite{lecun1998gradient} and \emph{ResNet-18}~\cite{he2016deep} are used as the backbone networks for training image classifiers in \ac{FL} system. All neural network models are implemented using PyTorch~\cite{paszke2019pytorch} framework and the source code has been made publicly available\footnote{\url{https://github.com/Rand2AI/GRNN}} for reproducing the results. We replace all \emph{ReLU} function with \emph{Sigmoid} function in order to ensure the model is second-order differentiable, which is as the same as \ac{DLG} and \ac{iDLG} in order to intractably compute the \ac{MSE} loss. Our method is also applicable to the network with \emph{ReLU} function where an second-order differentiable approximation to the \emph{ReLU} function can be used while the computation is intractable compared to using \emph{Sigmoid} function. The batch size varies across different experiments, and a comprehensive comparison study was carried out. \emph{RMSprop} optimizer with a learning rate of 0.0001 and a momentum of 0.99 is used for \ac{GRNN}. Regarding the loss function of \ac{GRNN}, we set the weights of \ac{TVLoss} to $1e-3$ and $1e-6$ for \emph{LeNet} and \emph{ResNet-18}, respectively. 

\subsection{Image Recovery}
\label{sec:dir}

\begin{table}
    \centering
    \caption{Examples of data leakage attack using the proposed \ac{GRNN} on the global model trained over one iteration.}
    \label{tab:TrainingExamples}
    \begin{tabular}{c|c|c}
         \hline
         \textbf{Dataset} & \textbf{Generated Data} & \textbf{True Data}  \\
         \hline
         MNIST & \makecell*[c]{\includegraphics[width=0.485\linewidth]{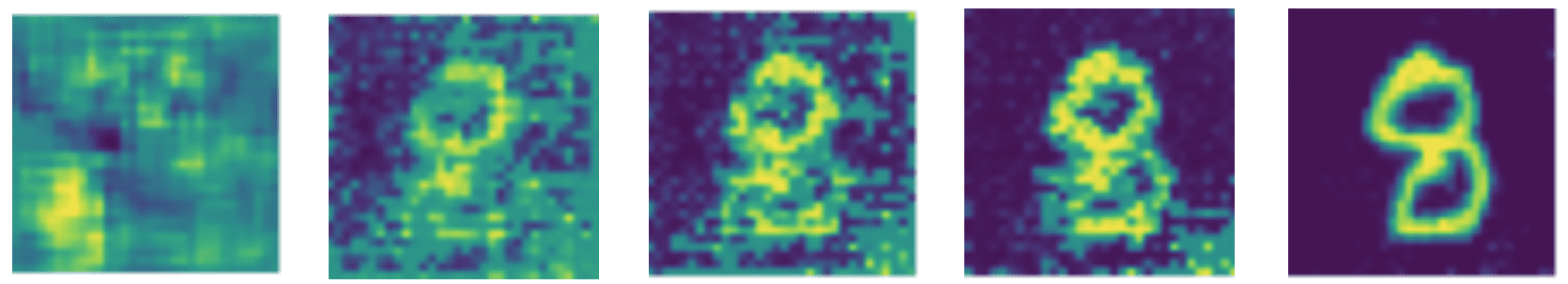}} & \makecell*[c]{\includegraphics[width=0.093\linewidth]{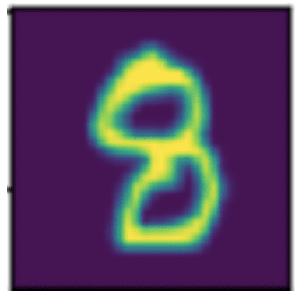}}\\
         \hline
         CIFAR-100 & \makecell*[c]{\includegraphics[width=0.485\linewidth]{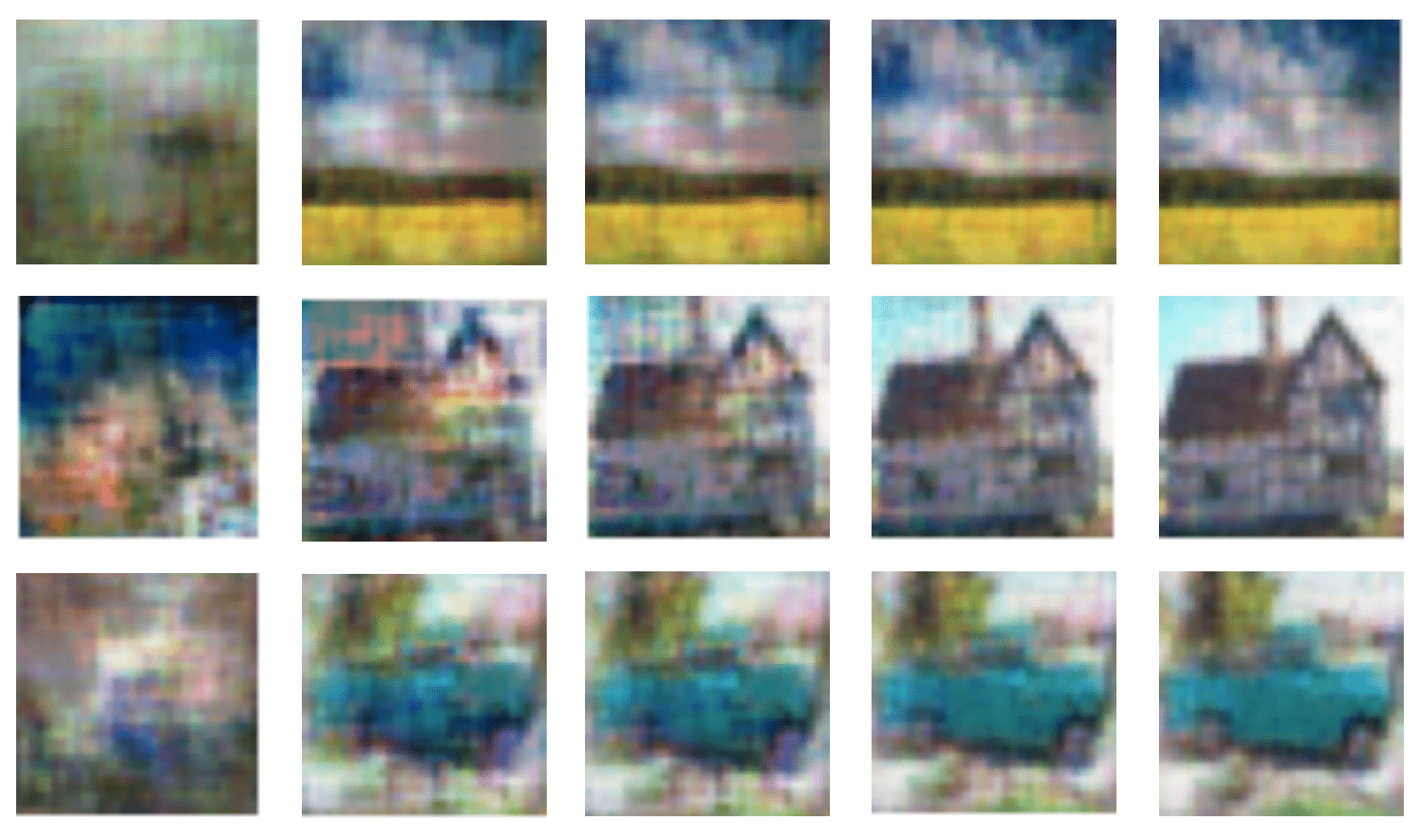}} & \makecell*[c]{\includegraphics[width=0.093\linewidth]{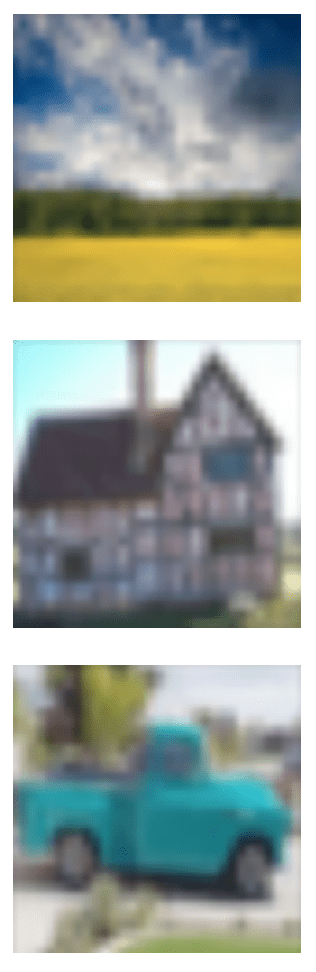}}\\
         \hline
         \ac{LFW} & \makecell*[c]{\includegraphics[width=0.485\linewidth]{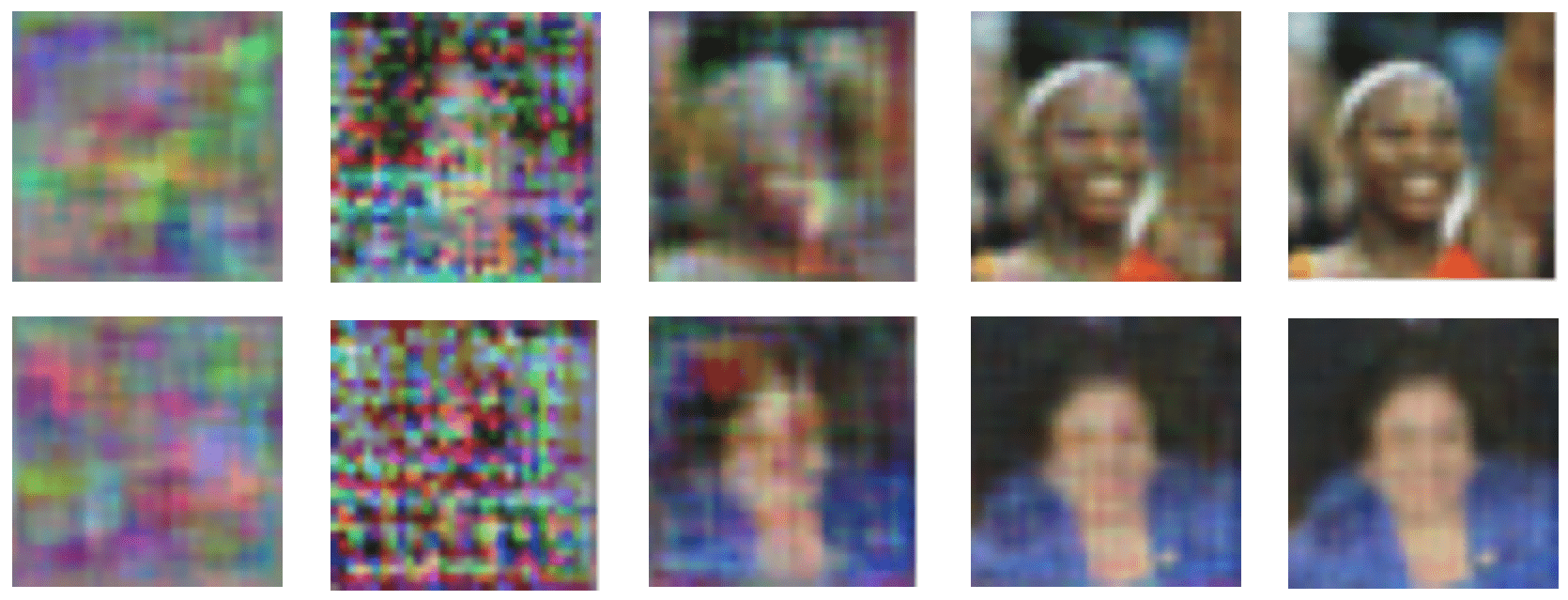}} & \makecell*[c]{\includegraphics[width=0.093\linewidth]{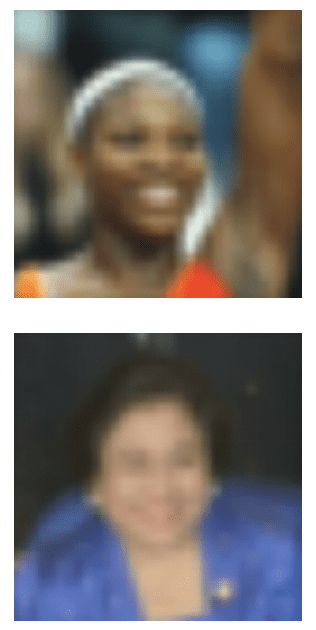}}\\
         \hline
    \end{tabular}
\end{table}

\begin{table}
    \centering
    \caption{Examples of data leakage attack using the proposed \ac{GRNN} on converged global model.}
    \label{tab:TrainingExamples2}
    \begin{tabular}{c|c|c}
         \hline
         \textbf{Dataset} & \textbf{Generated Data} & \textbf{True Data}  \\
         \hline
         MNIST & \makecell*[c]{\includegraphics[width=0.485\linewidth]{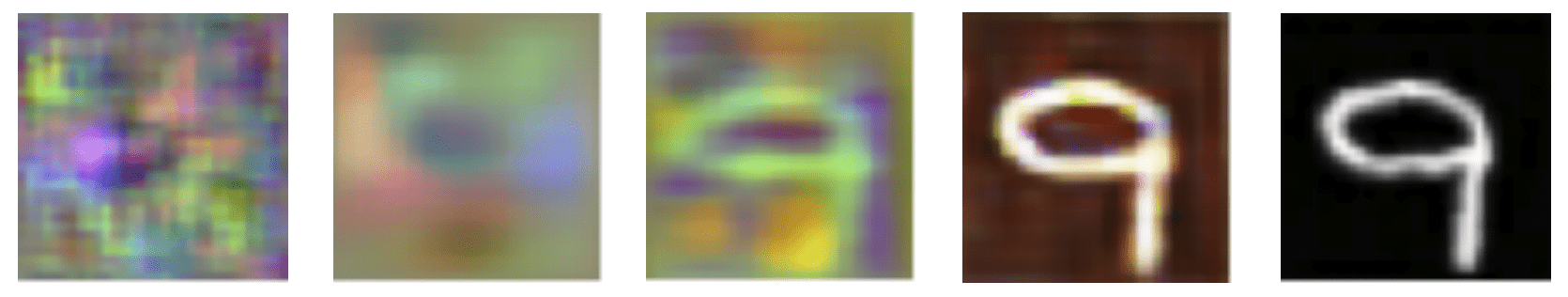}} & \makecell*[c]{\includegraphics[width=0.093\linewidth]{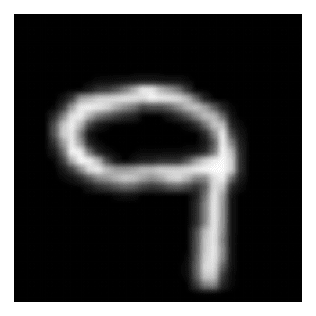}}\\
         \hline
         CIFAR-100 & \makecell*[c]{\includegraphics[width=0.485\linewidth]{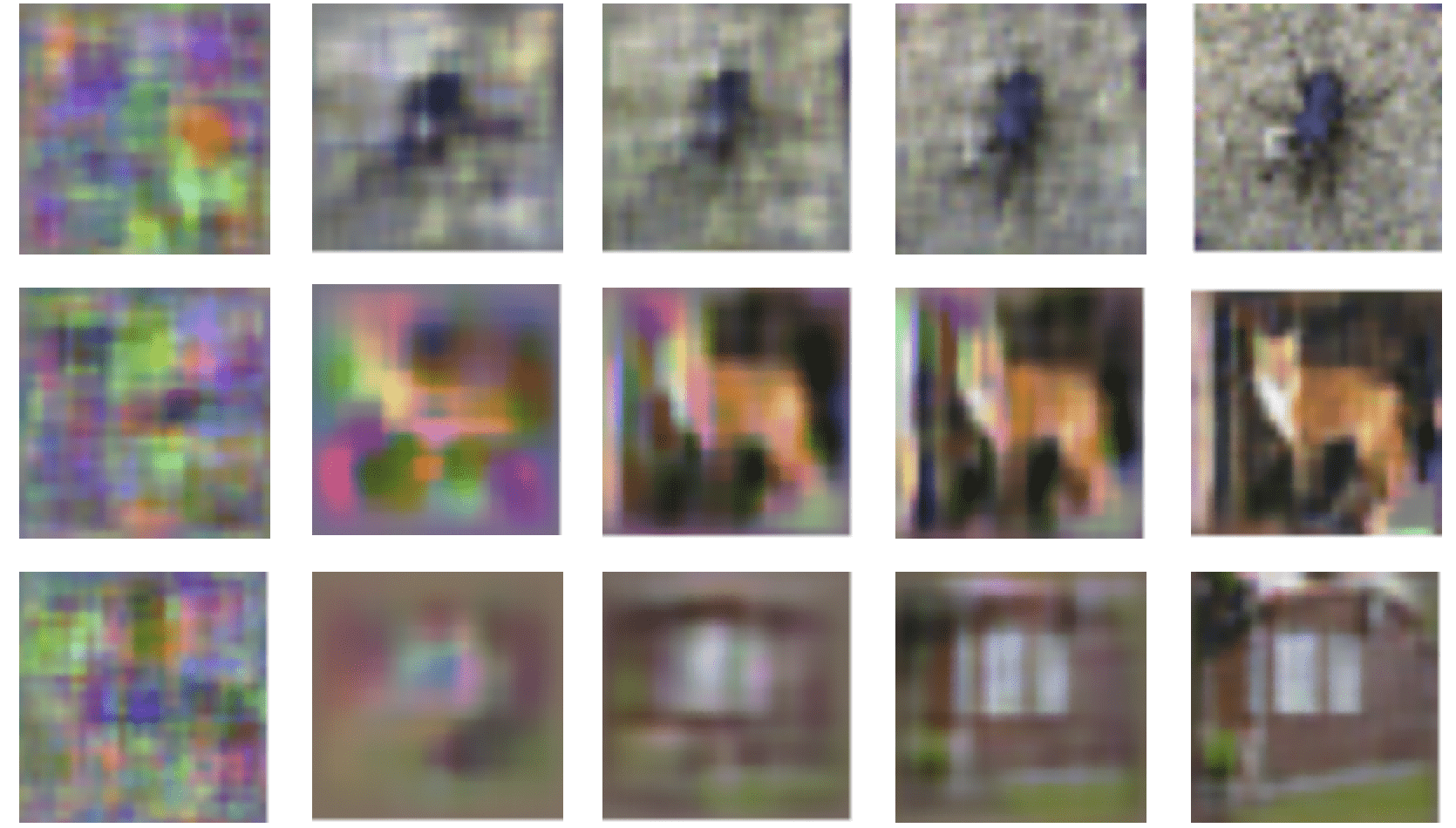}} & \makecell*[c]{\includegraphics[width=0.093\linewidth]{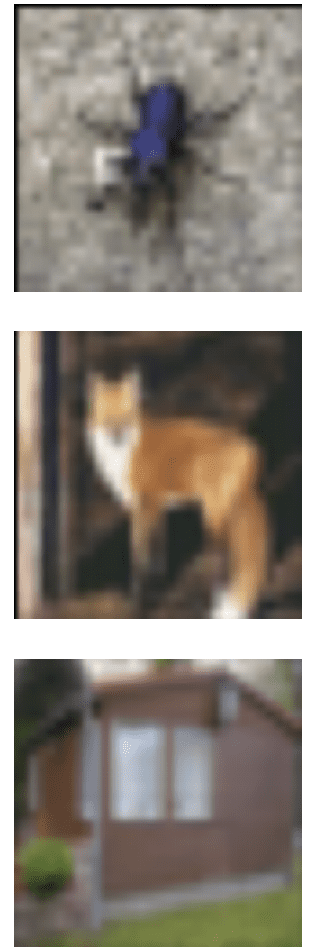}}\\
         \hline
         \ac{LFW} & \makecell*[c]{\includegraphics[width=0.485\linewidth]{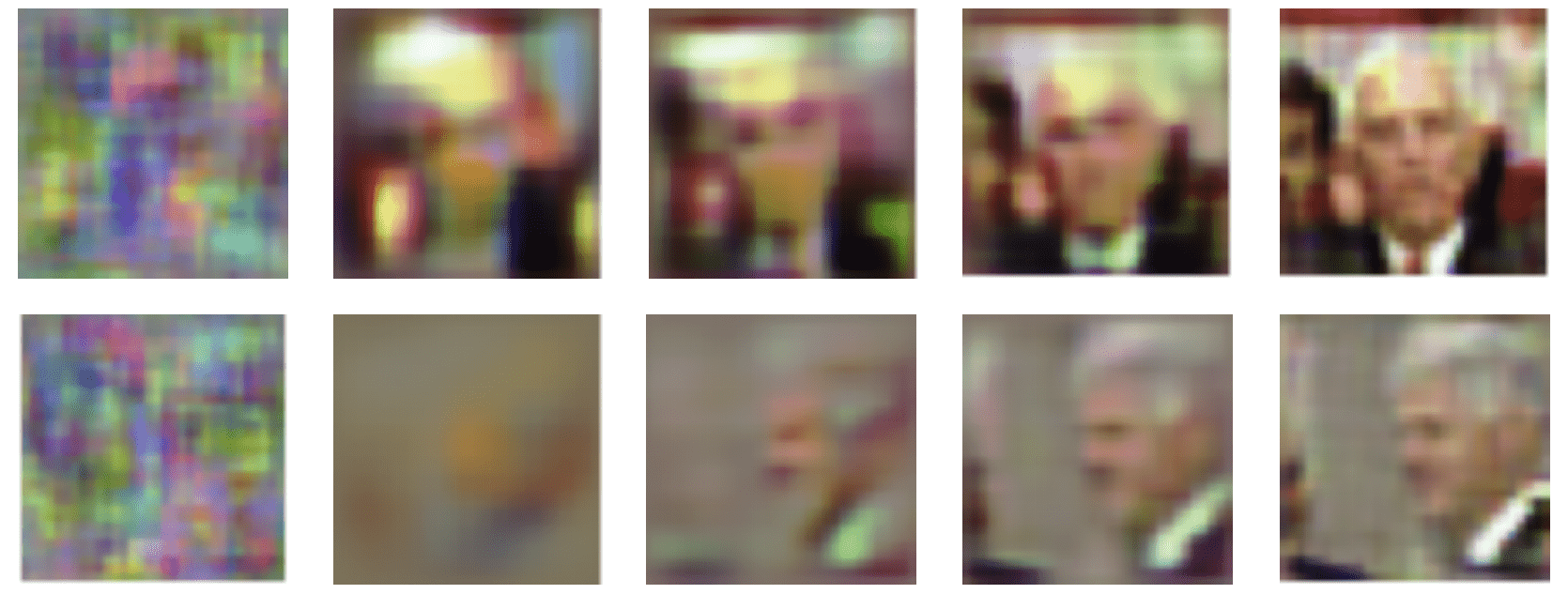}} & \makecell*[c]{\includegraphics[width=0.093\linewidth]{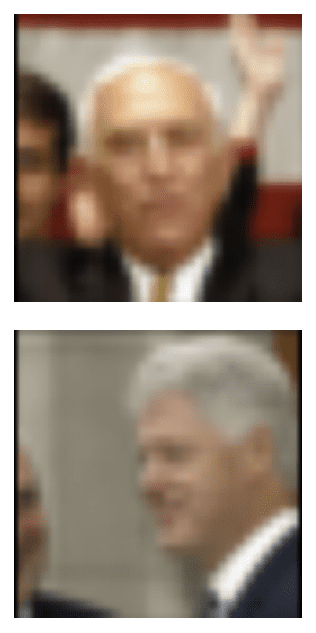}}\\
         \hline
    \end{tabular}
\end{table}

We first trained three \ac{CNN} based image classifiers over one iteration on MNIST, CIFAR-100, and \ac{LFW} datasets separately, and then used the proposed \ac{GRNN} to conduct data leakage attack on those models. In order to demonstrate that our method has no requirement on the convergence of the shared global model and deployment flexibility, we conducted two sets of experiments of the data leakage attack using \ac{GRNN}. One attack was carried out after the first iteration (non-converged state) and another one was carried out after the change of the loss of shared global model is sufficiently small $\delta G\le 1e-4$ (converged state). 

Some qualitative results are presented in Table~\ref{tab:TrainingExamples}, where the middle column shows the image data recovered from the true gradient. The images are recovered gradually with the number of iterations for training \ac{GRNN} increases, while the true gradient is obtained only at the first iteration of global \ac{FL} model, which indicates the gradient can lead to data leakage in \ac{FL} regardless the convergence of shared global model. Table~\ref{tab:TrainingExamples2} shows that image data generated from converged global model performs worse than from non-converged global model. We argue that the reason for lower performance is caused by a large proportion of zero gradients produced by the converged model. We also quantify the quality of generated image using \ac{PSNR} score, an objective standard for image evaluation which is defined as the logarithm of the ratio of the squared maximum value of RGB image fluctuation over \ac{MSE} between two images. The formal definition is given as such: $PSNR = 10 \cdot \lg(\frac{255^2}{MSE(img1, img2)})$. The higher \ac{PSNR} score, the higher the similarity between two images. Table \ref{tab:PSNRresult} gives the average \ac{PSNR} scores achieved by \ac{DLG} and \ac{GRNN} with the batch size of 1 using a non-converged and a converged global model. Overall, the non-converged model performs better than the converged model, except using \emph{ResNet-18} on the MNIST dataset, which achieves 0.70 dB less than the converged model (42.70 dB VS. 43.40 dB). \ac{DLG} failed to recover the training image on MNIST and \ac{LFW} datasets, as there is no visually recognizable image generated.

\begin{table}
\setlength{\tabcolsep}{10pt}
\begin{center}
\caption{Average \ac{PSNR} scores achieved by \ac{DLG} and \ac{GRNN} with the batch size of 1 using non-converged and converged global model. ``-" represents that there is no understandable visual image generated.}
\label{tab:PSNRresult}
\begin{tabular}{c|c|c|c|c}
\hline
\textbf{Method} & \textbf{Model} & \textbf{Dataset} & \textbf{Non-Converged} & \textbf{Converged}  \\
\hline
\multirow{3}{*}{DLG}
 & \multirow{3}{*}{LeNet}
 & MNIST & 50.90 & - \\
&& CIFAR-100 & \textbf{48.59} & \textbf{40.97} \\
&& LFW & 45.05 & - \\
\hline
\multirow{6}{*}{Ours}
 & \multirow{3}{*}{LeNet}
  & MNIST & \textbf{52.37} & 39.47 \\
&& CIFAR-100 & 47.58 & 38.44 \\
&& LFW & \textbf{45.57} & \textbf{38.17} \\

\cline{2-5} & \multirow{3}{*}{ResNet-18}
  & MNIST & 42.70 & \textbf{43.40} \\
&& CIFAR-100 & 38.85 & 38.03 \\
&& LFW & 39.60 & 38.02 \\
\hline
\end{tabular}
\end{center}
\end{table}

\begin{figure}
\subfigcapskip=-8pt
\centering
    \subfigure[]{\includegraphics[width=0.375\linewidth]{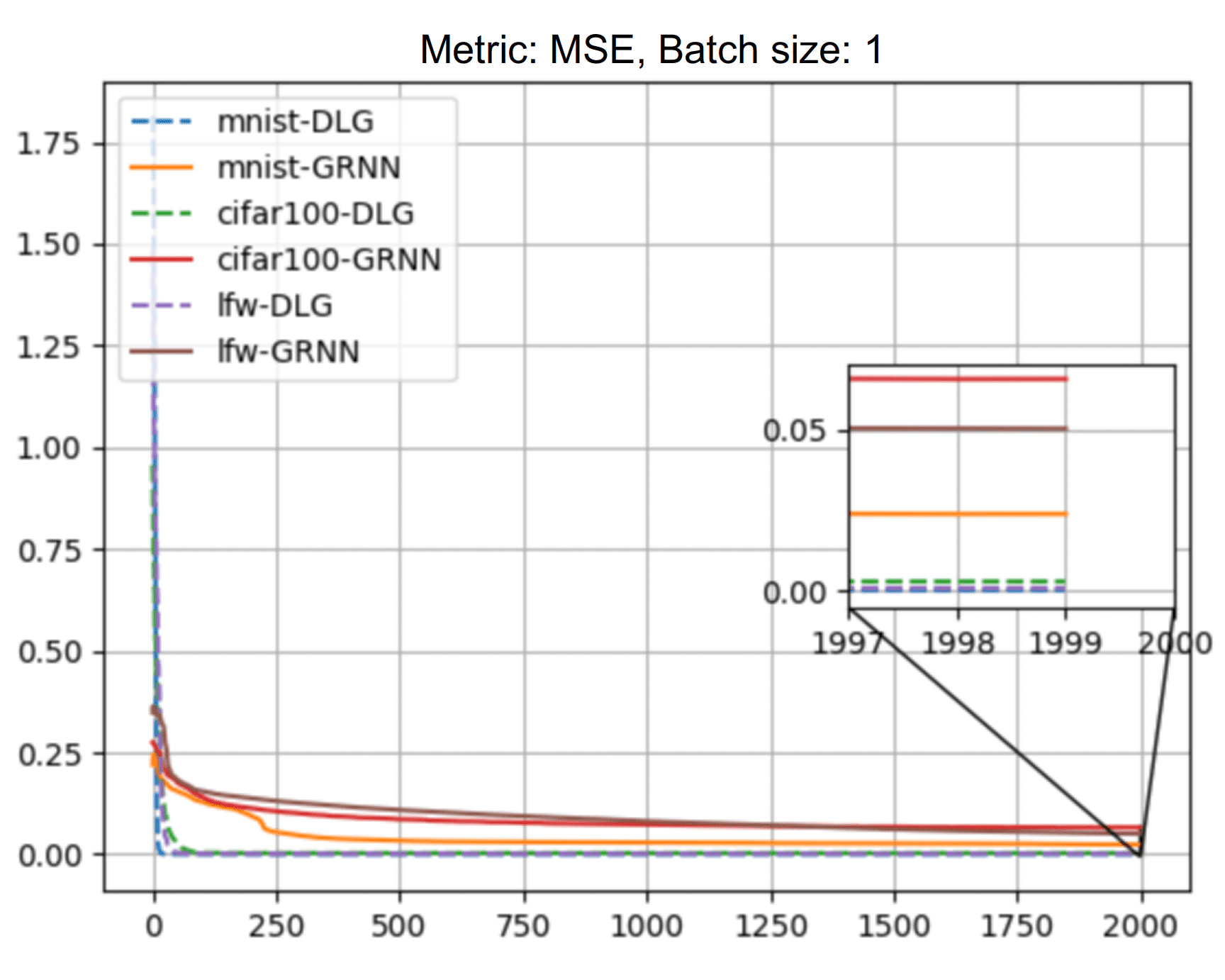}}
    \hspace{.5in}
    \subfigure[]{\includegraphics[width=0.375\linewidth]{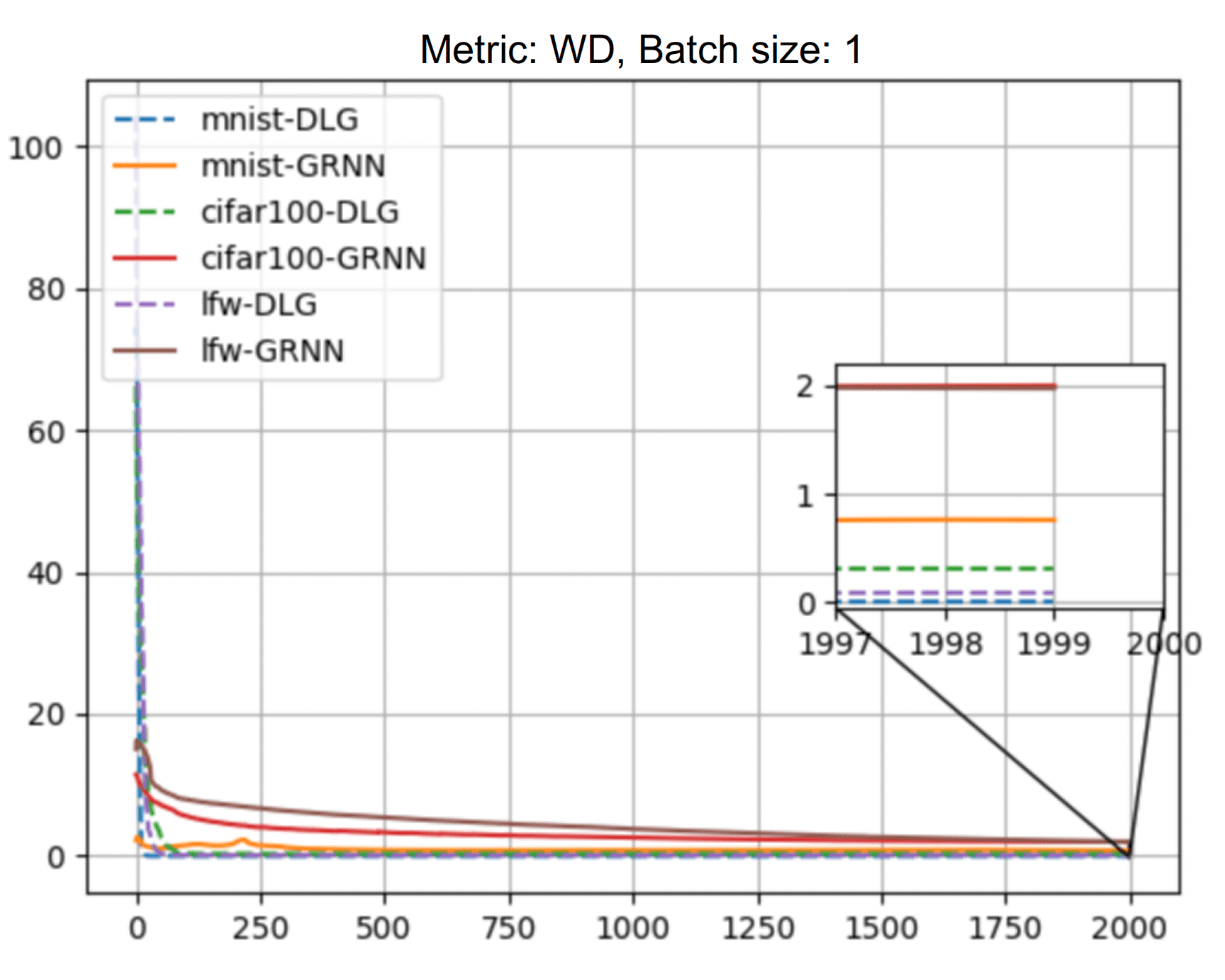}} \\
    \vspace{-.14in}
    \subfigure[]{\includegraphics[width=0.375\linewidth]{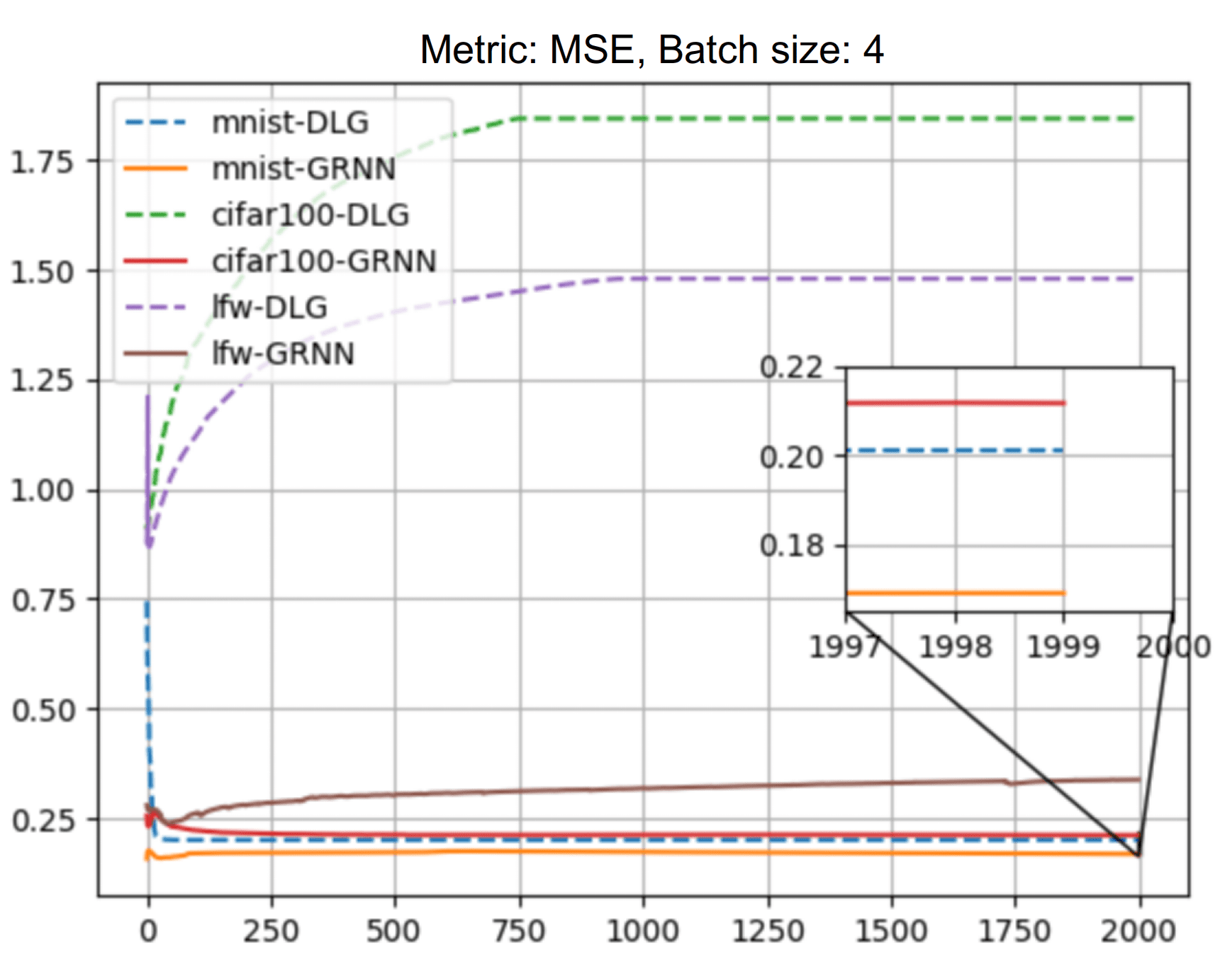}}
    \hspace{.5in}
    \subfigure[]{\includegraphics[width=0.375\linewidth]{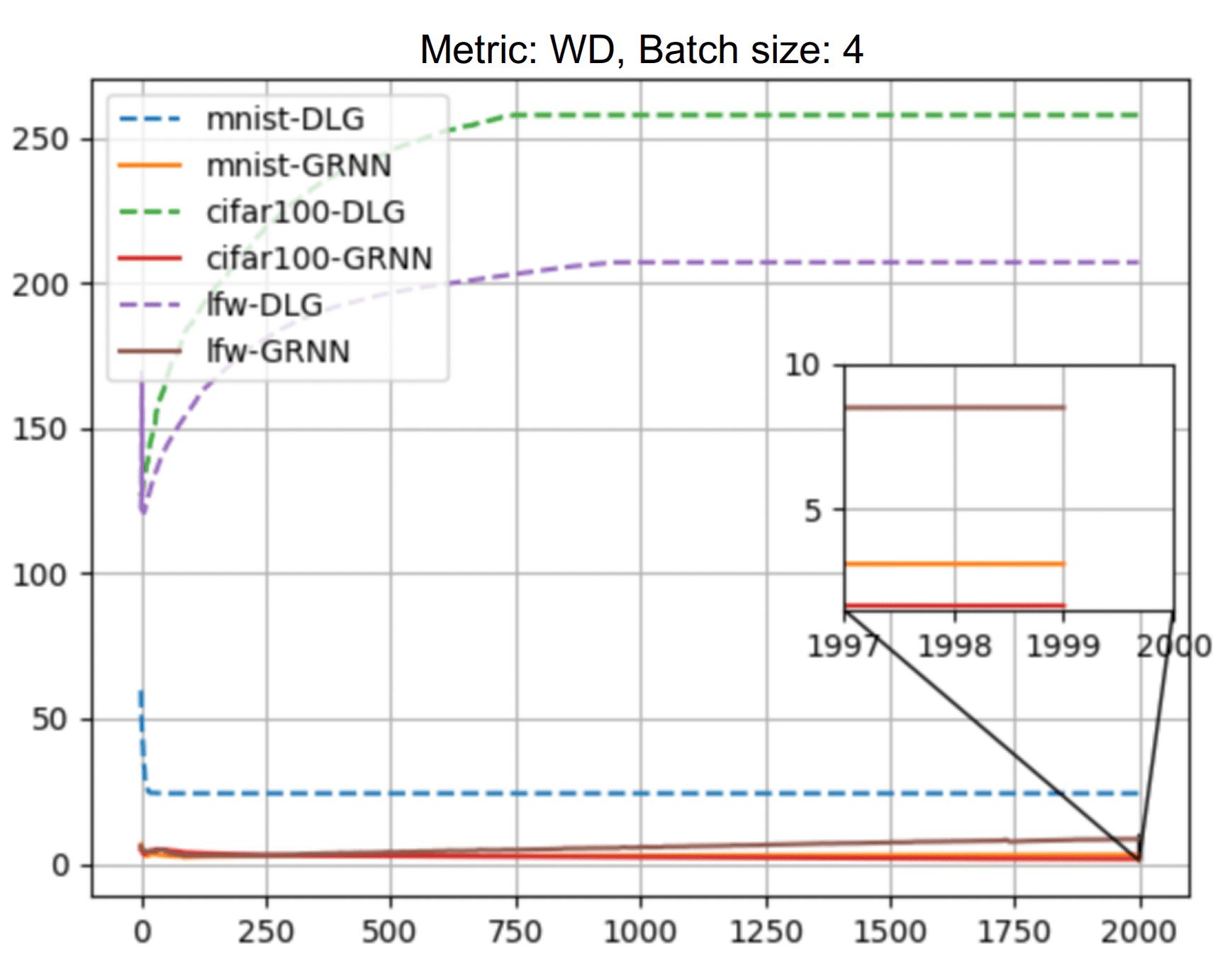}} \\
    \vspace{-.14in}
    \subfigure[]{\includegraphics[width=0.375\linewidth]{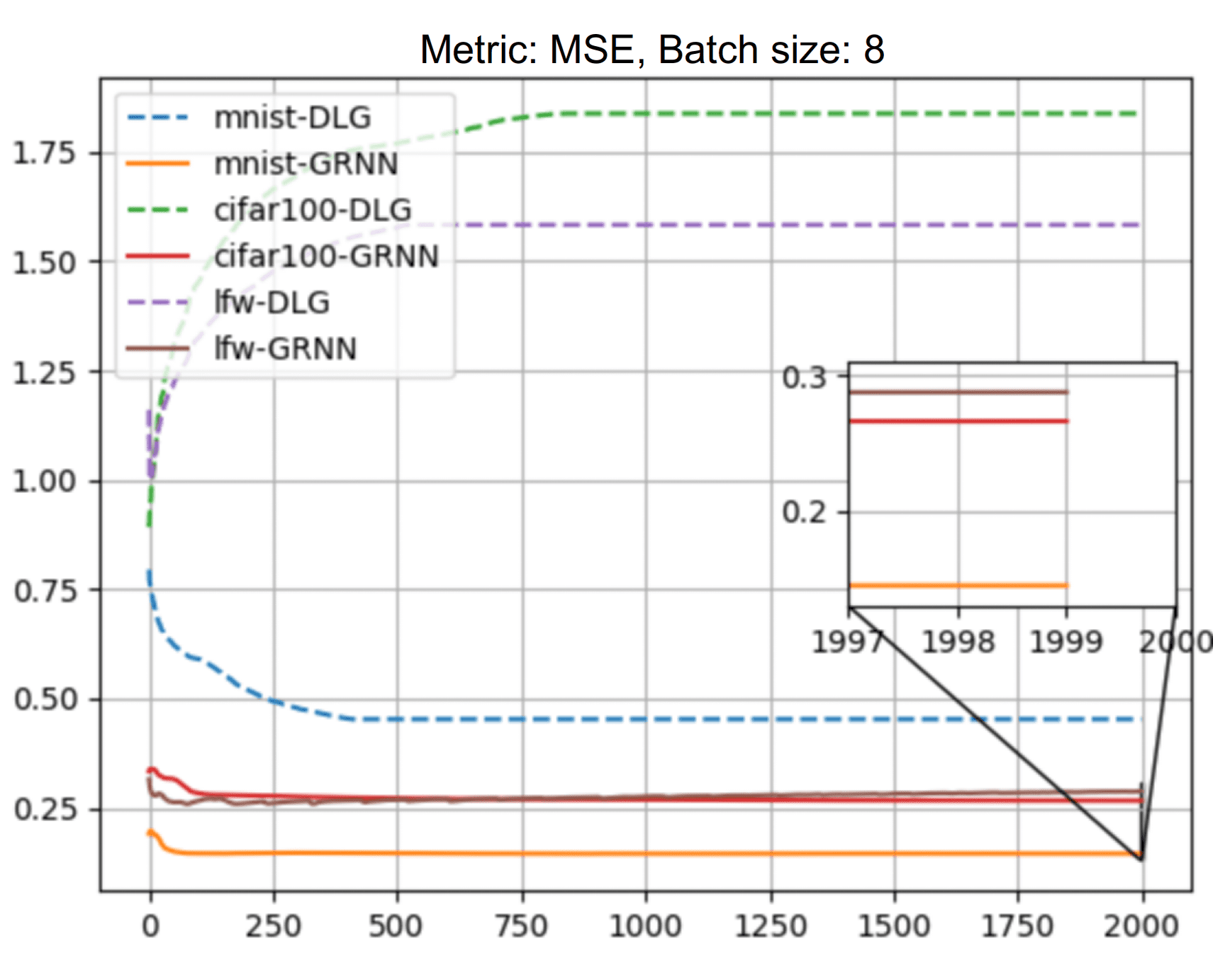}}
    \hspace{.5in}
    \subfigure[]{\includegraphics[width=0.375\linewidth]{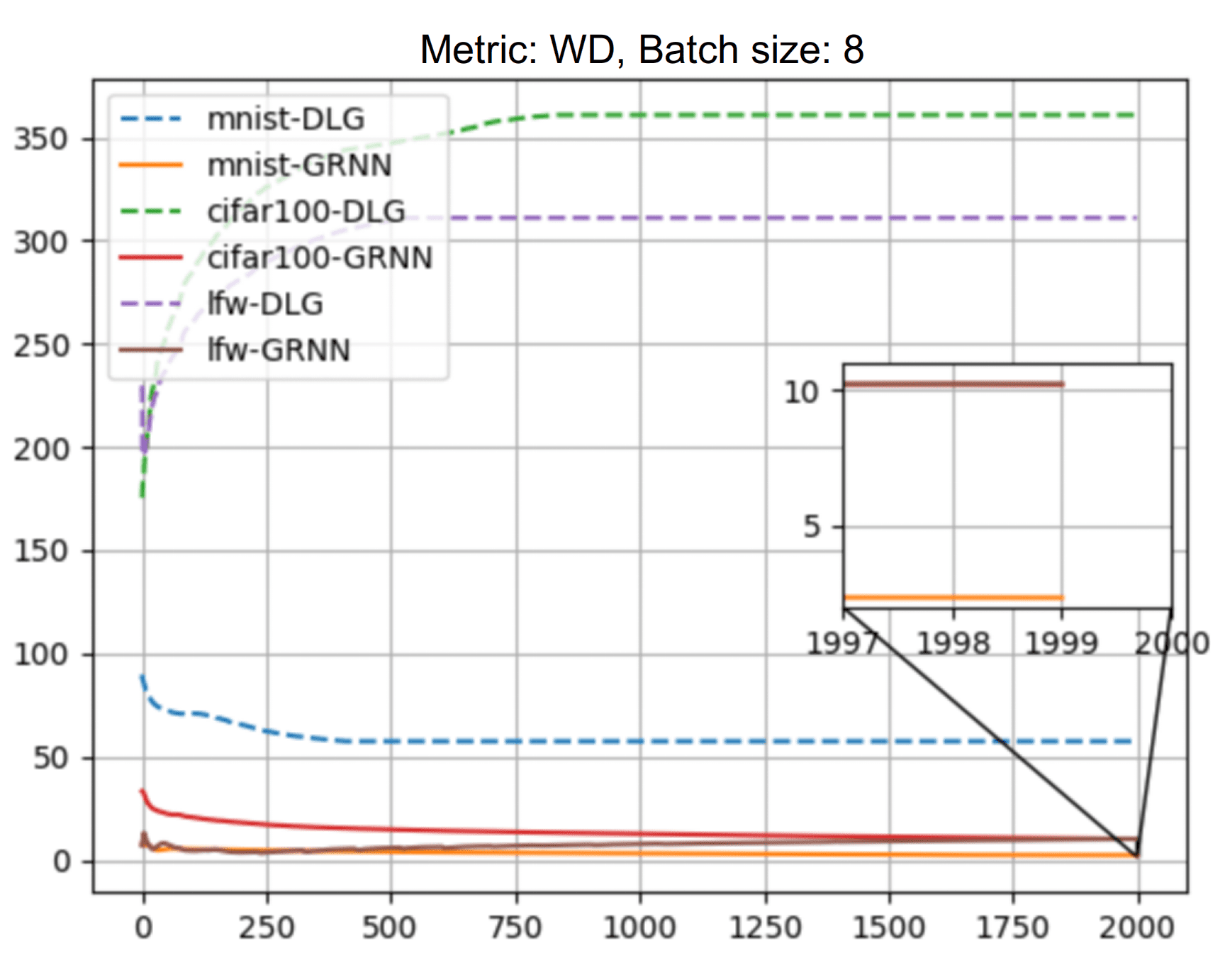}} \\
    \vspace{-.14in}
    \subfigure[]{\includegraphics[width=0.375\linewidth]{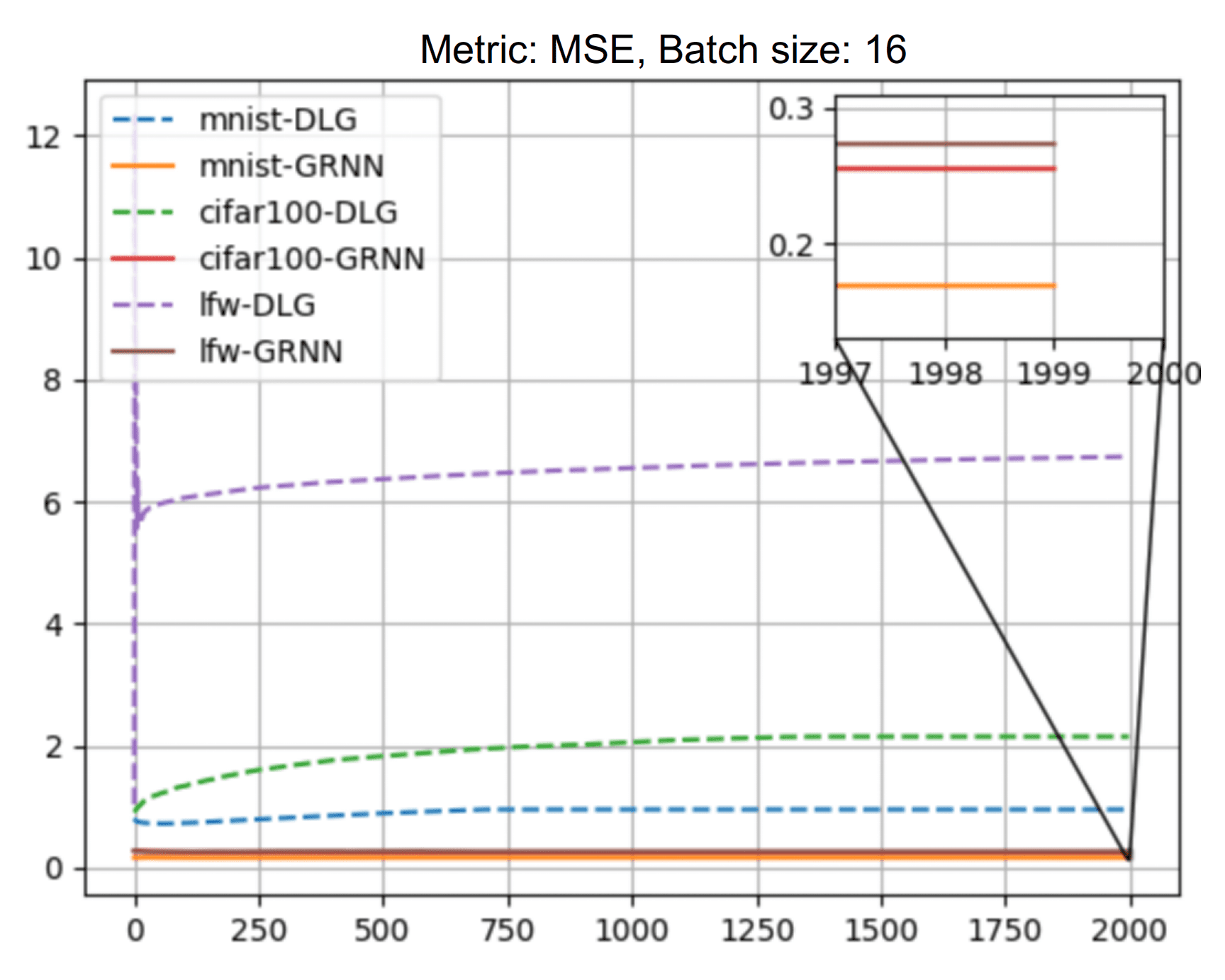}}
    \hspace{.5in}
    \subfigure[]{\includegraphics[width=0.375\linewidth]{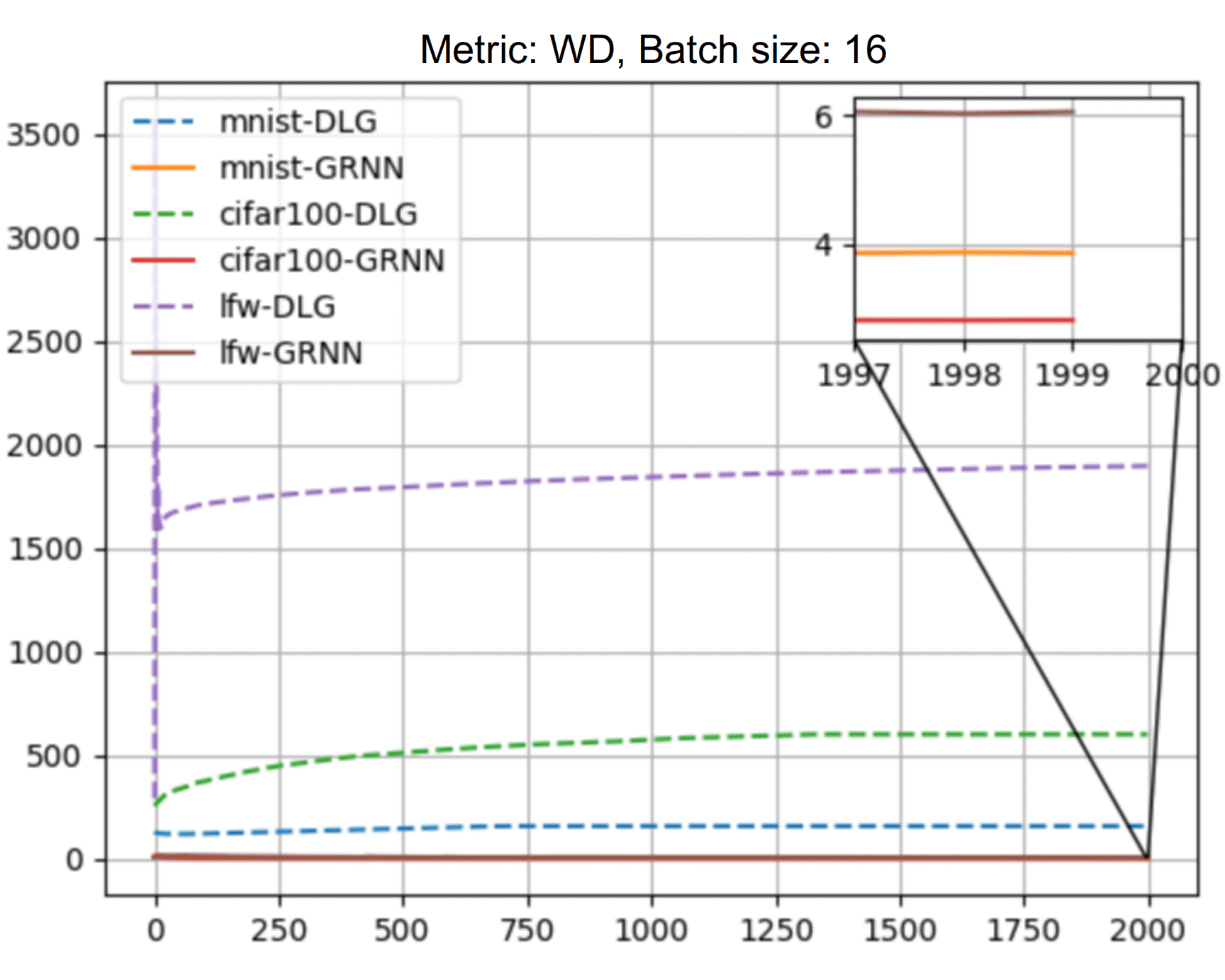}}
    \vspace{-.20in}
\caption{Distances between true and generated images with respect to training iteration for \ac{DLG} and \ac{GRNN} on three datasets. The horizontal axis corresponds to the number of training iterations of two attacking models and the vertical axis corresponds to the similarity metrics.}
\label{fig:TrainingResults}
\end{figure}

\begin{figure}
\centering
    \begin{center}
        \includegraphics[width=0.60\linewidth]{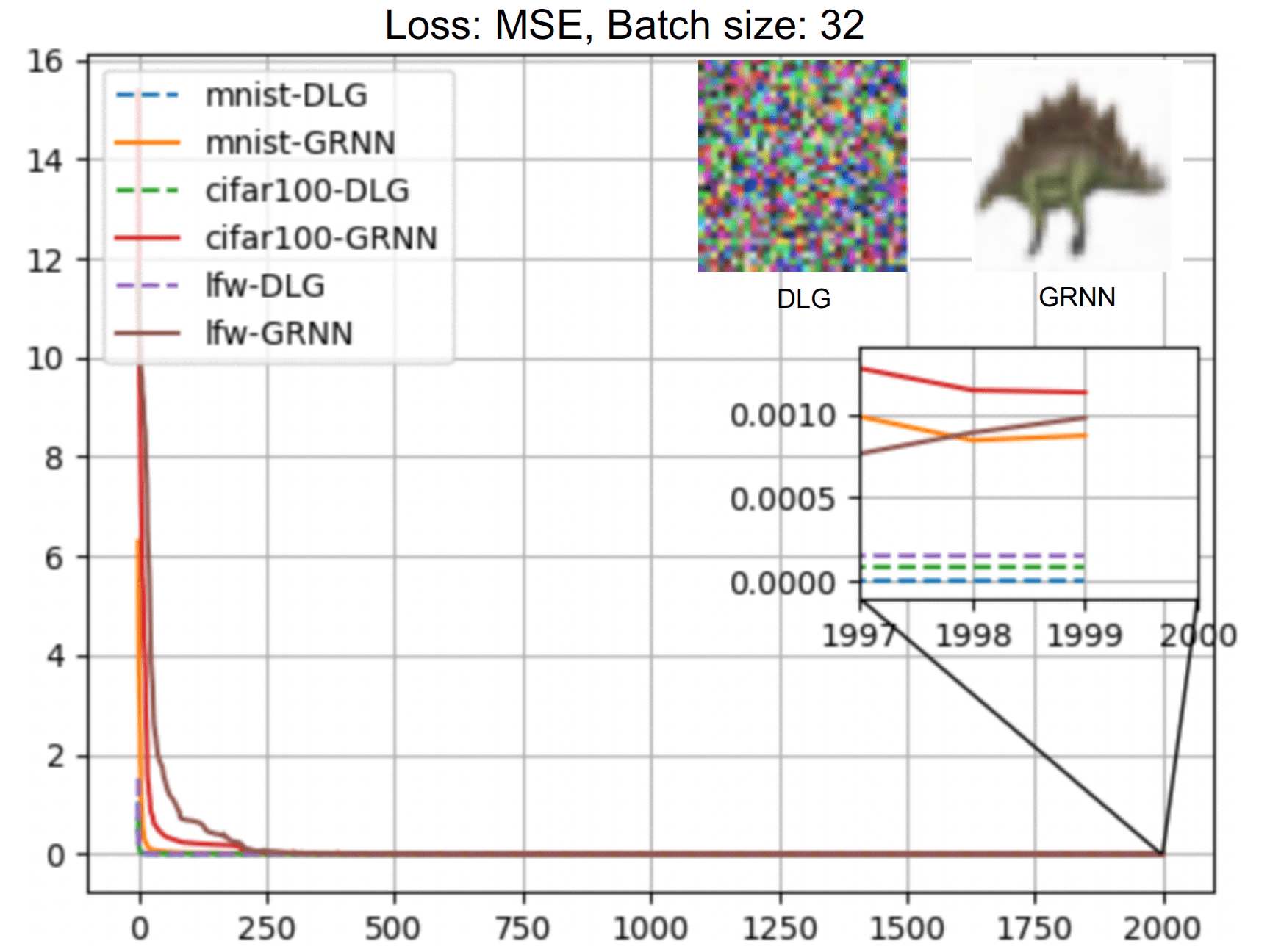}
    \end{center}
\caption{\ac{MSE} loss visualization for \ac{DLG} and \ac{GRNN} with batch size 32.}
\label{fig:TrainingLoss}
\end{figure}

To quantitatively evaluate the similarity of recovered images and true images, three metrics, namely \ac{MSE}, \ac{WD} and \ac{PSNR}~\cite{hore2010image} are computed. Fig.~\ref{fig:TrainingResults} shows the \ac{MSE} and \ac{WD} between recovered images and true images with respect to training iteration of the attacking model. \emph{LeNet} model as the global \ac{FL} model was used in this experiment. The dash lines and the solid lines correspond to the results of \ac{DLG} and \ac{GRNN}, respectively. Although \ac{GRNN} achieves slightly higher scores in \ac{MSE} and \ac{WD} when the batch size of 1 is used, our method is much more stable and significantly better when larger batch size is used. When the batch size increases to 4 and 8, \ac{DLG} only works on MNIST but fails on both CIFAR-100 and \ac{LFW}, therefore, the corresponding similarity measurements can not converge (see the green and purple dashed lines in Figs.~\ref{fig:TrainingResults} (c) (d), (e) and (f)). \ac{DLG} fails on all datasets with batch size of 16 while \ac{GRNN} is able to recover the image data consistently (see Figs.~\ref{fig:TrainingResults} (g) and (h)). We also notice that \ac{DLG} can well approximate the shared true gradient while generating a poor image. Fig.\ref{fig:TrainingLoss} shows that \ac{DLG} achieves smaller \ac{MSE} loss (Euclidean distance between true gradient and fake gradient) compared to our method, while it fails to recover the image data.

In Fig.~\ref{fig:TrainingLoss}, \ac{MSE} and \ac{WD} results from \ac{GRNN} are slightly higher than those from \ac{DLG} with a batch size of 1, however, the difference between these two generated set of images at this batch size is hardly discernible. Hence, we further calculated \ac{PSNR} to compare the pixel-wise similarity of the recovered images. Table~\ref{tab:PSNRresult} shows \ac{GRNN} achieves higher \ac{PSNR} on MNIST and \ac{LFW} datasets (+1.45\% and +0.52\% respectively) and lower on CIFAR-100 dataset (-1.01\%) using non-converged global model. Furthermore, our method achieved reasonable \ac{PSNR} score on attacking \emph{ResNet-18} model while \ac{DLG} always fails. Table \ref{tab:ExamplesComparison} shows some qualitative comparison of recovered images using both methods over different numbers of iterations. We can observe that \ac{DLG} recover the image pixel by pixel greedily, whereas \ac{GRNN} also ensures the appearance distribution to be consistent with the true image in a coarser scale, and object details are then gradually filled at a finer scale.

\begin{table}
    \centering
    \caption{Comparison of image recovery using \ac{DLG} and \ac{GRNN} over different numbers of iterations.}
    \label{tab:ExamplesComparison}
    \begin{tabular}{c|c|c}
        \hline
        \textbf{True Data} & \textbf{\ac{DLG}} &\textbf{\ac{GRNN}} \\
        \hline
        \makecell*[c]{\includegraphics[width=0.07\linewidth]{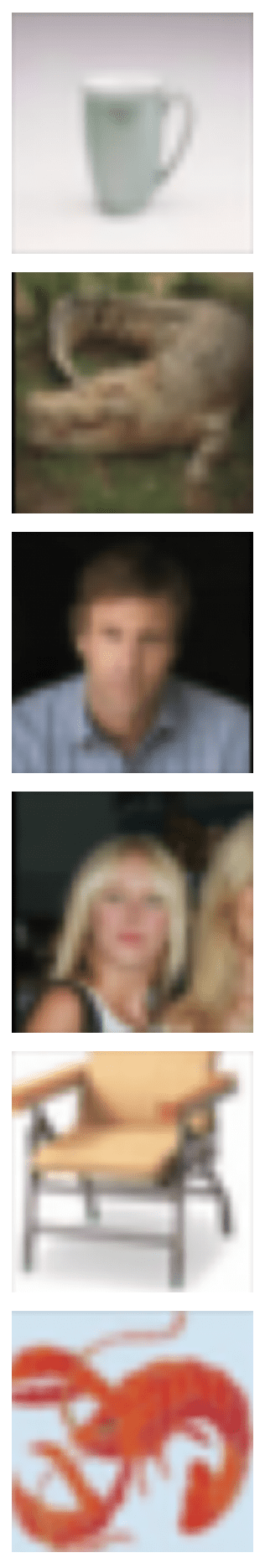}} & \makecell*[c]{\includegraphics[width=0.405\linewidth]{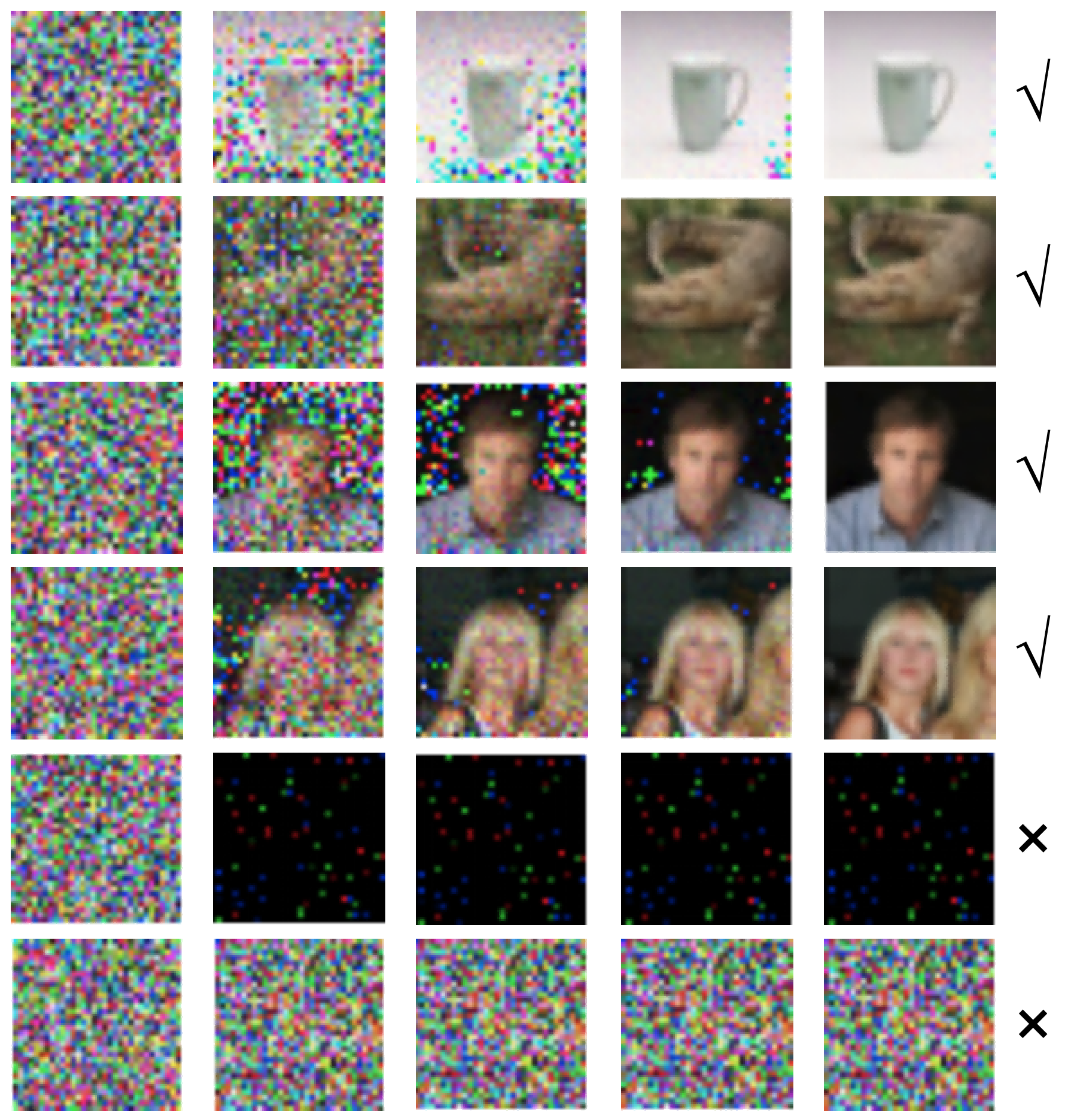}} & \makecell*[c]{\includegraphics[width=0.405\linewidth]{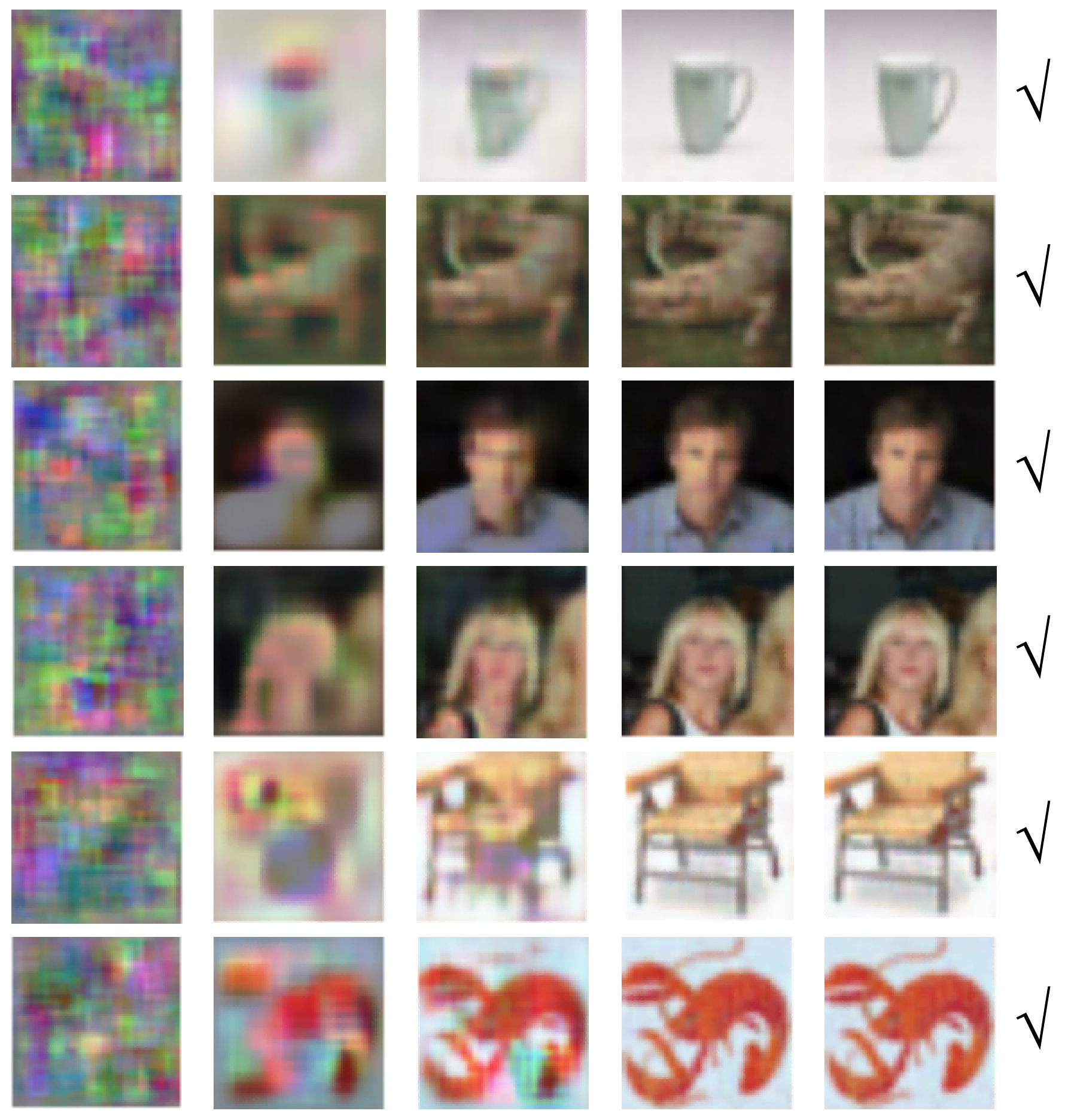}} \\
        \hline
    \end{tabular}
\end{table}

\subsubsection{Ablation Study on Batch Size}

A comparison study between the proposed \ac{GRNN} and \ac{DLG} was carried out using the same setting as described above, while we varied the training batch size for \ac{FL} model to evaluate the feasibility of data leakage attack. It is reasonable to consider that the data recovery is more challenging when the training batch size is increasing as the shared gradient is averaged over all images data in the batch, where information of an individual image is obscurely mixed. Table \ref{tab:AttackPerformance} lists the success and failure of attack for both two methods. We follow the same principle defined in \ac{DLG} paper, where a successful attack refers to recovering an image that is visually recognizable. On MNIST dataset, \ac{DLG} attack starts to fail when the size of training batch is larger than 8 and the \emph{LeNet} is used in \ac{FL}, whilst our method is able to recover the image data even with a batch size of 256. On CIFAR-100 and \ac{LFW} datasets, \ac{DLG} attack only works with a training batch size of 1, however, our method can still successfully perform the attack with a large batch size up to 64 and 128 respectively. Some failure examples of our method when a large training batch size is used can be found in Table~\ref{tab:FailureExample}. Although some failure examples show the consistency of color distribution and geometric similarity of objects, the appearance details are largely inconsistent or hard to match the original ones. We also show that the proposed \ac{GRNN} can successfully attack complex and large models, such as \emph{ResNet-18}.

\begin{table}
\setlength{\tabcolsep}{3pt}
\begin{center}
\caption{Data leakage attack with different training batch sizes for \ac{FL} model, where ``\checkmark'' refers to a success and ``$\times$'' refers to a failure.}
\label{tab:AttackPerformance}
\begin{tabular}{>{\centering\arraybackslash}c|c|c|>{\centering\arraybackslash}p{0.2in}|>{\centering\arraybackslash}p{0.2in}|>{\centering\arraybackslash}p{0.2in}|>{\centering\arraybackslash}p{0.2in}|>{\centering\arraybackslash}p{0.2in}|>{\centering\arraybackslash}p{0.2in}|>{\centering\arraybackslash}p{0.2in}|>{\centering\arraybackslash}p{0.2in}}
\hline
\textbf{Method} & \textbf{Model} & \diagbox{\textbf{Dataset}}{\textbf{\#Batch}} & \textbf{1} & \textbf{4} & \textbf{8} & \textbf{16} & \textbf{32} & \textbf{64} & \textbf{128} & \textbf{256}  \\
\hline
\multirow{3}{*}{DLG}
 & \multirow{3}{*}{LeNet}
 & MNIST & \checkmark & \checkmark & \checkmark & $\times$ & $\times$ & $\times$ & $\times$ & $\times$ \\
&& CIFAR-100 & \checkmark & $\times$ & $\times$ & $\times$ & $\times$ & $\times$ & $\times$ & $\times$ \\
&& LFW & \checkmark & $\times$ & $\times$ & $\times$ & $\times$ & $\times$ & $\times$ & $\times$  \\
\hline
\multirow{6}{*}{Ours}
 & \multirow{3}{*}{LeNet}
  & MNIST & \checkmark & \checkmark & \checkmark & \checkmark & \checkmark & \checkmark & \checkmark & \checkmark \\
&& CIFAR-100 & \checkmark & \checkmark & \checkmark & \checkmark & \checkmark & \checkmark & \checkmark & $\times$ \\
&& LFW & \checkmark & \checkmark & \checkmark & \checkmark & \checkmark & \checkmark & $\times$ & $\times$ \\

\cline{2-11} & \multirow{3}{*}{ResNet-18}
  & MNIST & \checkmark & \checkmark & \checkmark & \checkmark & $\times$ & - & - & - \\
&& CIFAR-100 & \checkmark & \checkmark & $\times$ & $\times$ & $\times$ & - & - & - \\
&& LFW & \checkmark & $\times$ & $\times$ & $\times$ & $\times$ & - & - & - \\
\hline
\end{tabular}
\end{center}
\end{table}

\begin{table}
\begin{center}
\caption{Examples of failed data leakage attack using the proposed \ac{GRNN} on \emph{LeNet} with batch sizes of 128 and 256. The top row shows the fake images recovered from the shared gradient and the bottom row shows corresponding true images in the private datasets.}
\label{tab:FailureExample}
\begin{tabular}{c|c|c}
\hline
\textbf{Dataset} & CIFAR-100 & \ac{LFW}\\
\hline
\textbf{Samples} & \makecell*[c]{\includegraphics[width=0.275\linewidth]{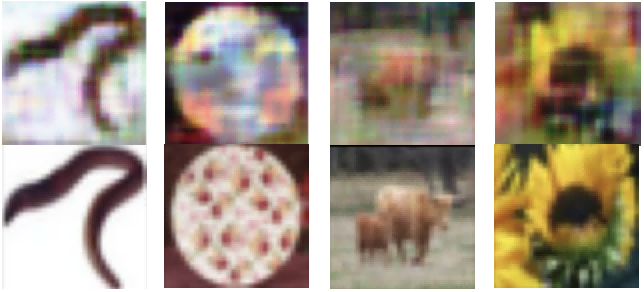}} & \makecell*[c]{\includegraphics[width=0.55\linewidth]{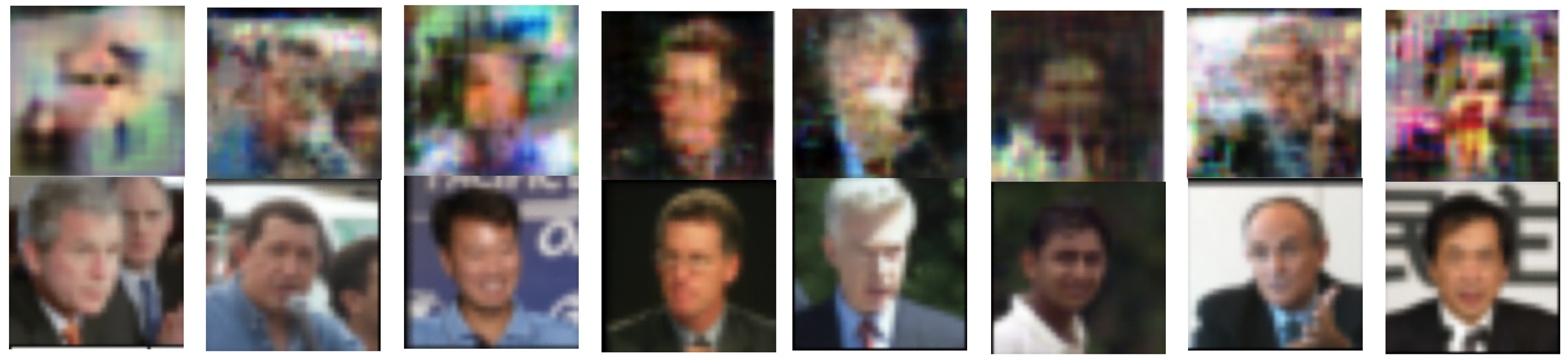}} \\
\hline
\end{tabular}
\end{center}
\end{table}

\begin{table}
    \centering
    \caption{Some randomly selected ground truth (GT) images and corresponding recovered images from \ac{GRNN} and \ac{DLG} with different batch sizes using \emph{LeNet}.}
    \label{tab:SelectedSamples}
    \begin{tabular}{c|c|c}
        \hline
        \multirow{2}{*}{\textbf{Method}} & \multicolumn{2}{c}{\textbf{\#Batch \& Images}} \\
        \cline{2-3}
        & 1 & 4 \\
        \hline
        GT   & \makecell*[c]{\includegraphics[width=0.4\linewidth]{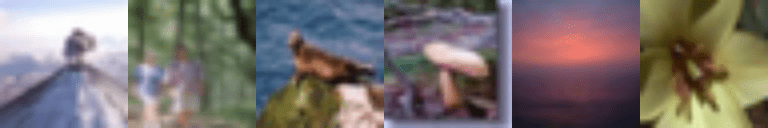}} 
             & \makecell*[c]{\includegraphics[width=0.4\linewidth]{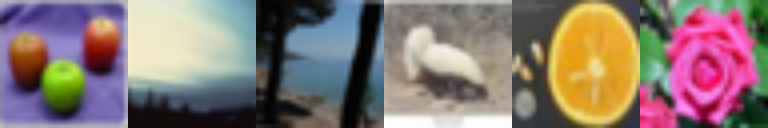}} \\
        GRNN & \makecell*[c]{\includegraphics[width=0.4\linewidth]{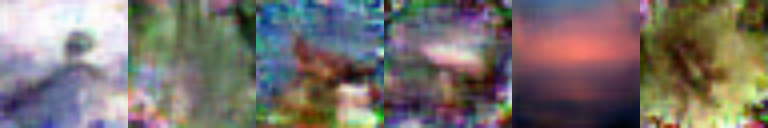}}
             & \makecell*[c]{\includegraphics[width=0.4\linewidth]{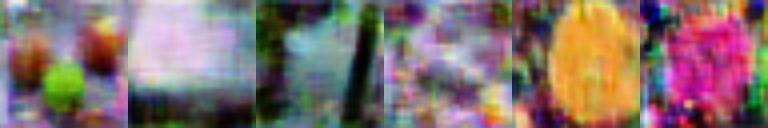}}\\
        DLG  & \makecell*[c]{\includegraphics[width=0.4\linewidth]{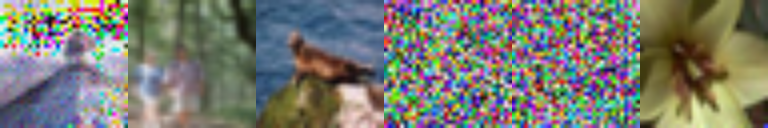}}
             & \makecell*[c]{\includegraphics[width=0.4\linewidth]{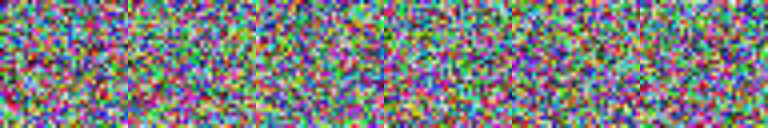}}\\
        \hline
        & 8 & 16 \\
        \hline
        GT   & \makecell*[c]{\includegraphics[width=0.4\linewidth]{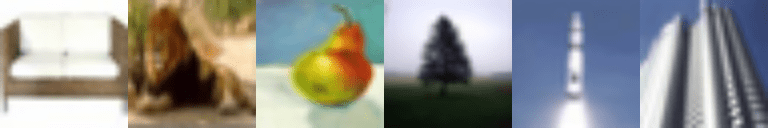}} 
             & \makecell*[c]{\includegraphics[width=0.4\linewidth]{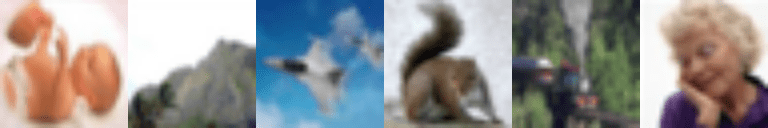}} \\
        GRNN & \makecell*[c]{\includegraphics[width=0.4\linewidth]{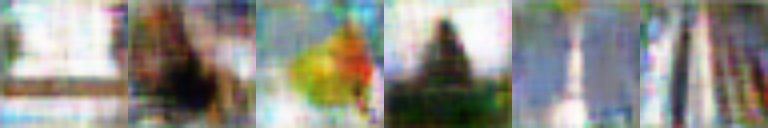}}
             & \makecell*[c]{\includegraphics[width=0.4\linewidth]{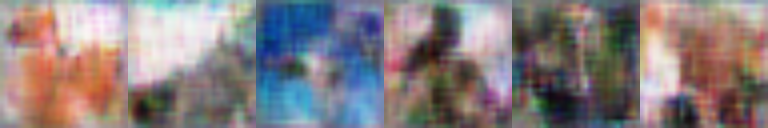}}\\
        DLG  & \makecell*[c]{\includegraphics[width=0.4\linewidth]{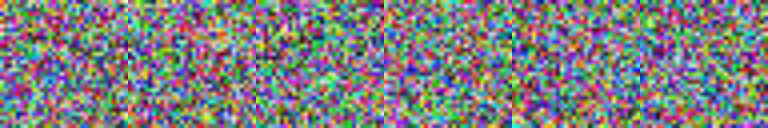}}
             & \makecell*[c]{\includegraphics[width=0.4\linewidth]{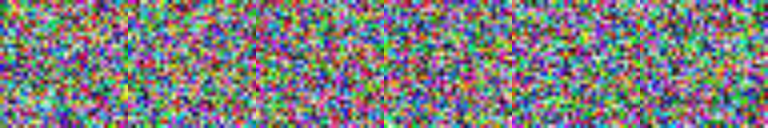}}\\
        \hline
        & 32 & 64 \\
        \hline
        GT   & \makecell*[c]{\includegraphics[width=0.4\linewidth]{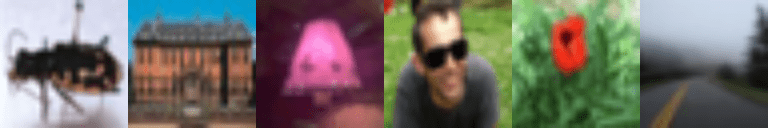}} 
             & \makecell*[c]{\includegraphics[width=0.4\linewidth]{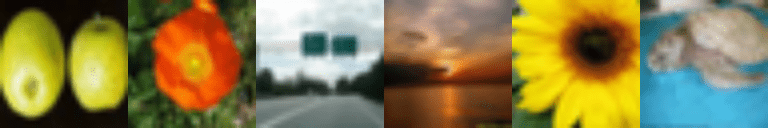}} \\
        GRNN & \makecell*[c]{\includegraphics[width=0.4\linewidth]{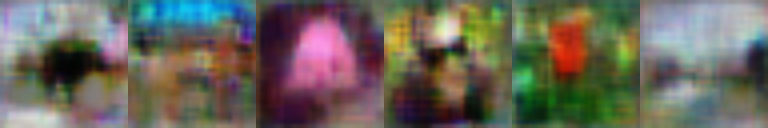}}
             & \makecell*[c]{\includegraphics[width=0.4\linewidth]{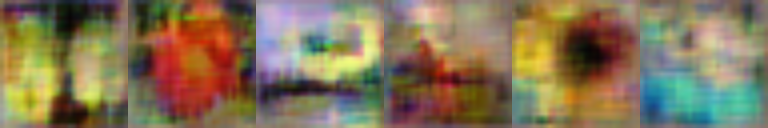}}\\
        DLG  & \makecell*[c]{\includegraphics[width=0.4\linewidth]{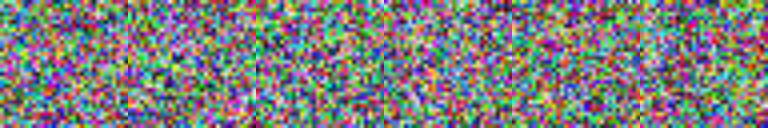}}
             & \makecell*[c]{\includegraphics[width=0.4\linewidth]{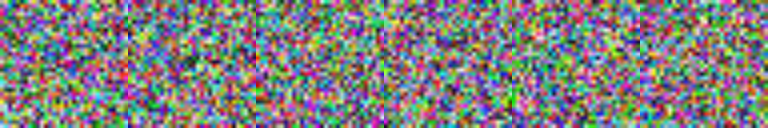}}\\
        \hline
        & 128 & 256 \\
        \hline
        GT   & \makecell*[c]{\includegraphics[width=0.4\linewidth]{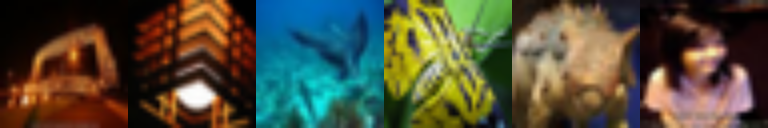}} 
             & \makecell*[c]{\includegraphics[width=0.4\linewidth]{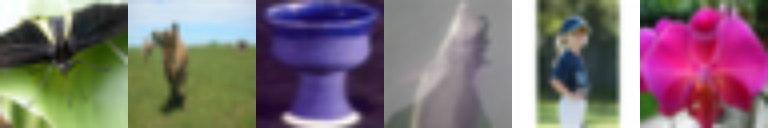}} \\
        GRNN & \makecell*[c]{\includegraphics[width=0.4\linewidth]{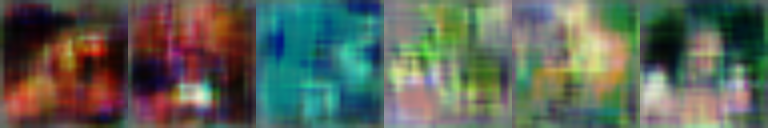}}
             & \makecell*[c]{\includegraphics[width=0.4\linewidth]{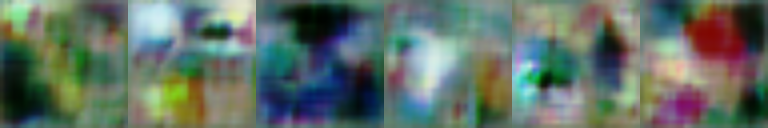}}\\
        DLG  & \makecell*[c]{\includegraphics[width=0.4\linewidth]{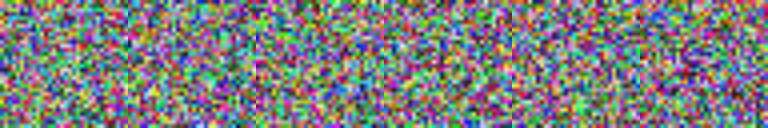}}
             & \makecell*[c]{\includegraphics[width=0.4\linewidth]{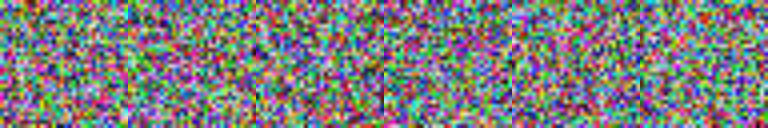}}\\
        \hline
    \end{tabular}
\end{table}

\subsubsection{Ablation Study on Success Rate}

\begin{figure}
\centering
    \subfigure[MSE]{\includegraphics[width=1.8in, height=1.7in]{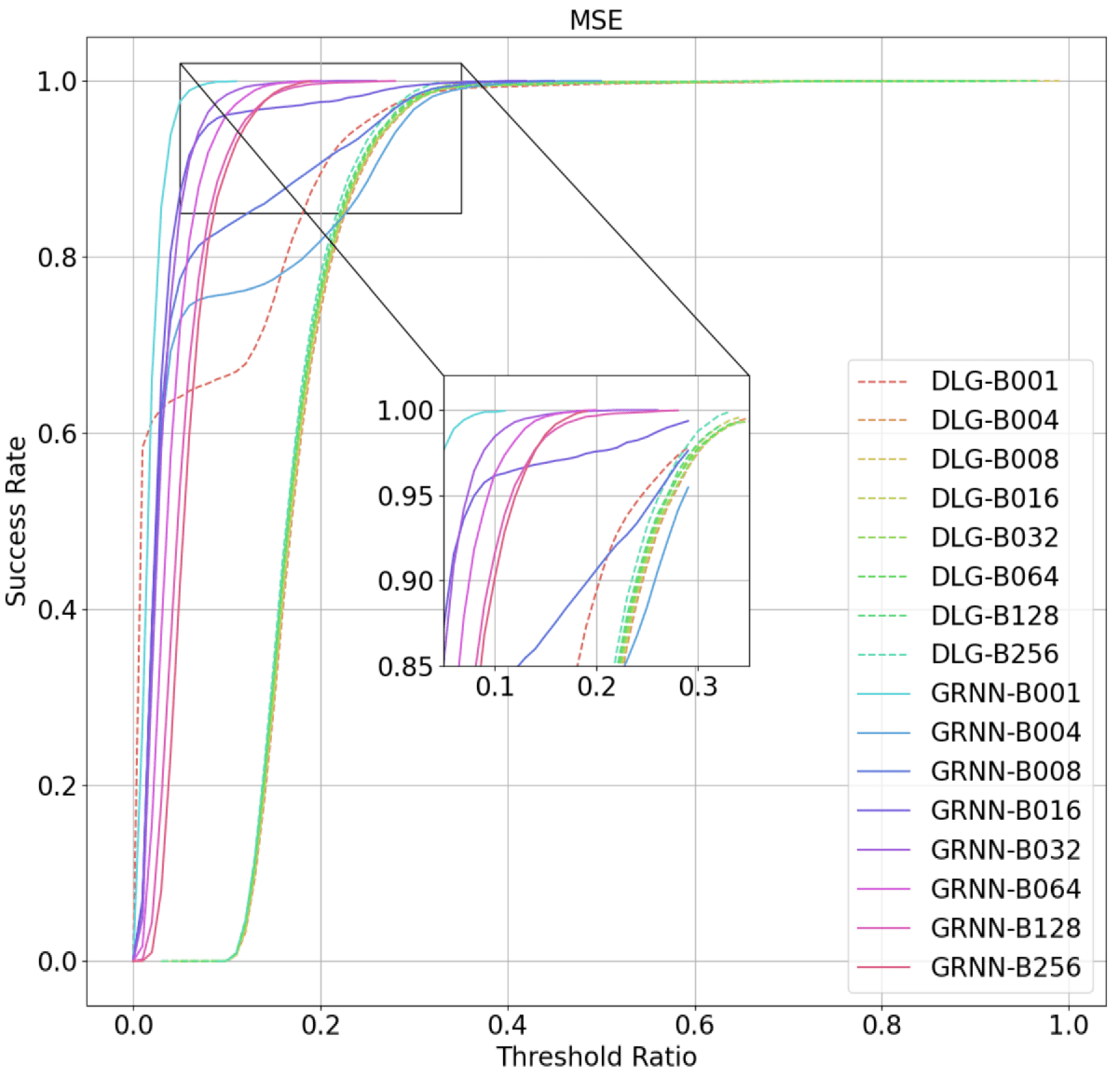}}
    \hspace{-.05in}
    \subfigure[PSNR]{\includegraphics[width=1.8in, height=1.7in]{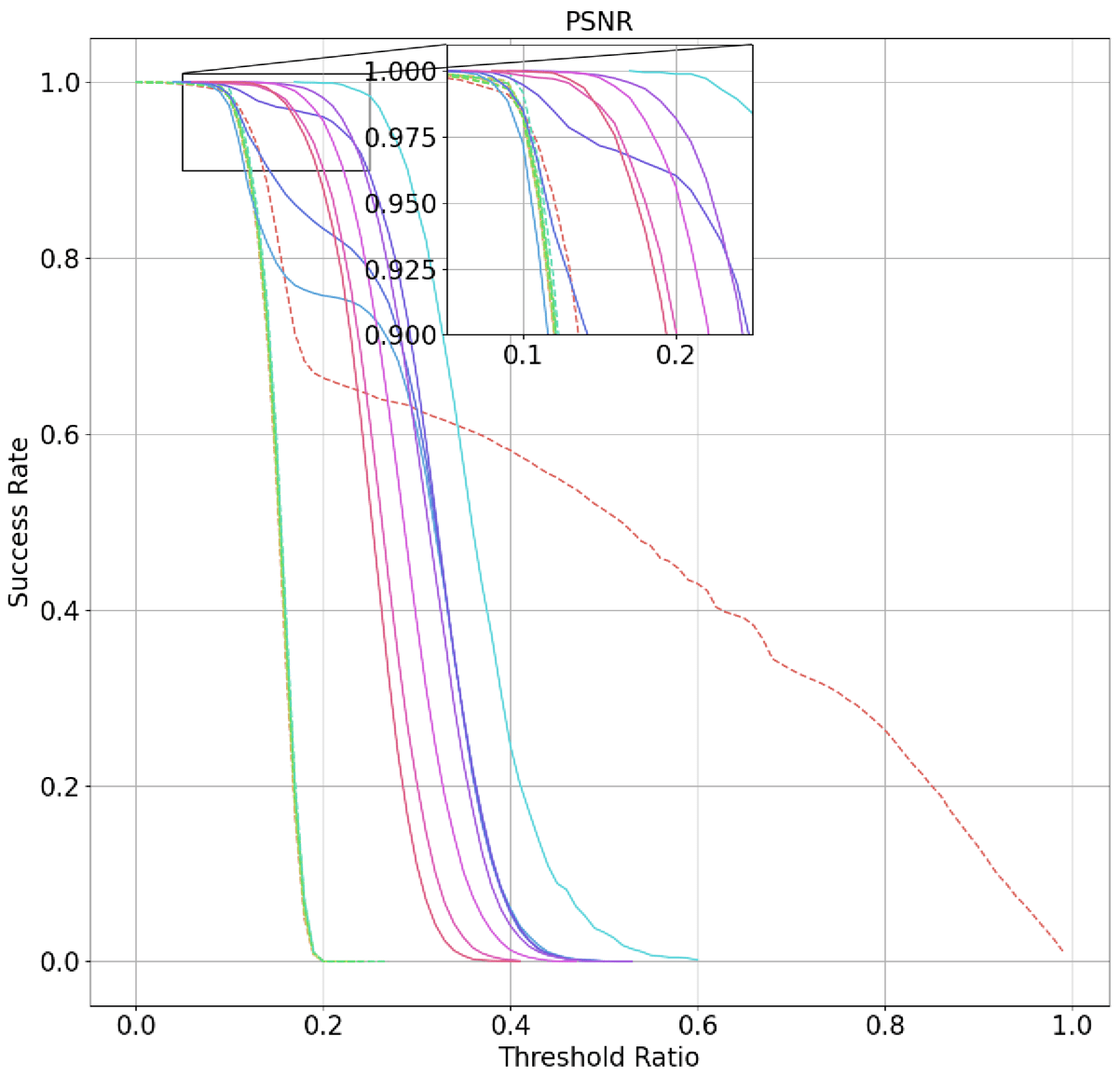}}
    \hspace{-.05in}
    \subfigure[SSIM]{\includegraphics[width=1.8in, height=1.7in]{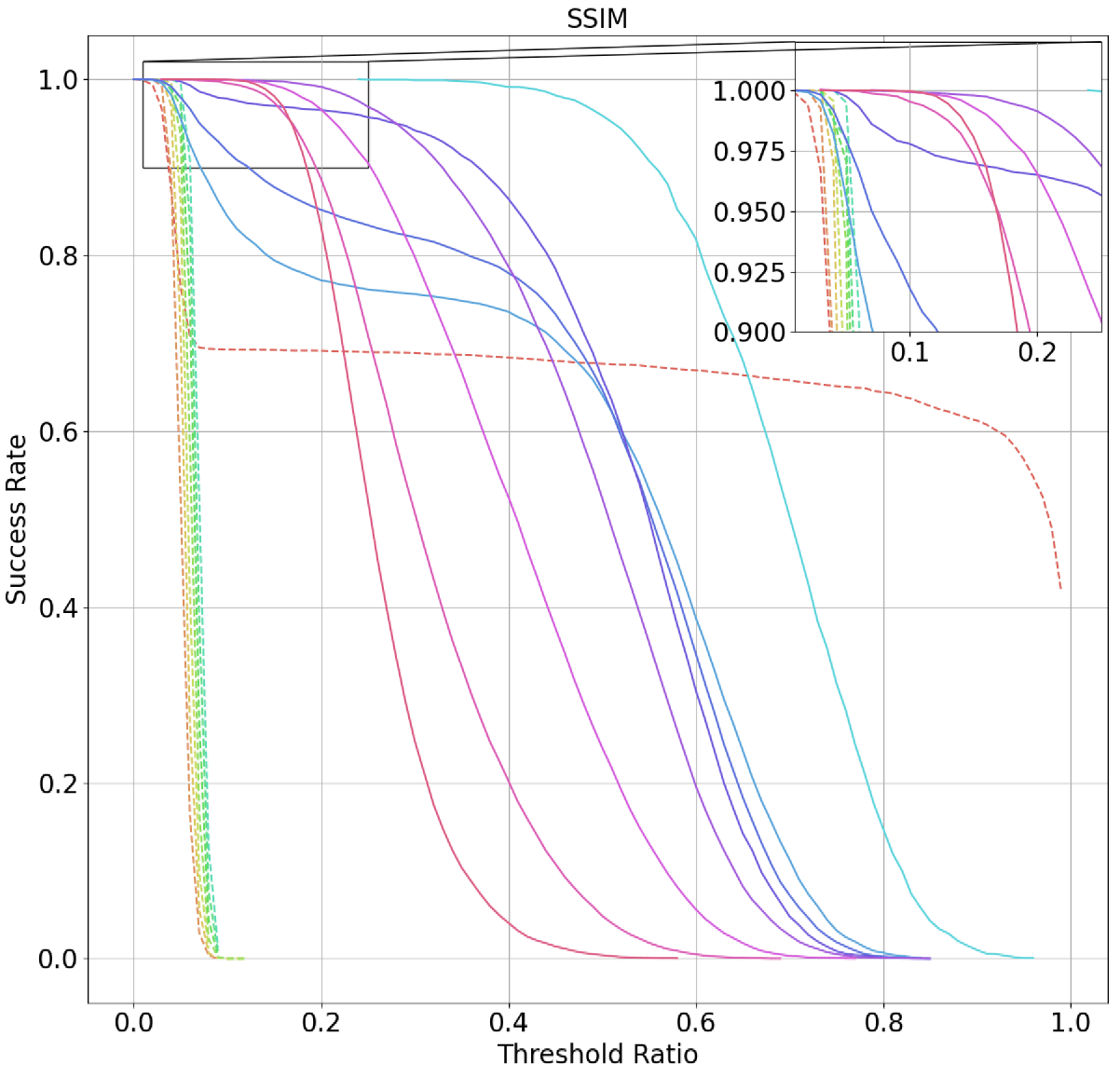}}
\caption{Successful attack rate for different batch size over normalized threshold ratio on \ac{MSE}, \ac{PSNR} and \ac{SSIM} similarity metrics. The dash line refers to the results from \ac{DLG}, and the solid line is from the proposed \ac{GRNN}}
\label{fig:AttachRate}
\end{figure}

To quantitatively evaluate the success rate of leakage attack, we conducted a comparison experiment on the testing set of CIFAR-100 using both \ac{GRNN} and \ac{DLG}. When the number of batch size is larger than 1, we treat the matching problems between ground-truth images and leaked images as a classic assignment problem given the similarity metrics, such as \ac{MSE} and \ac{PSNR}. In addition, \ac{SSIM} was also introduced to evaluate the structural discrepancy between two images. Fig.~\ref{fig:AttachRate} shows the success rates of both methods with different numbers of batch size against the similarity threshold for so-called ``successful attack''. The lower \ac{MSE} score indicates the better match, whereas, the higher \ac{PSNR} and \ac{SSIM} imply the better match. Fig.~\ref{fig:AttachRate} shows that given the same success rate, \ac{GRNN} achieves significantly lower \ac{MSE} scores and higher \ac{PSNR} and \ac{SSIM} apart from the batch size of 1. For example, with a success rate of 0.6, the \ac{MSE} threshold is around 0.05 for \ac{GRNN} and 0.17 for \ac{DLG}. Similarly, given the same threshold ratio, our method achieves a higher success rate. For instance, when a \ac{SSIM} threshold of 0.2 is used, most success rates of \ac{DLG} are dropped down to 0 apart from \ac{DLG} with a batch size of 1 that achieves around 0.7. However, the worst success rate for \ac{GRNN} is above 0.77. This experiment can further approve that the proposed method outperforms \ac{DLG} by a significant margin. More qualitative comparisons that were randomly selected from the generated images are illustrated in Table~\ref{tab:SelectedSamples}.

\subsubsection{Ablation Study on Image Resolution}

As aforementioned in Section~\ref{sec:grnn}, \ac{GRNN} is capable of handling different resolutions of images, due to the flexible number of upsampling blocks. We evaluate the recovery performance of larger resolutions than 32*32 using \emph{ResNet-18} as the local model and the dataset is CIFAR-100. Table~\ref{tab:largeImagePerformance} shows the performance with different image resolutions and batch sizes. First, we upsample the original image from 32 to 64, the results show that even with the batch size of 8, \ac{GRNN} is capable of recovering images from the shared gradient. Then we further explore the resolutions of 128*128 and 256*256, both experiments success with batch size of 1. Some qualitative results in resolutions are showed in Table~\ref{tab:largeImages}. Table~\ref{tab:InvertingGradients} shows the comparison results of \ac{GRNN} and \ac{IG} using a resolution of 256*256, the similarity between the recovered images and original images calculated with \ac{MSE}, \ac{PSNR} and \ac{SSIM} are also given. \ac{GRNN} outperforms \ac{IG} by a significant margin for all samples when the \ac{SSIM} metric is used. It is noticeable that our method is better than \ac{IG} virtually even though the \ac{PSNR} values of \ac{GRNN} are lower for some cases with the batch size of 4 and 8. Based on this study, we can conclude that our method is also capable of recovering the global structure and color appearance in the image when a large batch is used, while \ac{IG} likely produces virtually unrecognizable images.

\begin{table}
\setlength{\tabcolsep}{3pt}
\begin{center}
\caption{Data leakage attack using \ac{GRNN} with different image resolutions and sizes of training batch for \ac{FL} model, where ``\checkmark'' refers to a success and ``$\times$'' refers to a failure. The network is \emph{ResNet-18}, and the dataset is CIFAR-100.}
\label{tab:largeImagePerformance}
\begin{tabular}{>{\centering\arraybackslash}c|>{\centering\arraybackslash}p{0.2in}|>{\centering\arraybackslash}p{0.2in}|>{\centering\arraybackslash}p{0.2in}|>{\centering\arraybackslash}p{0.2in}|>{\centering\arraybackslash}p{0.2in}}
\hline
\diagbox{\textbf{Resolution}}{\textbf{\#Batch}} & \textbf{1} & \textbf{4} & \textbf{8} & \textbf{16} & \textbf{32}  \\
\hline
32*32 & \checkmark & \checkmark & \checkmark & \checkmark & \checkmark \\
64*64 & \checkmark & \checkmark & \checkmark & $\times$ & $\times$ \\
128*128 & \checkmark & $\times$ & $\times$ & $\times$ & $\times$  \\
256*256 & \checkmark & $\times$ & $\times$ & $\times$ & $\times$   \\
\hline
\end{tabular}
\end{center}
\end{table}

\begin{table}
    \centering
    \caption{Recovered images with different resolutions using \ac{GRNN}. The network is non-converged \emph{ResNet-18}, batch size is 1 and dataset is CIFAR-100.}
    \label{tab:largeImages}
    \begin{tabular}{c|c|c}
        \hline
        \textbf{Resolution} &\textbf{Recovered Images} & \textbf{Ground Truth}\\
        \hline
        32*32 & \makecell*[c]{\includegraphics[width=0.1\linewidth]{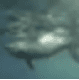}} \makecell*[c]{\includegraphics[width=0.1\linewidth]{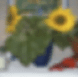}} & \makecell*[c]{\includegraphics[width=0.1\linewidth]{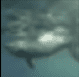}} \makecell*[c]{\includegraphics[width=0.1\linewidth]{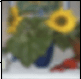}} \\
        \hline
        64*64 & \makecell*[c]{\includegraphics[width=0.1\linewidth]{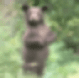}} \makecell*[c]{\includegraphics[width=0.1\linewidth]{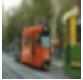}} & \makecell*[c]{\includegraphics[width=0.1\linewidth]{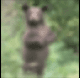}} \makecell*[c]{\includegraphics[width=0.1\linewidth]{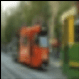}}\\
        \hline
        128*128 & \makecell*[c]{\includegraphics[width=0.1\linewidth]{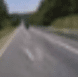}} \makecell*[c]{\includegraphics[width=0.1\linewidth]{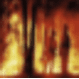}} & \makecell*[c]{\includegraphics[width=0.1\linewidth]{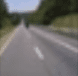}} \makecell*[c]{\includegraphics[width=0.1\linewidth]{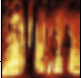}}\\
        \hline
        256*256 & \makecell*[c]{\includegraphics[width=0.1\linewidth]{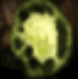}} \makecell*[c]{\includegraphics[width=0.1\linewidth]{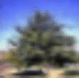}} & \makecell*[c]{\includegraphics[width=0.1\linewidth]{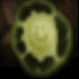}} \makecell*[c]{\includegraphics[width=0.1\linewidth]{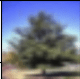}}\\
        \hline
    \end{tabular}
\end{table}

\begin{table}
    \centering
    \setlength{\tabcolsep}{3pt}
    \caption{Randomly selected images recovered by \ac{GRNN} and \ac{IG}. The network is non-converged \emph{ResNet-18}, resolution is 256*256, and dataset is CIFAR-100.}
    \label{tab:InvertingGradients}
    \begin{tabular}{c|c c c c c c c c c}
        \hline
        \multirow{2}{*}{\textbf{Method}} & \multicolumn{9}{c}{\textbf{\#Batch \& Images}} \\
        \cline{2-10}
        & \multicolumn{9}{c}{\textbf{1}} \\
        \hline
        GT  &\makecell*[c]{\includegraphics[width=0.06\linewidth]{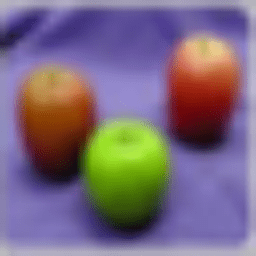}}
            &\makecell*[c]{\includegraphics[width=0.06\linewidth]{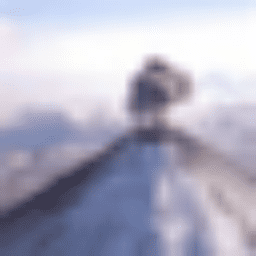}} 
            &\makecell*[c]{\includegraphics[width=0.06\linewidth]{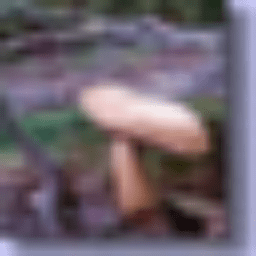}}
            &\makecell*[c]{\includegraphics[width=0.06\linewidth]{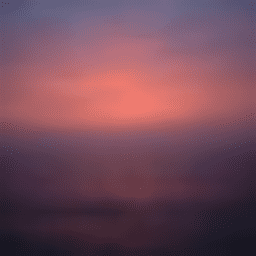}} 
            & \bigg| 
            &\makecell*[c]{\includegraphics[width=0.06\linewidth]{InvertingGradients/B1_True_1.png}}
            &\makecell*[c]{\includegraphics[width=0.06\linewidth]{InvertingGradients/B1_True_2.png}} 
            &\makecell*[c]{\includegraphics[width=0.06\linewidth]{InvertingGradients/B1_True_3.png}}
            &\makecell*[c]{\includegraphics[width=0.06\linewidth]{InvertingGradients/B1_True_4.png}} \\
        \hline
            & \multicolumn{4}{c}{GRNN} & | & \multicolumn{4}{c}{IG} \\
            & \makecell*[c]{\includegraphics[width=0.06\linewidth]{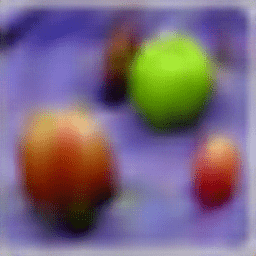}}
            & \makecell*[c]{\includegraphics[width=0.06\linewidth]{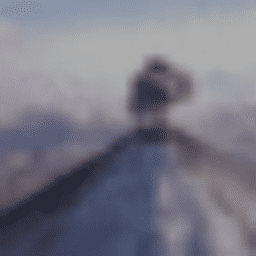}}
            & \makecell*[c]{\includegraphics[width=0.06\linewidth]{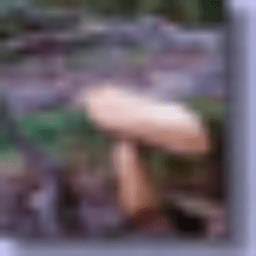}}
            & \makecell*[c]{\includegraphics[width=0.06\linewidth]{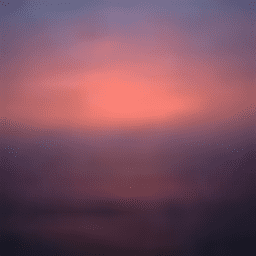}}
            & \bigg| 
            & \makecell*[c]{\includegraphics[width=0.06\linewidth]{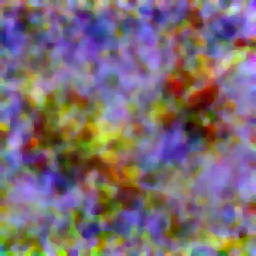}}
            & \makecell*[c]{\includegraphics[width=0.06\linewidth]{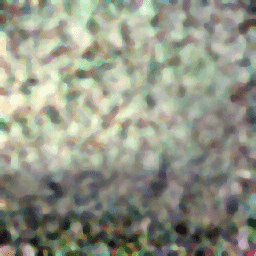}}
            & \makecell*[c]{\includegraphics[width=0.06\linewidth]{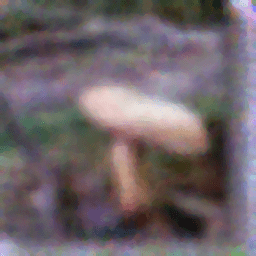}}
            & \makecell*[c]{\includegraphics[width=0.06\linewidth]{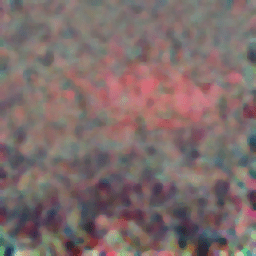}}\\
        MSE     & 4107.90 & \textbf{3656.38} & \textbf{45.75} & \textbf{53.21} 
                & | 
                & \textbf{2272.82} & 4853.13 & 94.19 & 1729.50 
                \\
        PSNR    & \textbf{38.29}   & 37.88   & \textbf{40.43} & \textbf{39.82} 
                & | 
                & 38.09   & \textbf{37.98}   & 39.56 & 38.11 
                \\
        SSIM    & \textbf{0.49}    & \textbf{0.90}    & \textbf{0.95}  & \textbf{0.98}  
                & | 
                & 0.31    & 0.26    & 0.85  & 0.55  
                \\
        \hline
        \hline
        & \multicolumn{9}{c}{\textbf{4}} \\
        \hline
        GT   & \makecell*[c]{\includegraphics[width=0.06\linewidth]{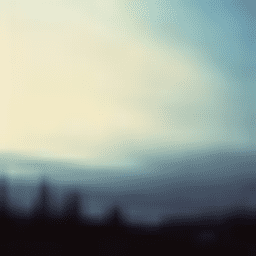}}
             & \makecell*[c]{\includegraphics[width=0.06\linewidth]{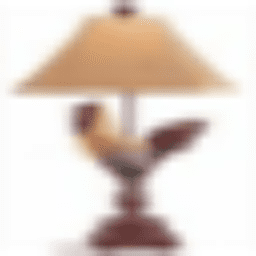}}
             & \makecell*[c]{\includegraphics[width=0.06\linewidth]{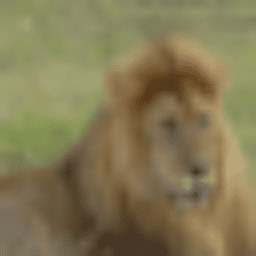}}
             & \makecell*[c]{\includegraphics[width=0.06\linewidth]{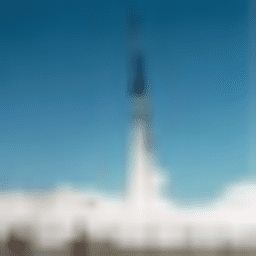}}
             & \bigg| 
             & \makecell*[c]{\includegraphics[width=0.06\linewidth]{InvertingGradients/B4_True_1.png}}
             & \makecell*[c]{\includegraphics[width=0.06\linewidth]{InvertingGradients/B4_True_2.png}}
             & \makecell*[c]{\includegraphics[width=0.06\linewidth]{InvertingGradients/B4_True_3.png}}
             & \makecell*[c]{\includegraphics[width=0.06\linewidth]{InvertingGradients/B4_True_4.png}}\\
        \hline
            & \multicolumn{4}{c}{GRNN} & | & \multicolumn{4}{c}{IG} \\
            & \makecell*[c]{\includegraphics[width=0.06\linewidth]{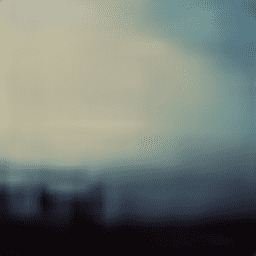}}
            & \makecell*[c]{\includegraphics[width=0.06\linewidth]{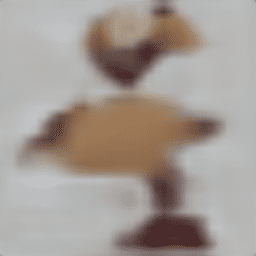}}
            & \makecell*[c]{\includegraphics[width=0.06\linewidth]{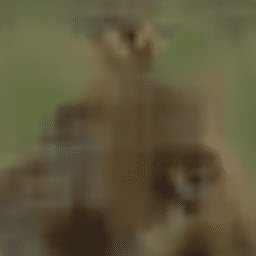}}
            & \makecell*[c]{\includegraphics[width=0.06\linewidth]{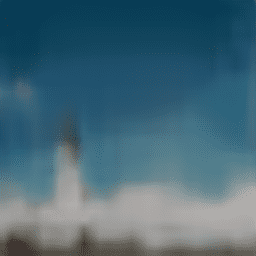}}
            & \bigg|
            & \makecell*[c]{\includegraphics[width=0.06\linewidth]{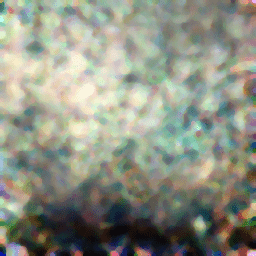}}
            & \makecell*[c]{\includegraphics[width=0.06\linewidth]{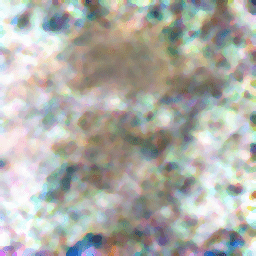}}
            & \makecell*[c]{\includegraphics[width=0.06\linewidth]{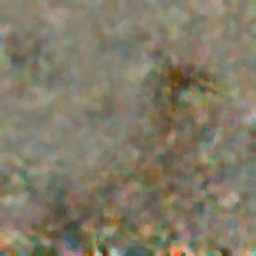}}
            & \makecell*[c]{\includegraphics[width=0.06\linewidth]{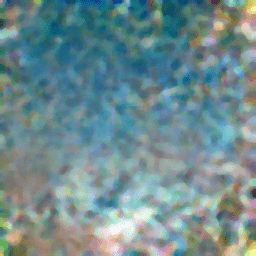}}\\
        MSE   & \textbf{1482.89} & 5996.54 & 1391.17 & \textbf{1799.29} 
             & | 
             & 2261.19 & \textbf{4481.52} & \textbf{716.74} & 2147.60 
             \\
        PSNR    & 37.97   & 38.03   & 37.83   & 37.92   
                & | 
                & \textbf{38.07}   & \textbf{38.05}   & \textbf{38.27}  & \textbf{38.12}   
                \\
        SSIM    & \textbf{0.90}    & \textbf{0.61}    & \textbf{0.75}    & \textbf{0.86}    
                & | 
                & 0.39    & 0.34    & 0.56   & 0.40    
                \\
        \hline
        \hline
        & \multicolumn{9}{c}{\textbf{8}} \\
        \hline
        GT   & \makecell*[c]{\includegraphics[width=0.06\linewidth]{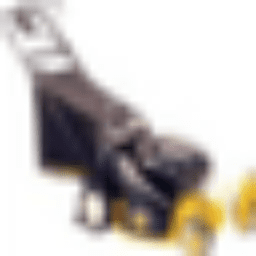}}
             & \makecell*[c]{\includegraphics[width=0.06\linewidth]{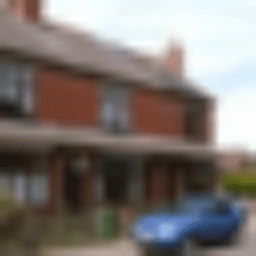}}
             & \makecell*[c]{\includegraphics[width=0.06\linewidth]{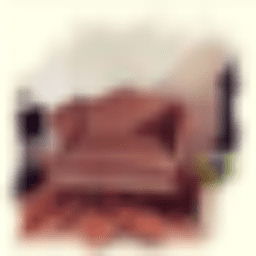}}
             & \makecell*[c]{\includegraphics[width=0.06\linewidth]{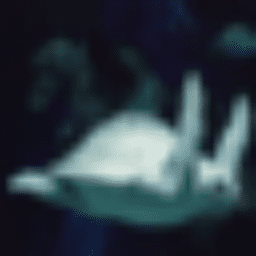}}
             & \bigg| 
             & \makecell*[c]{\includegraphics[width=0.06\linewidth]{InvertingGradients/B8_True_1.png}}
             & \makecell*[c]{\includegraphics[width=0.06\linewidth]{InvertingGradients/B8_True_2.png}}
             & \makecell*[c]{\includegraphics[width=0.06\linewidth]{InvertingGradients/B8_True_3.png}}
             & \makecell*[c]{\includegraphics[width=0.06\linewidth]{InvertingGradients/B8_True_4.png}}\\
        \hline
            & \multicolumn{4}{c}{GRNN} & | & \multicolumn{4}{c}{IG} \\
            & \makecell*[c]{\includegraphics[width=0.06\linewidth]{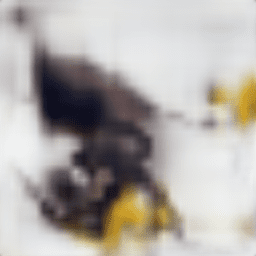}}
            & \makecell*[c]{\includegraphics[width=0.06\linewidth]{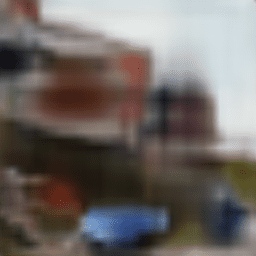}}
            & \makecell*[c]{\includegraphics[width=0.06\linewidth]{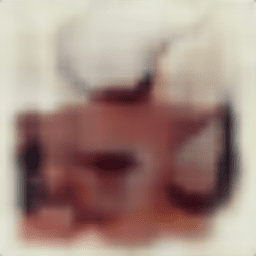}}
            & \makecell*[c]{\includegraphics[width=0.06\linewidth]{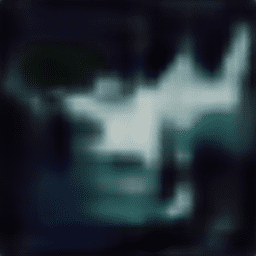}}
            & \bigg| 
            & \makecell*[c]{\includegraphics[width=0.06\linewidth]{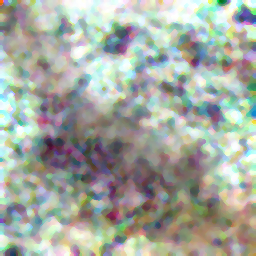}}
            & \makecell*[c]{\includegraphics[width=0.06\linewidth]{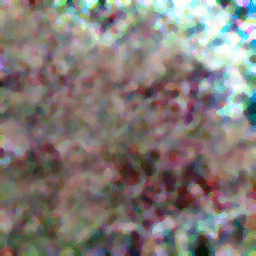}}
            & \makecell*[c]{\includegraphics[width=0.06\linewidth]{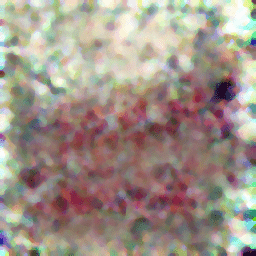}}
            & \makecell*[c]{\includegraphics[width=0.06\linewidth]{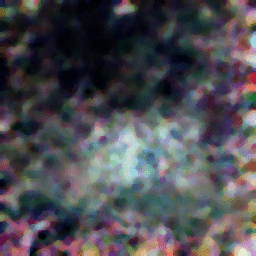}}\\
        MSE   & 4874.68 & \textbf{1280.33} & \textbf{1441.29} & 3380.90 
        & | 
        & \textbf{4836.14} & 3309.53 & 2693.98 & \textbf{3059.81} 
        \\
        PSNR  & 37.94   & \textbf{38.14}   & \textbf{38.09}   & \textbf{38.67}   
        & | 
        & \textbf{38.01}   & 38.08   & 38.06   & 38.11   
        \\
        SSIM  & \textbf{0.54}    & \textbf{0.59}    & \textbf{0.70}    & \textbf{0.54}    
        & | 
        & 0.24    & 0.28    & 0.35    & 0.33    
        \\
        \hline
    \end{tabular}
\end{table}

\subsubsection{Ablation Study on Loss Function}

\ac{MSE} loss is widely used in regression tasks, however, it can be easily biased to the outlier or noisy data point at a pixel-wise level. We believe that the distribution information embedded in the gradient vectors indicates the global structure of the image data. Therefore, we introduced \ac{WD} distance to measure the geometric discrepancy between the fake gradient and true gradient and guide the image generation process. In addition, we carried out comparison experiments to evaluate different combinations of loss functions including \ac{MSE}, \ac{WD}, \ac{TVLoss} and \ac{CD}. The results can be found in Table \ref{tab:lossCompare}. The experimental results show that the proposed loss objective combining \ac{MSE}, \ac{WD} and \ac{TVLoss} achieves the best performance. It is noteworthy mentioning that the color distortion and artifact can be suppressed by using \ac{TVLoss} and \ac{WD} jointly as they can effectively penalize the spurious noises locally and globally.

\begin{table}
    \centering
    \caption{Comparison of image recovery using different loss functions on MNIST dataset.}
    \label{tab:lossCompare}
    \begin{tabular}{c|c|c}
        \hline
        \textbf{Loss Function} &\textbf{Recovered Images} & \textbf{Ground Truth}\\
        \hline
        \ac{MSE} &  \makecell*[c]{\includegraphics[width=0.1\linewidth]{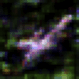}} \makecell*[c]{\includegraphics[width=0.1\linewidth]{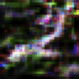}} & \makecell*[c]{\includegraphics[width=0.1\linewidth]{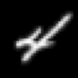}} \makecell*[c]{\includegraphics[width=0.1\linewidth]{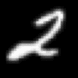}} \\
        \hline
        \ac{WD} &  \makecell*[c]{\includegraphics[width=0.1\linewidth]{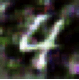}} \makecell*[c]{\includegraphics[width=0.1\linewidth]{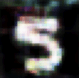}} & \makecell*[c]{\includegraphics[width=0.1\linewidth]{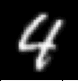}} \makecell*[c]{\includegraphics[width=0.1\linewidth]{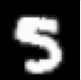}} \\
        \hline
        \ac{CD} &  \makecell*[c]{\includegraphics[width=0.1\linewidth]{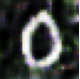}} \makecell*[c]{\includegraphics[width=0.1\linewidth]{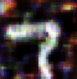}} & \makecell*[c]{\includegraphics[width=0.1\linewidth]{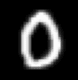}} \makecell*[c]{\includegraphics[width=0.1\linewidth]{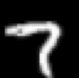}} \\
        \hline
        \ac{MSE} \& \ac{WD} & \makecell*[c]{\includegraphics[width=0.1\linewidth]{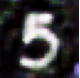}} \makecell*[c]{\includegraphics[width=0.1\linewidth]{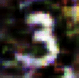}} & \makecell*[c]{\includegraphics[width=0.1\linewidth]{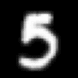}} \makecell*[c]{\includegraphics[width=0.1\linewidth]{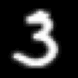}}\\
        \hline
        \ac{MSE} \& \ac{CD} & \makecell*[c]{\includegraphics[width=0.1\linewidth]{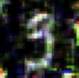}} \makecell*[c]{\includegraphics[width=0.1\linewidth]{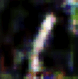}} & \makecell*[c]{\includegraphics[width=0.1\linewidth]{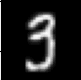}} \makecell*[c]{\includegraphics[width=0.1\linewidth]{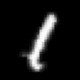}}\\
        \hline
        \ac{MSE} \& \ac{CD} \& \ac{TVLoss} &\makecell*[c]{\includegraphics[width=0.1\linewidth]{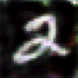}} \makecell*[c]{\includegraphics[width=0.1\linewidth]{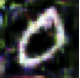}} & \makecell*[c]{\includegraphics[width=0.1\linewidth]{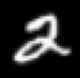}} \makecell*[c]{\includegraphics[width=0.1\linewidth]{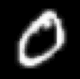}}\\
        \hline
        \ac{MSE} \& \ac{WD} \& \ac{TVLoss} &\makecell*[c]{\includegraphics[width=0.1\linewidth]{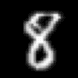}} \makecell*[c]{\includegraphics[width=0.1\linewidth]{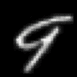}} & \makecell*[c]{\includegraphics[width=0.1\linewidth]{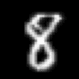}} \makecell*[c]{\includegraphics[width=0.1\linewidth]{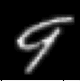}}\\
        \hline
    \end{tabular}
\end{table}

\subsection{Label Inference}
\label{sec:dli}
In addition to image data recovery, we also investigated the performance of label inference in those experiments. Table~\ref{tab:lia} provides the comparative results of label inference accuracy (\%) for \ac{DLG} and \ac{GRNN}, where each experiment was repeated 10 times, the mean and standard deviation were reported. We can observe that the label inference accuracy decreases while the size of the training batch increases for both \ac{DLG} and \ac{GRNN}. However, our method outperforms \ac{DLG} in all experiments except the one on MNIST with a batch size of 8. Furthermore, \ac{GRNN} is significantly better than \ac{DLG} when a large batch size is used. For example, \ac{GRNN} achieves 99.84\% on \ac{LFW} using \emph{LeNet} and a batch size of 256, whereas \ac{DLG} only obtains 79.69\% using the same setting. Note that having a correct label prediction does not necessarily indicate the corresponding image data can be recovered successfully (see Table \ref{tab:AttackPerformance}). We can conclude that recovering image information is much more challenging than recovering labels as the distribution of image data is in a much higher dimension than its corresponding label. The accuracy of label inference on \emph{ResNet-18} achieves 100\% in almost all experiments except the one on \ac{LFW} with a batch size of 32 (94.06\%), which is higher and more stable than \emph{LeNet}. \emph{ResNet-18} has much more trainable parameters than \emph{LeNet}, therefore, the gradient with a larger number of elements is much more informative for finding decision boundaries for the classification task.

\begin{table}
\setlength{\tabcolsep}{2.8pt}
\begin{center}
\caption{Comparison of label inference accuracy (\%) using \ac{DLG} and \ac{GRNN}, where L. and R. refer to LeNet and ResNet-18 respectively.}
\label{tab:lia}
\begin{tabular}{c|c|c|c|c|c|c|c|c|c|c||c}
\hline
 \multicolumn{3}{c|}{\textbf{\#Batch}} & \textbf{1} & \textbf{4} & \textbf{8} & \textbf{16} & \textbf{32} & \textbf{64} & \textbf{128} & \textbf{256} & \textbf{Avg} \\
\hline
\multirow{6}{*}{\ac{DLG}}
 & \multirow{6}{*}{L.}
 & \multirow{2}{*}{MNIST} & \textbf{100.0$\pm$} & 97.50$\pm$ & \textbf{100.0$\pm$} & 94.38$\pm$ & 90.62$\pm$ & 90.31$\pm$ & 91.41$\pm$ & 92.42$\pm$ & 94.23$\pm$ \\
&&                        & \textbf{0.0} & 0.0750 & \textbf{0.0} & 0.0337 & 0.0242 & 0.0260 & 0.0152 & 0.0181 & 0.0254 \\
\cline{3-12}
&& \multirow{2}{*}{CIFAR} & \textbf{100.0$\pm$} & 97.50$\pm$ & 98.75$\pm$ & 97.50$\pm$ & 92.50$\pm$ & 89.84$\pm$ & 86.33$\pm$ & 85.70$\pm$ & 93.51$\pm$ \\
&&                            & \textbf{0.0} & 0.0750 & 0.0375 & 0.0306 & 0.0287 & 0.0329 & 0.0295 & 0.0289 & 0.0329 \\
\cline{3-12}
&& \multirow{2}{*}{LFW} & \textbf{100.0$\pm$}  & 80.00$\pm$ & 90.00$\pm$ & 85.00$\pm$ & 88.44$\pm$ & 70.62$\pm$ & 89.53$\pm$ & 79.69$\pm$  & 85.41$\pm$ \\
&&                      & \textbf{0.0}  & 0.3317 & 0.1561 & 0.2358 & 0.1683 & 0.3924 & 0.2985 & 0.3985 & 0.2477 \\
\hline
\hline

\multirow{12}{*}{Ours}
 & \multirow{6}{*}{L.}
 & \multirow{2}{*}{MNIST}     & \textbf{100.0$\pm$} & \textbf{100.0$\pm$} & 97.50$\pm$ & 97.50$\pm$ & 97.50$\pm$ & \textbf{96.25$\pm$} & \textbf{96.25$\pm$} & \textbf{96.37$\pm$} & 97.73$\pm$ \\
&&                            & \textbf{0.0} & \textbf{0.0} & 0.0500 & 0.0415 & 0.0187 & \textbf{0.0188} & \textbf{0.0171} & \textbf{0.0165} & 0.0203 \\
\cline{3-12}
&& \multirow{2}{*}{CIFAR} & \textbf{100.0$\pm$} & \textbf{100.0$\pm$} & \textbf{100.0$\pm$} & 99.38$\pm$ & 99.69$\pm$ & \textbf{98.91$\pm$} & \textbf{98.75$\pm$} & \textbf{96.80$\pm$} & 99.19$\pm$ \\
&&                            & \textbf{0.0} & \textbf{0.0} & \textbf{0.0} & 0.0188 & 0.0094 & \textbf{0.0122} & \textbf{0.0080} & \textbf{0.0108} & 0.0074 \\
\cline{3-12}
&& \multirow{2}{*}{\ac{LFW}}       & \textbf{100.0$\pm$} & 97.50$\pm$ & 98.75$\pm$ & \textbf{100.0$\pm$} & \textbf{100.0$\pm$} & \textbf{99.69$\pm$} & \textbf{99.69$\pm$} & \textbf{99.84$\pm$} & \textbf{99.43$\pm$} \\
&&                            & \textbf{0.0} & 0.0750 & 0.0375 & \textbf{0.0} & \textbf{0.0} & \textbf{0.0062} & \textbf{0.0071} & \textbf{0.0019} & \textbf{0.0160} \\

\cline{2-12}
 & \multirow{6}{*}{R.}
 & \multirow{2}{*}{MNIST}     & \textbf{100.0$\pm$} & \textbf{100.0$\pm$} & \textbf{100.0$\pm$} & \textbf{100.0$\pm$} & \textbf{100.0$\pm$} & - & - & - & \textbf{100.0$\pm$} \\
&&                            & \textbf{0.0} & \textbf{0.0} & \textbf{0.0} & \textbf{0.0} & \textbf{0.0} & - & - & - & \textbf{0.0} \\
\cline{3-12}
&& \multirow{2}{*}{CIFAR} & \textbf{100.0$\pm$} & \textbf{100.0$\pm$} & \textbf{100.0$\pm$} & \textbf{100.0$\pm$} & \textbf{100.0$\pm$} & - & - & - & \textbf{100.0$\pm$} \\
&&                            & \textbf{0.0} & \textbf{0.0} & \textbf{0.0} & \textbf{0.0} & \textbf{0.0} & - & - & - & \textbf{0.0} \\
\cline{3-12}
&& \multirow{2}{*}{\ac{LFW}}       & \textbf{100.0$\pm$} & \textbf{100.0$\pm$} & \textbf{100.0$\pm$} & \textbf{100.0$\pm$} & 94.06$\pm$ & - & - & - & 98.81$\pm$ \\
&&                            & \textbf{0.0} & \textbf{0.0} & \textbf{0.0} & \textbf{0.0} & 0.0452 & - & - & - & 0.0090 \\
\hline
\end{tabular}
\end{center}
\end{table}

\begin{table}
\setlength{\tabcolsep}{7pt}
\begin{center}
\caption{Performances of different network architectures, where training accuracy refers to predicted results of true images and relevant ground truth label. Re-identification accuracy is from predicted results of fake images and relevant ground truth labels. \ac{DLG} and \ac{GRNN} both use \emph{LeNet} as backbone. Training and testing samples are from VGG-Face dataset. Res18 represents to ResNet-18 and Dense121 is DenseNet-121.}
\label{tab:reid}
\begin{tabular}{c|c|c|c|c|c|c|c}
\hline
\multirow{2}{*}{\textbf{Method}} &\multirow{2}{*}{\textbf{Network}} & \multirow{2}{*}{\textbf{Train Acc}} & \multirow{2}{*}{\textbf{\#B}} &\multicolumn{3}{c|}{\textbf{Re-identification Accuracy}} & \multirow{2}{*}{\textbf{Sample No.}} \\
\cline{5-7}
&&&& \textbf{Top-1} & \textbf{Top-3} & \textbf{Top-5}\\
\hline
\multirow{2}{*}{DLG} & Res18 & 97.27\% & 1 & 25.14\% & 45.57\% & 51.86\% & 700 \\
\cline{2-8}
 & Dense121 & 97.11\% & 1 & \textbf{15.57\%} & 33.57\% & 42.14\% & 700\\
\hline
\hline
\multirow{8}{*}{GRNN}
 & \multirow{4}{*}{Res18} & \multirow{4}{*}{97.27\%} & 1 & \textbf{30.66\%} & \textbf{74.79\%} & \textbf{88.40\%} & 700\\
&&& 4 & 17.45\% & 24.64\% & 31.11\% & 1112\\ 
&&& 8 & 6.14\% & 13.58\% & 22.13\% & 2224\\ 
&&& 16 & 2.90\% & 10.43\% & 22.08\% & 4352\\ 
\cline{2-8}
 & \multirow{4}{*}{Dense121} & \multirow{4}{*}{97.11\%} & 1 & 11.46\% & \textbf{43.12\%} & \textbf{63.03\%} & 700\\ 
&&& 4 & 9.53\% & 19.87\% & 26.80\% & 1112\\  
&&& 8 & 3.06\% & 10.25\% & 19.83\% & 2224\\ 
&&& 16 & 1.52\% & 8.23\% & 19.12\% & 4352\\ 
\hline
\end{tabular}
\end{center}
\end{table}

\begin{table}
\setlength{\tabcolsep}{3pt}
\begin{center}
\caption{The re-identification results of true images and their corresponding generated fake images.}
\label{tab:reid.examples}
\begin{tabular}{c|c|c|c}
\hline
\textbf{Ground Truth Label} & \textbf{\#42} & \textbf{\#26} & \textbf{\#69}\\
\hline
\hline
\textbf{True Input Image} & 
\makecell*[c]{\includegraphics[width=0.1\linewidth]{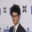}} & \makecell*[c]{\includegraphics[width=0.1\linewidth]{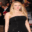}}&
\makecell*[c]{\includegraphics[width=0.1\linewidth]{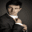}} \\
\hline
\textbf{Top-3 Labels} & 
\textbf{\#42} | \#97 | \#94 & 
\textbf{\#26} | \#22 | \#64 &
\#22 | \#64 | \#23 \\
\hline
\textbf{Top-3 Images} & 
\makecell*[c]{\includegraphics[width=0.2\linewidth]{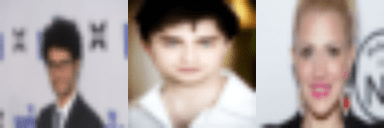}} & \makecell*[c]{\includegraphics[width=0.2\linewidth]{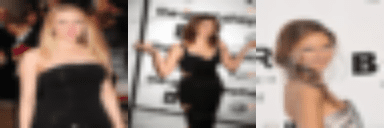}}&
\makecell*[c]{\includegraphics[width=0.2\linewidth]{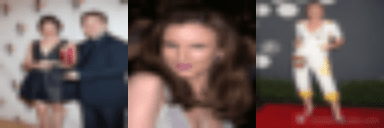}} \\
\hline
\textbf{Top-3 Confidences} & 
40.70\% | 10.84\% | 4.08\% & 
90.34\% | 4.04\% | 1.75\% &
66.10\% | 5.02\% | 4.71\% \\
\hline
\hline
\textbf{Generated Input Image} & 
\makecell*[c]{\includegraphics[width=0.1\linewidth]{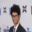}} & \makecell*[c]{\includegraphics[width=0.1\linewidth]{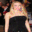}}&
\makecell*[c]{\includegraphics[width=0.1\linewidth]{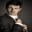}}\\
\hline
\textbf{Top-3 Labels} & 
\textbf{\#42} | \#4 | \#22 & 
\#57 | \#31 | \#75 &
\textbf{\#69} | \#76 | \#88 \\
\hline
\textbf{Top-3 Images} & 
\makecell*[c]{\includegraphics[width=0.2\linewidth]{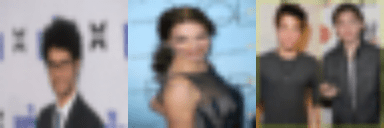}} & \makecell*[c]{\includegraphics[width=0.2\linewidth]{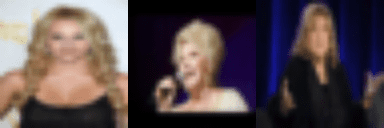}}&
\makecell*[c]{\includegraphics[width=0.2\linewidth]{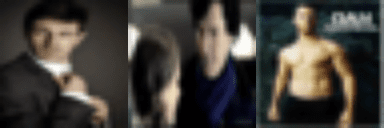}} \\
\hline
\textbf{Top-3 Confidences} & 
11.89\% | 7.58\% | 7.46\% & 
73.10\% | 7.88\% | 3.91\% &
59.52\% | 7.96\% | 3.29\% \\
\hline


\end{tabular}
\end{center}
\end{table}

We also noticed that the number of label classes has an impact on its inference performance. In our experiment, MNIST, CIFAR-100, and \ac{LFW} have 10 classes, 100 classes, and 5749 classes, respectively. The average accuracy of \ac{DLG} is 94.23\% on MNIST while decreasing to 93.51\% and 85.41\% on CIFAR-100 and \ac{LFW}, respectively. In contrast, \ac{GRNN} achieves higher accuracy on \ac{LFW} compared to the other two datasets. In \ac{DLG}, the label is obtained via updating the image input and label input jointly during the backward pass phase in \ac{SGD}, whereas in \ac{GRNN}, the label is calculated by the fake label generator in the forward pass phase. We believe that by adding the image data and label generators, \ac{GRNN} can better capture the correspondence between image data and its label in the joint latent space and, furthermore, can generate more individualized images with respect to different classes.


\subsection{Face Re-Identification}
As shown in Table \ref{tab:ExamplesComparison}, recovered images look almost the same as their corresponding true images, there are still slight deviations that may produce classification misleading results if we use generated data to replace the original ones. This behavior of \ac{DL} methods are well documented in the literature, \emph{e.g.} \cite{moosavi2016deepfool,modas2019sparsefool,shamsabadi2020colorfool}. Taking face recognition as an example, the recovered face image may look identical to its original image visually, however, it cannot produce correct prediction by the face recognition model to identify the person, see Table~\ref{tab:reid.examples}. Therefore, we applied the face re-identification experiment to evaluate the feasibility of data leakage attack using \ac{GRNN}. We first used the \ac{GRNN} to recover the face image data from the \ac{FL} system during the training stage. Then, the fake image was passed to the face recognition model as input to predict the identity label. The success of re-identification was counted if the prediction label matches the true label. VGG-Face dataset was used in this experiment which contains 2622 identities. We used the top 100 identities that have the most image samples for training. We selected the successfully recovered images data from the output of \ac{GRNN} which ended up with 700 fake face images in total when batch size is 1, and 1112 images, 2224 images, and 4352 images for batch size 4, 8, and 16 respectively. In the meantime, we trained two face recognition models using \emph{ResNet-18} and \emph{DenseNet-121} using the same training set. Table \ref{tab:reid} reports the top-1, top-3 and top-5 accuracies of re-identification of those fake face images. Although the top-1 accuracy on \emph{ResNet-18} and \emph{DenseNet-121} are 30.66\% and 11.46\% when batch size is 1, they are significantly better than random prediction (1\%). The re-identification accuracy increases dramatically when we consider using top-3 and top-5 metrics. The recovery performance becomes worse with the increasing batch size, so the accuracy decreases as well. We say that the face recognition deep models are sensitive to the minor perturbation that is produced during the image recovery process. The images generated using \ac{GRNN} achieve significantly higher accuracies compared to \ac{DLG}, \emph{i.e.} $+5.52\%$, $+29.22\%$ and $+36.54\%$ can be achieved in \emph{Top-1}, \emph{Top-3} and \emph{Top-5} accuracies using \emph{ResNet-18}.

\subsection{Defense Strategy}
\label{sec:ds}

\begin{table}[t]
\setlength{\tabcolsep}{5pt}
\begin{center}
\caption{Average \ac{PSNR} scores with different noise types and scales. ``$\times$'' means the method fails the experiment, whereas \ac{PSNR} score is given only if it successes. \ac{DLG} failed completely, as it had no visible success among all the experiments.}
\label{tab:DPresults}
\begin{tabular}{c|c|c|c|c|c|c|c}
\hline
\textbf{Method} & \textbf{Dataset} & \diagbox{\textbf{Type}}{\textbf{Scale}} & \textbf{\#Batch} & \textbf{1e-1} & \textbf{1e-2} & \textbf{1e-3} & \textbf{1e-4}  \\
\hline
\multirow{12}{*}{DLG} &
\multirow{4}{*}{MNIST}
& \multirow{2}{*}{Gaussian}
  & 1 & $\times$ & $\times$ & $\times$ & $\times$ \\
&&& 4 & $\times$ & $\times$ & $\times$ & $\times$ \\
\cline{3-8}
&& \multirow{2}{*}{Laplacian}
  & 1 & $\times$ & $\times$ & $\times$ & $\times$ \\
&&& 4 & $\times$ & $\times$ & $\times$ & $\times$ \\
\cline{2-8}
&\multirow{4}{*}{CIFAR-100}
& \multirow{2}{*}{Gaussian}
  & 1 & $\times$ & $\times$ & $\times$ & $\times$ \\
&&& 4 & $\times$ & $\times$ & $\times$ & $\times$ \\
\cline{3-8}
&& \multirow{2}{*}{Laplacian}
  & 1 & $\times$ & $\times$ & $\times$ & $\times$ \\
&&& 4 & $\times$ & $\times$ & $\times$ & $\times$ \\
\cline{2-8}
&\multirow{4}{*}{LFW}
& \multirow{2}{*}{Gaussian}
  & 1 & $\times$ & $\times$ & $\times$ & $\times$ \\
&&& 4 & $\times$ & $\times$ & $\times$ & $\times$ \\
\cline{3-8}
&& \multirow{2}{*}{Laplacian}
  & 1 & $\times$ & $\times$ & $\times$ & $\times$ \\
&&& 4 & $\times$ & $\times$ & $\times$ & $\times$ \\
\hline
\hline
\multirow{24}{*}{Ours} &
\multirow{8}{*}{MNIST}
& \multirow{4}{*}{Gaussian}
  & 1 & 39.73 & 39.73 & 47.20 & 52.51 \\
&&& 4 & $\times$ & 39.66 & 39.67 & 44.43 \\
&&& 8 & $\times$ & 40.11 & 39.65 & 40.71 \\
&&&16 & $\times$ & $\times$ & 39.73 & 39.78 \\
\cline{3-8}
&& \multirow{4}{*}{Laplacian}
  & 1 & 39.69 & 39.55 & 45.08 & 51.80 \\
&&& 4 & $\times$ & 39.67 & 39.83 & 45.39 \\
&&& 8 & $\times$ & $\times$ & 39.66 & 41.02 \\
&&&16 & $\times$ & $\times$ & 39.62 & 40.00 \\
\cline{2-8}
&\multirow{8}{*}{CIFAR-100} 
& \multirow{4}{*}{Gaussian} 
  & 1 & $\times$ & 38.18 & 41.78 & 45.75 \\
&&& 4 & $\times$ & $\times$ & 38.29 & 40.64 \\
&&& 8 & $\times$ & $\times$ & 38.18 & 39.26 \\
&&&16 & $\times$ & $\times$ & $\times$ & 38.65 \\
\cline{3-8}
&& \multirow{4}{*}{Laplacian} 
  & 1 & $\times$ & 38.13 & 40.17 & 46.20 \\
&&& 4 & $\times$ & $\times$ & 38.14 & 40.32 \\
&&& 8 & $\times$ & $\times$ & 38.08 & 39.14 \\
&&&16 & $\times$ & $\times$ & $\times$ & $\times$ \\
\cline{2-8}
&\multirow{8}{*}{LFW} 
& \multirow{4}{*}{Gaussian} 
  & 1 & $\times$ & 38.15 & 41.72 & 46.26 \\
&&& 4 & $\times$ & $\times$ & 38.25 & 40.59  \\
&&& 8 & $\times$ & $\times$ & $\times$ & $\times$ \\
&&&16 & $\times$ & $\times$ & $\times$ & $\times$ \\
\cline{3-8}
&& \multirow{4}{*}{Laplacian} 
  & 1 & $\times$ & 38.09 & 40.96 & 46.09 \\
&&& 4 & $\times$ & $\times$ & 38.19 & 40.26 \\
&&& 8 & $\times$ & $\times$ & $\times$ & $\times$ \\
&&&16 & $\times$ & $\times$ & $\times$ & $\times$ \\
\hline
\end{tabular}
\end{center}
\end{table}

\begin{figure}[ht]
\centering
    \subfigure{\includegraphics[width=0.39\linewidth]{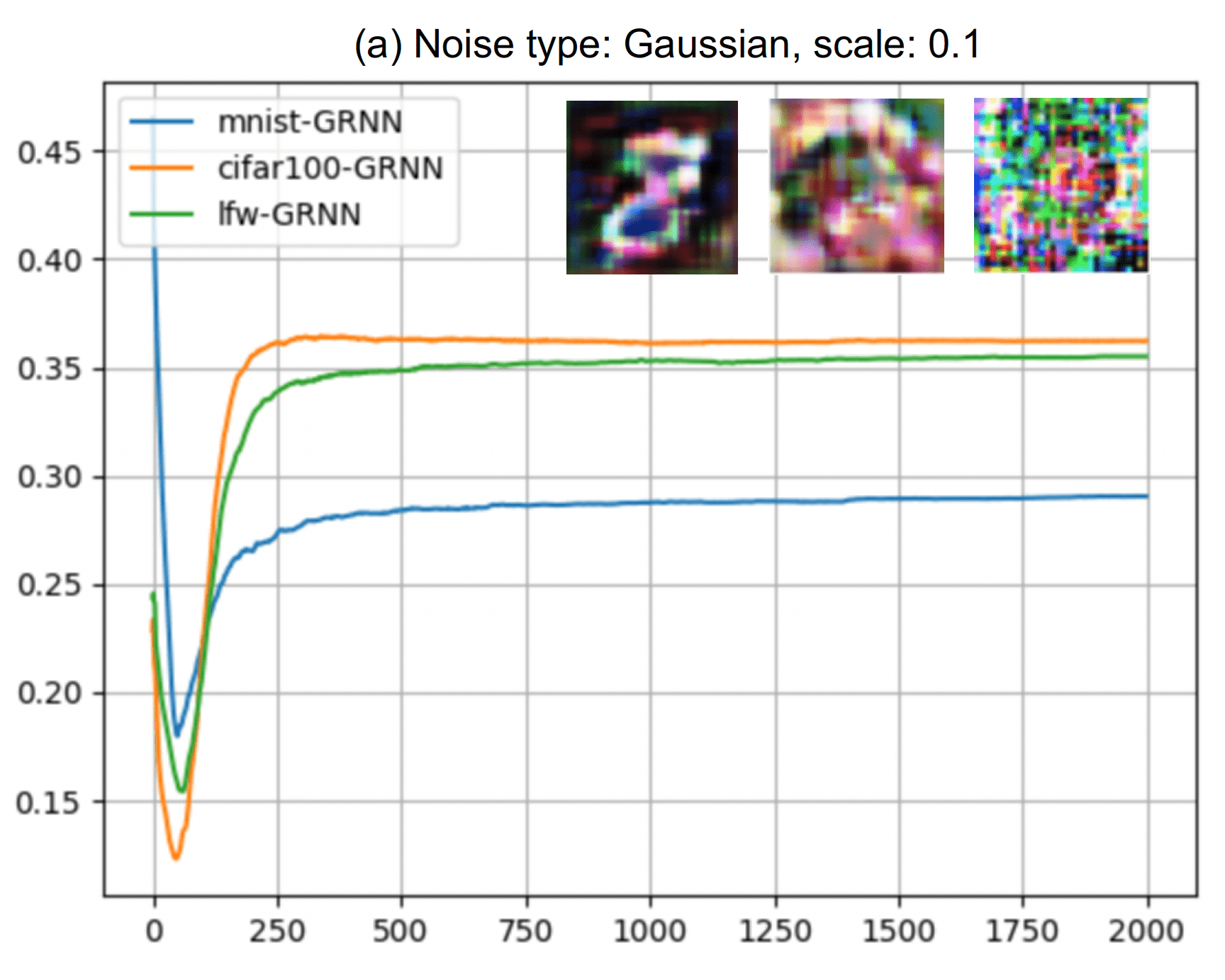}}
    \hspace{.5in}
    \subfigure{\includegraphics[width=0.39\linewidth]{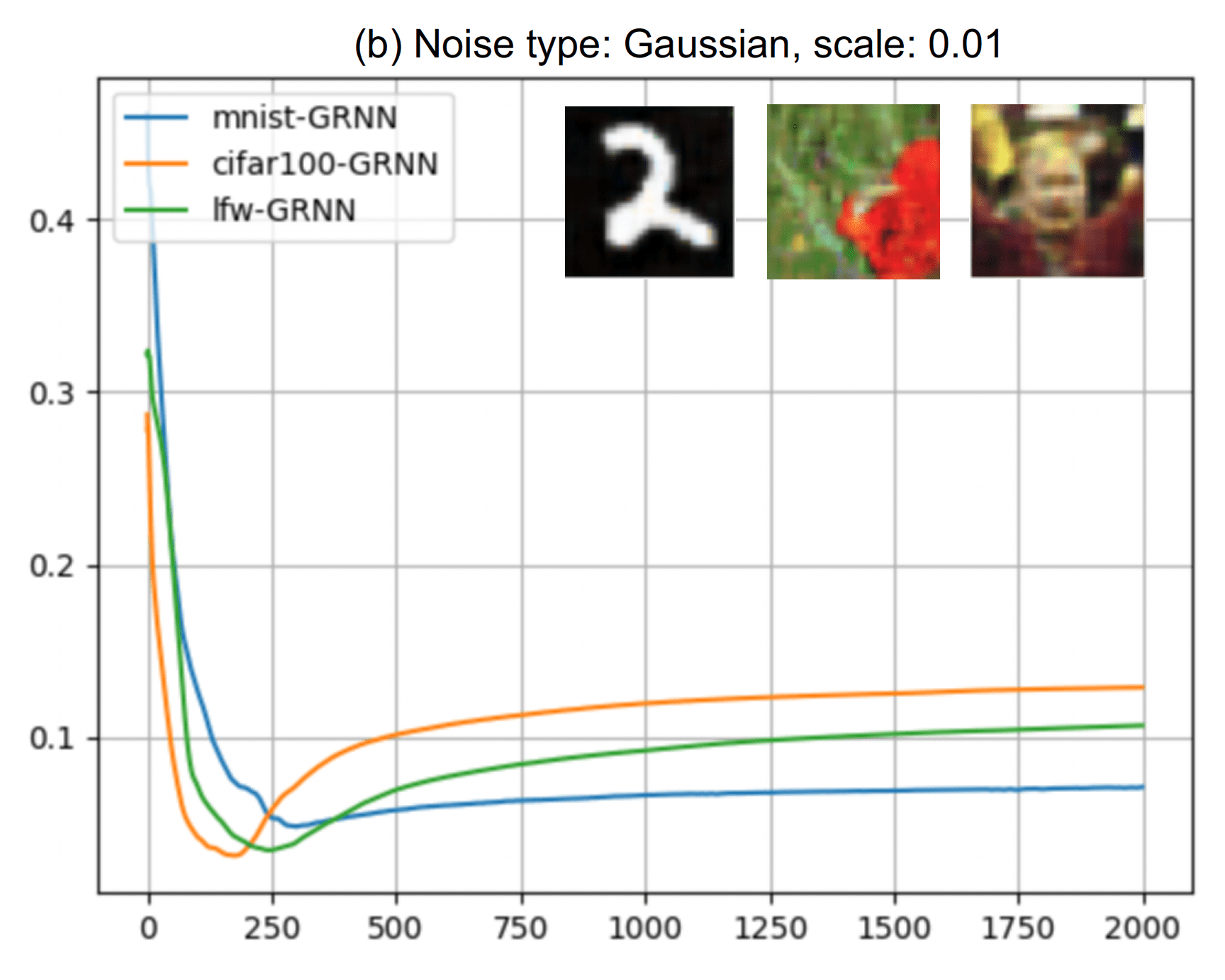}} \\
    \vspace{-.2in}
    \subfigure{\includegraphics[width=0.39\linewidth]{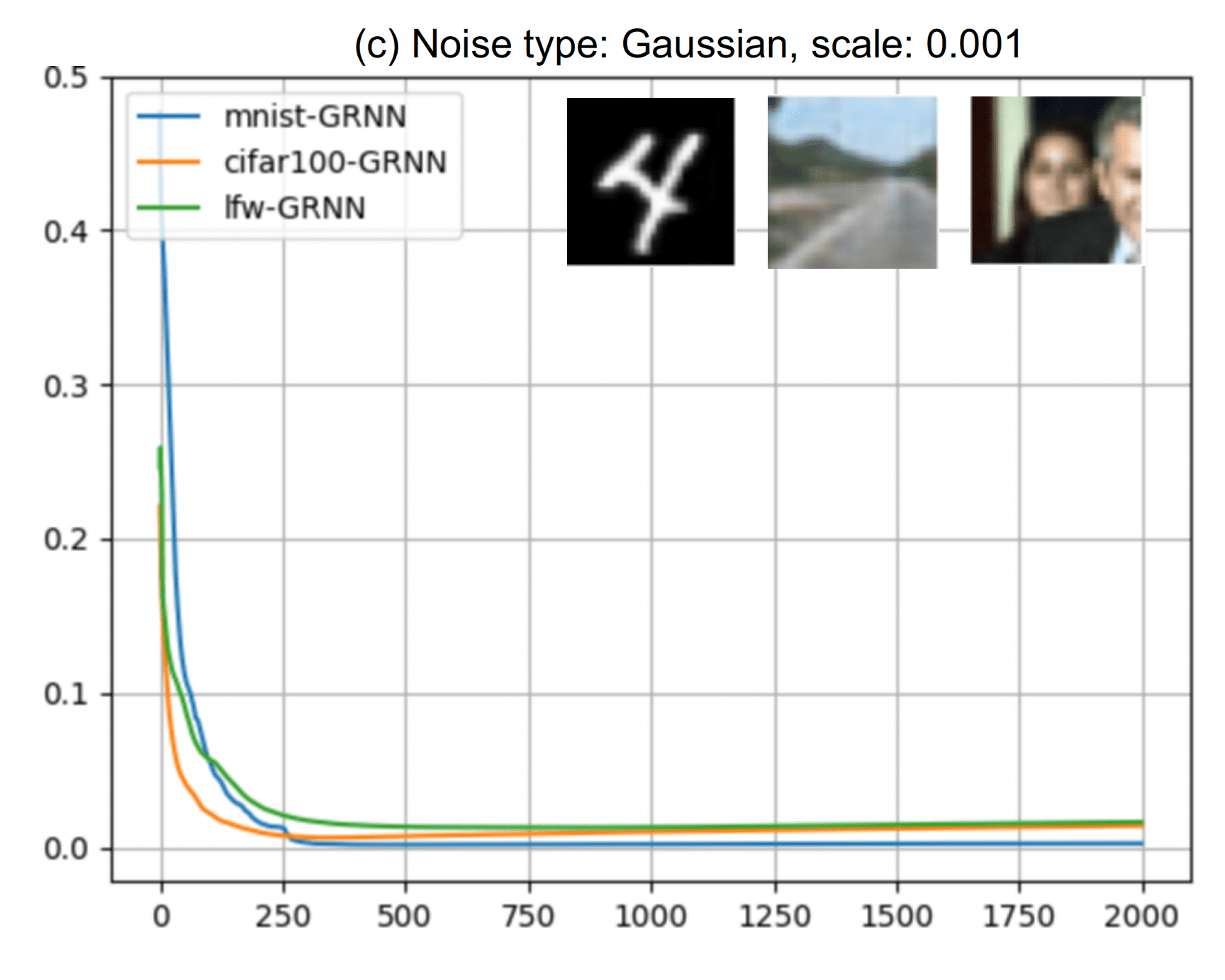}}
    \hspace{.5in}
    \subfigure{\includegraphics[width=0.39\linewidth]{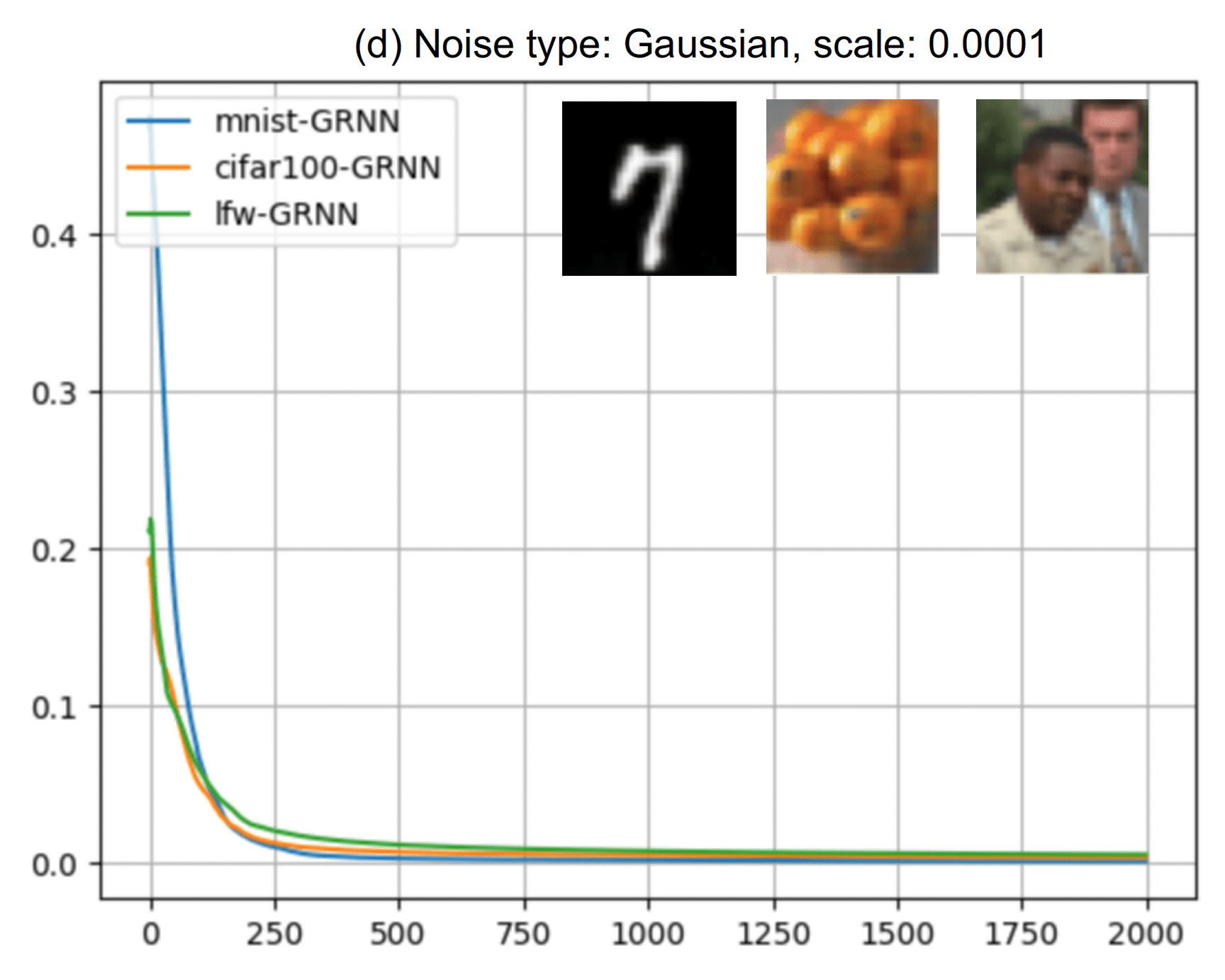}} \\
    \vspace{-.2in}
    \subfigure{\includegraphics[width=0.39\linewidth]{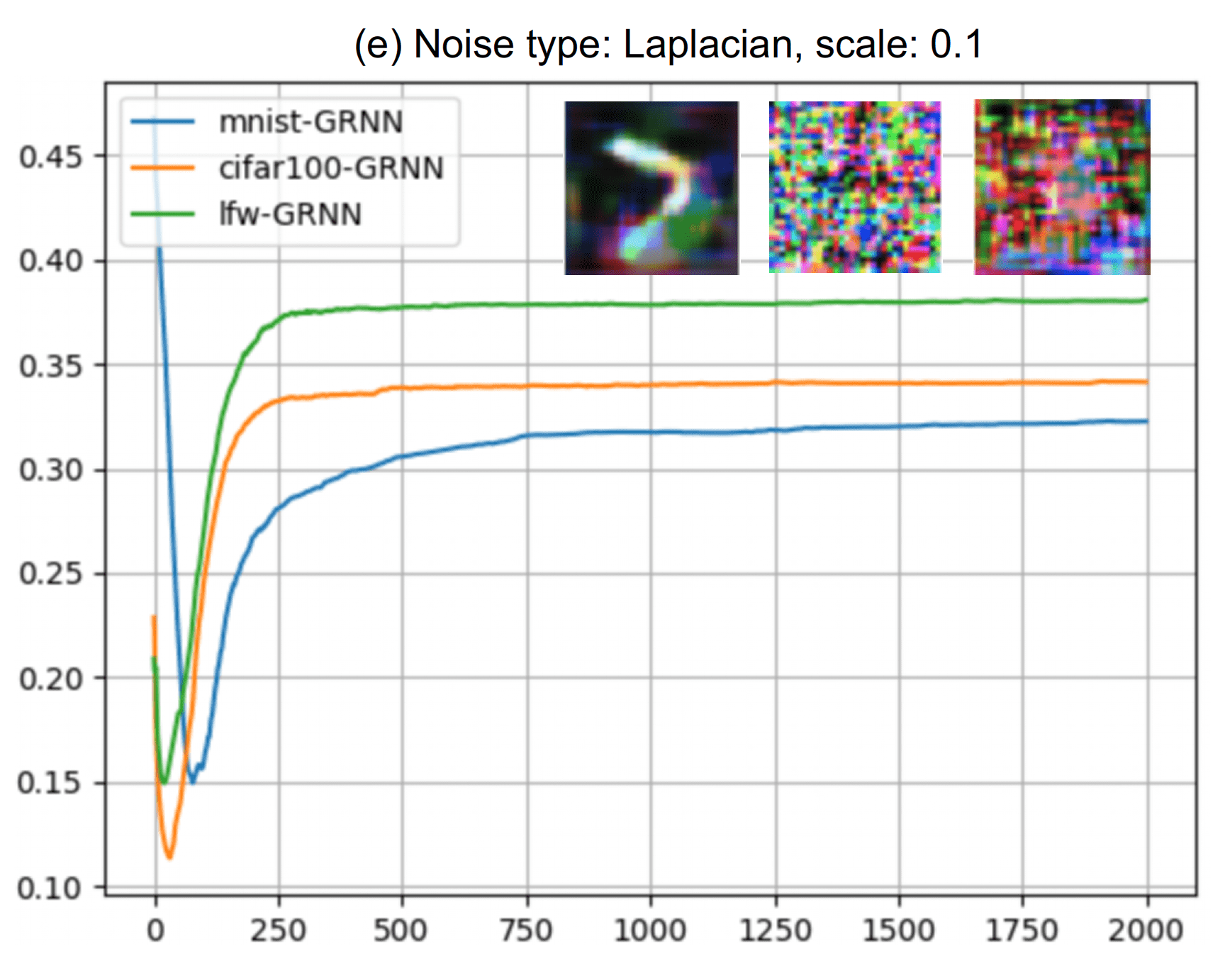}}
    \hspace{.5in}
    \subfigure{\includegraphics[width=0.39\linewidth]{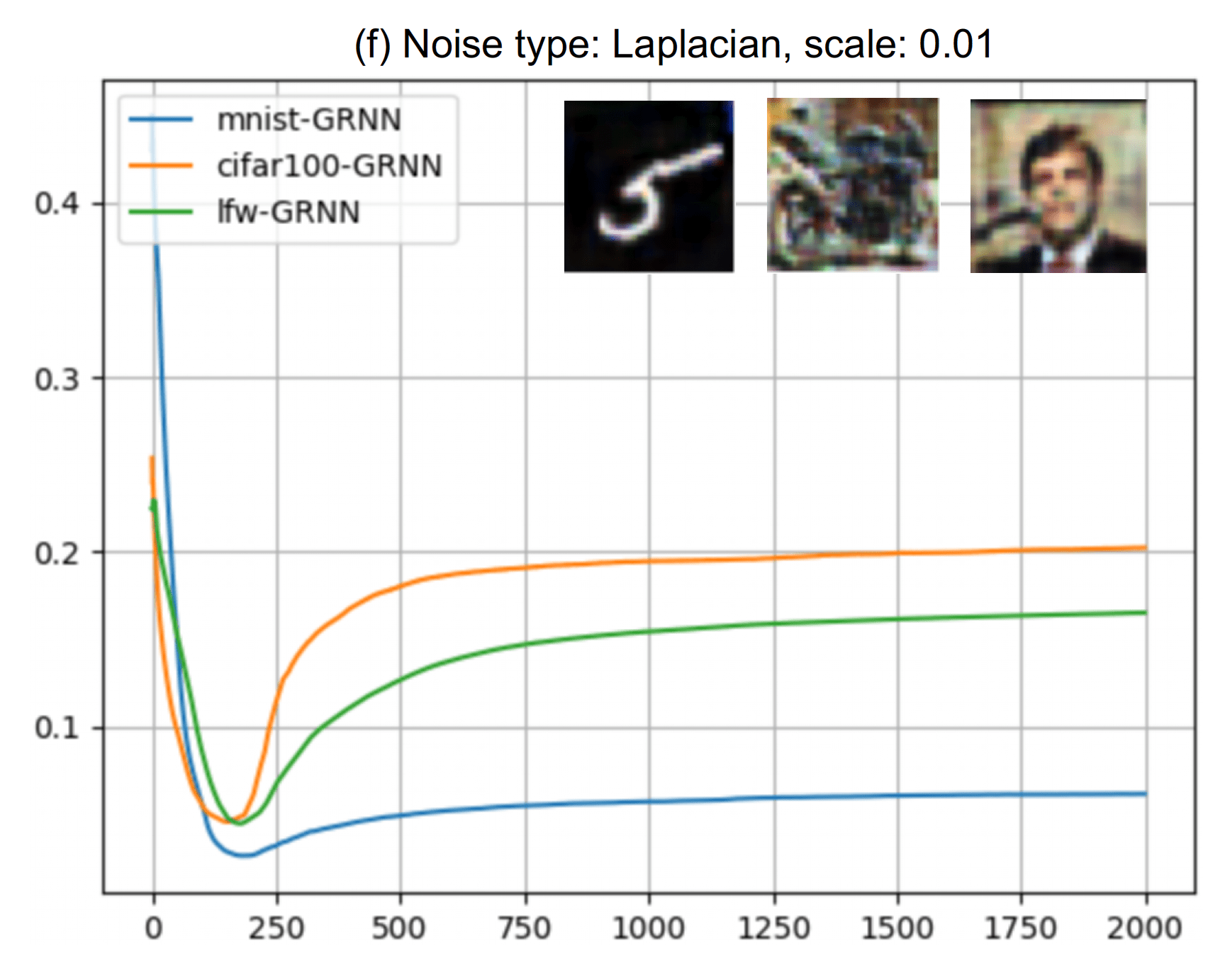}} \\
    \vspace{-.2in}
    \subfigure{\includegraphics[width=0.39\linewidth]{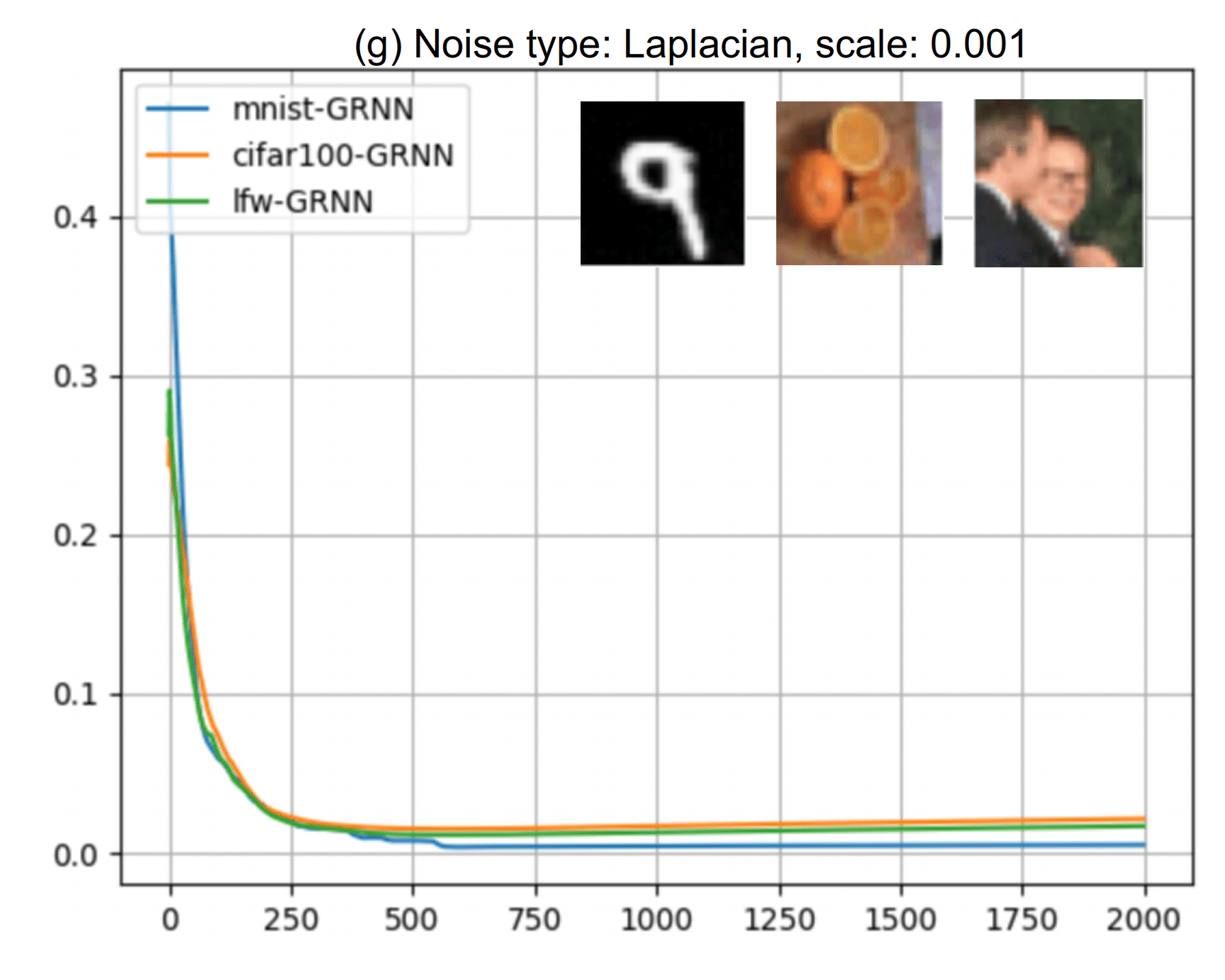}}
    \hspace{.5in}
    \subfigure{\includegraphics[width=0.39\linewidth]{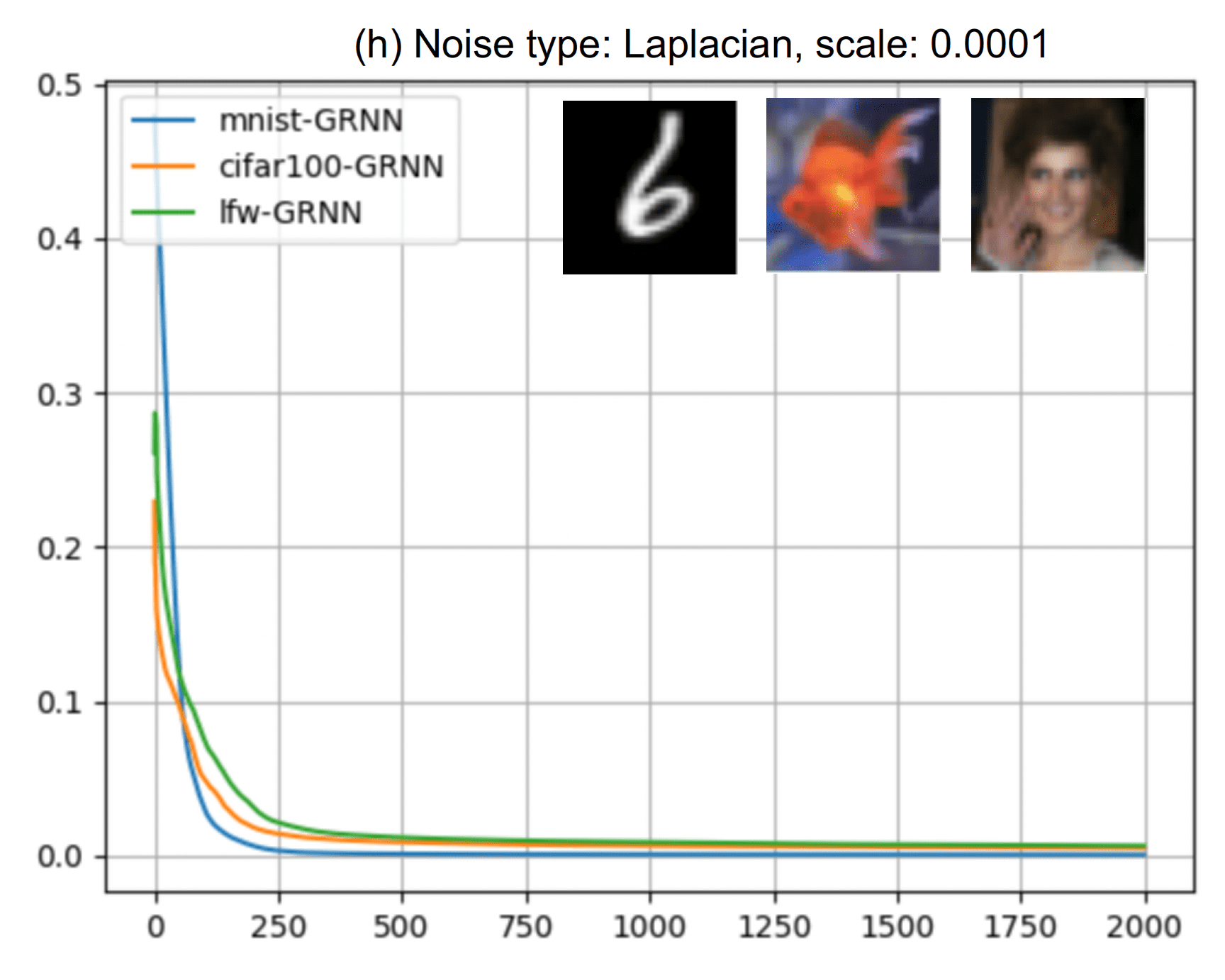}}
    \vspace{-.2in}
\caption{\ac{MSE} distances between true images and generated fake images during training and illustration of generated fake images from \ac{GRNN} with different noise types and scales on three datasets. (a) - (d) are the results of adding Gaussian noise, and (e) - (h) are the results of adding Laplacian noise. The horizontal axis is the number of iteration for training the attack model.}
\label{fig:DPexamples}
\end{figure}

The most relevant defense approach for \ac{GRNN} is noise addition, where, in our scenario, the clients can add a level of Gaussian or Laplacian noise onto the shared gradient. We empirically evaluated the effectiveness of \ac{GRNN} when the noise addition defense strategy was used. In this study, the \emph{LeNet} was used as the global \ac{FL} model with different batch sizes. As for the Laplacian mechanism, it adds Laplacian-distributed noise to function $f$. In this paper, we set the l1-sensitivity $\Delta f$ to be 1 and varied $\epsilon$, which can be defined as: $\lambda=\frac{\Delta f}{\epsilon} \in [1e-1, 1e-4]$. As for the Gaussian mechanism, it also adds randomness with a normal distribution. Technically, the Gaussian mechanism uses l2-sensitivity and parameter $\delta$ is counted on. We simplify the Gaussian mechanism to add the noise whose distribution has 0 mean and only ranges standard deviation from 1e-1 to 1e-4. Fig.~\ref{fig:DPexamples} shows the \ac{MSE} distance between recovered image and true image when different levels and types of noises are added. \ac{GRNN} fails to recover the image when a high level of noise is added to the gradient (see Figs.~\ref{fig:DPexamples} (a) and (e)), however, this usually leads to poor performance on the global \ac{FL} model as the noisy gradients are aggregated. We observed that \ac{GRNN} is able to recover the image data successfully and obtains reasonable results when the scale of noise is reduced to 0.01 (see Figs.~\ref{fig:DPexamples} (b) and (f)). The average \ac{PSNR} scores are presented in Table \ref{tab:DPresults}. Compared to the no-defense approach shown in Table \ref{tab:PSNRresult}, the noise added to the gradient can result in decreasing of \ac{PSNR} of the generated images, which indicates the effectiveness of the noise addition strategy. However, the proposed \ac{GRNN} is still capable of recovering image data when a high level of noise is added to the gradient, \emph{i.e.} 1e-2. The \ac{PSNR} scores with Gaussian noise scale of 1e-4 on MNIST dataset are even larger than that without noise (52.51 dB VS. 52.37 dB), as well on \ac{LFW} dataset (46.26 dB VS. 45.57 dB). On the other hand, we found that even the image recovery fails, the label can still be inferred correctly using our \ac{GRNN}. However, \ac{DLG} totally fails to recover images and inference labels in all of our experiments.

\section{Conclusion}
\label{sec:cc}
In this paper, we proposed a data leakage attack method, namely \ac{GRNN}, for \ac{FL} system which is capable of recovering both data and its corresponding label. Compared to the state-of-the-art methods, the proposed method is much more stable when a large resolution and a batch size are used. It also outperforms state-of-the-art in terms of fidelity of recovered data and accuracy of label inference. Meanwhile, the experimental results on face re-identification task suggest that \ac{GRNN} outperforms \ac{DLG} by a margin in terms of \emph{Top-1}, \emph{Top-3} and \emph{Top-5} accuracies. We also discussed the potential defense strategies and empirically evaluate the performance of \ac{GRNN} when noise addition defense is applied. We conclude that our method can successfully and consistently recover the data in \ac{FL} when a high level of noise is added to the gradient. The implementation of our method is publicly available to ensure its reproducibility. 

\input{manuscript.bbl}

\end{document}

%% file: manuscript.bbl

%% file: manuscript.bbl

\begin{thebibliography}{48}


\ifx \showCODEN    \undefined \def \showCODEN     #1{\unskip}     \fi
\ifx \showDOI      \undefined \def \showDOI       #1{#1}\fi
\ifx \showISBNx    \undefined \def \showISBNx     #1{\unskip}     \fi
\ifx \showISBNxiii \undefined \def \showISBNxiii  #1{\unskip}     \fi
\ifx \showISSN     \undefined \def \showISSN      #1{\unskip}     \fi
\ifx \showLCCN     \undefined \def \showLCCN      #1{\unskip}     \fi
\ifx \shownote     \undefined \def \shownote      #1{#1}          \fi
\ifx \showarticletitle \undefined \def \showarticletitle #1{#1}   \fi
\ifx \showURL      \undefined \def \showURL       {\relax}        \fi
\providecommand\bibfield[2]{#2}
\providecommand\bibinfo[2]{#2}
\providecommand\natexlab[1]{#1}
\providecommand\showeprint[2][]{arXiv:#2}

\bibitem[\protect\citeauthoryear{Aono, Hayashi, Wang, Moriai,
  et~al\mbox{.}}{Aono et~al\mbox{.}}{2017}]%
        {aono2017privacy}
\bibfield{author}{\bibinfo{person}{Yoshinori Aono}, \bibinfo{person}{Takuya
  Hayashi}, \bibinfo{person}{Lihua Wang}, \bibinfo{person}{Shiho Moriai},
  {et~al\mbox{.}}} \bibinfo{year}{2017}\natexlab{}.
\newblock \showarticletitle{Privacy-preserving deep learning via additively
  homomorphic encryption}.
\newblock \bibinfo{journal}{\emph{IEEE Trans. on IFS}} \bibinfo{volume}{13},
  \bibinfo{number}{5} (\bibinfo{year}{2017}), \bibinfo{pages}{1333--1345}.
\newblock


\bibitem[\protect\citeauthoryear{Arjovsky, Chintala, and Bottou}{Arjovsky
  et~al\mbox{.}}{2017}]%
        {arjovsky2017wasserstein}
\bibfield{author}{\bibinfo{person}{Martin Arjovsky}, \bibinfo{person}{Soumith
  Chintala}, {and} \bibinfo{person}{L{\'e}on Bottou}.}
  \bibinfo{year}{2017}\natexlab{}.
\newblock \showarticletitle{Wasserstein generative adversarial networks}. In
  \bibinfo{booktitle}{\emph{PMLR ICML}}. \bibinfo{pages}{214--223}.
\newblock


\bibitem[\protect\citeauthoryear{Armknecht, Boyd, Carr, Gj{\o}steen,
  J{\"a}schke, Reuter, and Strand}{Armknecht et~al\mbox{.}}{2015}]%
        {armknecht2015guide}
\bibfield{author}{\bibinfo{person}{Frederik Armknecht}, \bibinfo{person}{Colin
  Boyd}, \bibinfo{person}{Christopher Carr}, \bibinfo{person}{Kristian
  Gj{\o}steen}, \bibinfo{person}{Angela J{\"a}schke},
  \bibinfo{person}{Christian~A Reuter}, {and} \bibinfo{person}{Martin Strand}.}
  \bibinfo{year}{2015}\natexlab{}.
\newblock \showarticletitle{A Guide to Fully Homomorphic Encryption.}
\newblock \bibinfo{journal}{\emph{IACR Cryptol.}}  \bibinfo{volume}{2015}
  (\bibinfo{year}{2015}), \bibinfo{pages}{1192}.
\newblock


\bibitem[\protect\citeauthoryear{Aslett, Esperan{\c{c}}a, and Holmes}{Aslett
  et~al\mbox{.}}{2015}]%
        {aslett2015encrypted}
\bibfield{author}{\bibinfo{person}{Louis~JM Aslett}, \bibinfo{person}{Pedro~M
  Esperan{\c{c}}a}, {and} \bibinfo{person}{Chris~C Holmes}.}
  \bibinfo{year}{2015}\natexlab{}.
\newblock \showarticletitle{Encrypted statistical machine learning: new privacy
  preserving methods}.
\newblock \bibinfo{journal}{\emph{arXiv preprint arXiv:1508.06845}}
  (\bibinfo{year}{2015}).
\newblock


\bibitem[\protect\citeauthoryear{Bonawitz, Ivanov, Kreuter, Marcedone, McMahan,
  Patel, Ramage, Segal, and Seth}{Bonawitz et~al\mbox{.}}{2016}]%
        {bonawitz2016practical}
\bibfield{author}{\bibinfo{person}{Keith Bonawitz}, \bibinfo{person}{Vladimir
  Ivanov}, \bibinfo{person}{Ben Kreuter}, \bibinfo{person}{Antonio Marcedone},
  \bibinfo{person}{H~Brendan McMahan}, \bibinfo{person}{Sarvar Patel},
  \bibinfo{person}{Daniel Ramage}, \bibinfo{person}{Aaron Segal}, {and}
  \bibinfo{person}{Karn Seth}.} \bibinfo{year}{2016}\natexlab{}.
\newblock \showarticletitle{Practical secure aggregation for federated learning
  on user-held data}.
\newblock \bibinfo{journal}{\emph{arXiv preprint arXiv:1611.04482}}
  (\bibinfo{year}{2016}).
\newblock


\bibitem[\protect\citeauthoryear{Bost, Popa, Tu, and Goldwasser}{Bost
  et~al\mbox{.}}{2015}]%
        {bost2015machine}
\bibfield{author}{\bibinfo{person}{Raphael Bost}, \bibinfo{person}{Raluca~Ada
  Popa}, \bibinfo{person}{Stephen Tu}, {and} \bibinfo{person}{Shafi
  Goldwasser}.} \bibinfo{year}{2015}\natexlab{}.
\newblock \showarticletitle{Machine learning classification over encrypted
  data.}. In \bibinfo{booktitle}{\emph{NDSS}}, Vol.~\bibinfo{volume}{4324}.
  \bibinfo{pages}{4325}.
\newblock


\bibitem[\protect\citeauthoryear{Bu, Dong, Long, and Su}{Bu
  et~al\mbox{.}}{2020}]%
        {bu2020deep}
\bibfield{author}{\bibinfo{person}{Zhiqi Bu}, \bibinfo{person}{Jinshuo Dong},
  \bibinfo{person}{Qi Long}, {and} \bibinfo{person}{Weijie~J Su}.}
  \bibinfo{year}{2020}\natexlab{}.
\newblock \showarticletitle{Deep learning with Gaussian differential privacy}.
\newblock \bibinfo{journal}{\emph{Harvard data science review}}
  \bibinfo{volume}{2020}, \bibinfo{number}{23} (\bibinfo{year}{2020}).
\newblock


\bibitem[\protect\citeauthoryear{Dauphin, Fan, Auli, and Grangier}{Dauphin
  et~al\mbox{.}}{2017}]%
        {dauphin2017language}
\bibfield{author}{\bibinfo{person}{Yann~N Dauphin}, \bibinfo{person}{Angela
  Fan}, \bibinfo{person}{Michael Auli}, {and} \bibinfo{person}{David
  Grangier}.} \bibinfo{year}{2017}\natexlab{}.
\newblock \showarticletitle{Language modeling with gated convolutional
  networks}. In \bibinfo{booktitle}{\emph{PMLR ICML}}.
  \bibinfo{pages}{933--941}.
\newblock


\bibitem[\protect\citeauthoryear{Dwork}{Dwork}{2006}]%
        {dwork2006differential}
\bibfield{author}{\bibinfo{person}{Cynthia Dwork}.}
  \bibinfo{year}{2006}\natexlab{}.
\newblock \showarticletitle{Differential privacy}. In
  \bibinfo{booktitle}{\emph{Springer ICALP}}. \bibinfo{pages}{1--12}.
\newblock


\bibitem[\protect\citeauthoryear{Dwork, Roth, et~al\mbox{.}}{Dwork
  et~al\mbox{.}}{2014}]%
        {dwork2014algorithmic}
\bibfield{author}{\bibinfo{person}{Cynthia Dwork}, \bibinfo{person}{Aaron
  Roth}, {et~al\mbox{.}}} \bibinfo{year}{2014}\natexlab{}.
\newblock \showarticletitle{The algorithmic foundations of differential
  privacy.}
\newblock \bibinfo{journal}{\emph{Foundations and Trends in TCS}}
  \bibinfo{volume}{9}, \bibinfo{number}{3-4} (\bibinfo{year}{2014}),
  \bibinfo{pages}{211--407}.
\newblock


\bibitem[\protect\citeauthoryear{Fredrikson, Jha, and Ristenpart}{Fredrikson
  et~al\mbox{.}}{2015}]%
        {fredrikson2015model}
\bibfield{author}{\bibinfo{person}{Matt Fredrikson}, \bibinfo{person}{Somesh
  Jha}, {and} \bibinfo{person}{Thomas Ristenpart}.}
  \bibinfo{year}{2015}\natexlab{}.
\newblock \showarticletitle{Model inversion attacks that exploit confidence
  information and basic countermeasures}. In \bibinfo{booktitle}{\emph{ACM
  CCS}}. \bibinfo{pages}{1322--1333}.
\newblock


\bibitem[\protect\citeauthoryear{Gasc{\'o}n, Schoppmann, Balle, Raykova,
  Doerner, Zahur, and Evans}{Gasc{\'o}n et~al\mbox{.}}{2016}]%
        {gascon2016secure}
\bibfield{author}{\bibinfo{person}{Adri{\`a} Gasc{\'o}n},
  \bibinfo{person}{Phillipp Schoppmann}, \bibinfo{person}{Borja Balle},
  \bibinfo{person}{Mariana Raykova}, \bibinfo{person}{Jack Doerner},
  \bibinfo{person}{Samee Zahur}, {and} \bibinfo{person}{David Evans}.}
  \bibinfo{year}{2016}\natexlab{}.
\newblock \showarticletitle{Secure Linear Regression on Vertically Partitioned
  Datasets.}
\newblock \bibinfo{journal}{\emph{IACR Cryptol.}}  \bibinfo{volume}{2016}
  (\bibinfo{year}{2016}), \bibinfo{pages}{892}.
\newblock


\bibitem[\protect\citeauthoryear{Geiping, Bauermeister, Dr{\"o}ge, and
  Moeller}{Geiping et~al\mbox{.}}{2020}]%
        {geiping2020inverting}
\bibfield{author}{\bibinfo{person}{Jonas Geiping}, \bibinfo{person}{Hartmut
  Bauermeister}, \bibinfo{person}{Hannah Dr{\"o}ge}, {and}
  \bibinfo{person}{Michael Moeller}.} \bibinfo{year}{2020}\natexlab{}.
\newblock \showarticletitle{Inverting Gradients--How easy is it to break
  privacy in federated learning?}
\newblock \bibinfo{journal}{\emph{arXiv preprint arXiv:2003.14053}}
  (\bibinfo{year}{2020}).
\newblock


\bibitem[\protect\citeauthoryear{Goodfellow, Pouget-Abadie, Mirza, Xu,
  Warde-Farley, Ozair, Courville, and Bengio}{Goodfellow et~al\mbox{.}}{2014}]%
        {goodfellow2014generative}
\bibfield{author}{\bibinfo{person}{Ian Goodfellow}, \bibinfo{person}{Jean
  Pouget-Abadie}, \bibinfo{person}{Mehdi Mirza}, \bibinfo{person}{Bing Xu},
  \bibinfo{person}{David Warde-Farley}, \bibinfo{person}{Sherjil Ozair},
  \bibinfo{person}{Aaron Courville}, {and} \bibinfo{person}{Yoshua Bengio}.}
  \bibinfo{year}{2014}\natexlab{}.
\newblock \showarticletitle{Generative adversarial nets}. In
  \bibinfo{booktitle}{\emph{NIPS}}. \bibinfo{pages}{2672--2680}.
\newblock


\bibitem[\protect\citeauthoryear{Gulrajani, Ahmed, Arjovsky, Dumoulin, and
  Courville}{Gulrajani et~al\mbox{.}}{2017}]%
        {gulrajani2017improved}
\bibfield{author}{\bibinfo{person}{Ishaan Gulrajani}, \bibinfo{person}{Faruk
  Ahmed}, \bibinfo{person}{Mart{\'\i}n Arjovsky}, \bibinfo{person}{Vincent
  Dumoulin}, {and} \bibinfo{person}{Aaron~C Courville}.}
  \bibinfo{year}{2017}\natexlab{}.
\newblock \showarticletitle{Improved Training of Wasserstein GANs}.
\newblock   \bibinfo{volume}{30} (\bibinfo{year}{2017}).
\newblock


\bibitem[\protect\citeauthoryear{Hao, Li, Xu, Liu, and Yang}{Hao
  et~al\mbox{.}}{2019}]%
        {hao2019towards}
\bibfield{author}{\bibinfo{person}{Meng Hao}, \bibinfo{person}{Hongwei Li},
  \bibinfo{person}{Guowen Xu}, \bibinfo{person}{Sen Liu}, {and}
  \bibinfo{person}{Haomiao Yang}.} \bibinfo{year}{2019}\natexlab{}.
\newblock \showarticletitle{Towards efficient and privacy-preserving federated
  deep learning}. In \bibinfo{booktitle}{\emph{IEEE ICC}}.
  \bibinfo{pages}{1--6}.
\newblock


\bibitem[\protect\citeauthoryear{Hardy, Henecka, Ivey-Law, Nock, Patrini,
  Smith, and Thorne}{Hardy et~al\mbox{.}}{2017}]%
        {hardy2017private}
\bibfield{author}{\bibinfo{person}{Stephen Hardy}, \bibinfo{person}{Wilko
  Henecka}, \bibinfo{person}{Hamish Ivey-Law}, \bibinfo{person}{Richard Nock},
  \bibinfo{person}{Giorgio Patrini}, \bibinfo{person}{Guillaume Smith}, {and}
  \bibinfo{person}{Brian Thorne}.} \bibinfo{year}{2017}\natexlab{}.
\newblock \showarticletitle{Private federated learning on vertically
  partitioned data via entity resolution and additively homomorphic
  encryption}.
\newblock \bibinfo{journal}{\emph{arXiv preprint arXiv:1711.10677}}
  (\bibinfo{year}{2017}).
\newblock


\bibitem[\protect\citeauthoryear{He, Zhang, Ren, and Sun}{He
  et~al\mbox{.}}{2016}]%
        {he2016deep}
\bibfield{author}{\bibinfo{person}{Kaiming He}, \bibinfo{person}{Xiangyu
  Zhang}, \bibinfo{person}{Shaoqing Ren}, {and} \bibinfo{person}{Jian Sun}.}
  \bibinfo{year}{2016}\natexlab{}.
\newblock \showarticletitle{Deep residual learning for image recognition}. In
  \bibinfo{booktitle}{\emph{IEEE CVPR}}. \bibinfo{pages}{770--778}.
\newblock


\bibitem[\protect\citeauthoryear{Hesamifard, Takabi, and Ghasemi}{Hesamifard
  et~al\mbox{.}}{2017}]%
        {hesamifard2017cryptodl}
\bibfield{author}{\bibinfo{person}{Ehsan Hesamifard}, \bibinfo{person}{Hassan
  Takabi}, {and} \bibinfo{person}{Mehdi Ghasemi}.}
  \bibinfo{year}{2017}\natexlab{}.
\newblock \showarticletitle{Cryptodl: Deep neural networks over encrypted
  data}.
\newblock \bibinfo{journal}{\emph{arXiv preprint arXiv:1711.05189}}
  (\bibinfo{year}{2017}).
\newblock


\bibitem[\protect\citeauthoryear{Hesamifard, Takabi, and Ghasemi}{Hesamifard
  et~al\mbox{.}}{2019}]%
        {hesamifard2019deep}
\bibfield{author}{\bibinfo{person}{Ehsan Hesamifard}, \bibinfo{person}{Hassan
  Takabi}, {and} \bibinfo{person}{Mehdi Ghasemi}.}
  \bibinfo{year}{2019}\natexlab{}.
\newblock \showarticletitle{Deep neural networks classification over encrypted
  data}. In \bibinfo{booktitle}{\emph{ACM CODASPY}}. \bibinfo{pages}{97--108}.
\newblock


\bibitem[\protect\citeauthoryear{Hitaj, Ateniese, and Perez-Cruz}{Hitaj
  et~al\mbox{.}}{2017}]%
        {hitaj2017deep}
\bibfield{author}{\bibinfo{person}{Briland Hitaj}, \bibinfo{person}{Giuseppe
  Ateniese}, {and} \bibinfo{person}{Fernando Perez-Cruz}.}
  \bibinfo{year}{2017}\natexlab{}.
\newblock \showarticletitle{Deep models under the GAN: information leakage from
  collaborative deep learning}. In \bibinfo{booktitle}{\emph{ACM CCS}}.
  \bibinfo{pages}{603--618}.
\newblock


\bibitem[\protect\citeauthoryear{Hore and Ziou}{Hore and Ziou}{2010}]%
        {hore2010image}
\bibfield{author}{\bibinfo{person}{Alain Hore} {and} \bibinfo{person}{Djemel
  Ziou}.} \bibinfo{year}{2010}\natexlab{}.
\newblock \showarticletitle{Image quality metrics: PSNR vs. SSIM}. In
  \bibinfo{booktitle}{\emph{IEEE ICPR}}. \bibinfo{pages}{2366--2369}.
\newblock


\bibitem[\protect\citeauthoryear{Huang, Mattar, Berg, and Learned-Miller}{Huang
  et~al\mbox{.}}{2008}]%
        {huang2008labeled}
\bibfield{author}{\bibinfo{person}{Gary~B Huang}, \bibinfo{person}{Marwan
  Mattar}, \bibinfo{person}{Tamara Berg}, {and} \bibinfo{person}{Eric
  Learned-Miller}.} \bibinfo{year}{2008}\natexlab{}.
\newblock \showarticletitle{Labeled faces in the wild: A database forstudying
  face recognition in unconstrained environments}. In
  \bibinfo{booktitle}{\emph{Workshop on faces in'Real-Life'Images: detection,
  alignment, and recognition}}.
\newblock


\bibitem[\protect\citeauthoryear{Iandola, Moskewicz, Ashraf, and
  Keutzer}{Iandola et~al\mbox{.}}{2016}]%
        {iandola2016firecaffe}
\bibfield{author}{\bibinfo{person}{Forrest~N Iandola},
  \bibinfo{person}{Matthew~W Moskewicz}, \bibinfo{person}{Khalid Ashraf}, {and}
  \bibinfo{person}{Kurt Keutzer}.} \bibinfo{year}{2016}\natexlab{}.
\newblock \showarticletitle{Firecaffe: near-linear acceleration of deep neural
  network training on compute clusters}. In \bibinfo{booktitle}{\emph{IEEE
  CVPR}}. \bibinfo{pages}{2592--2600}.
\newblock


\bibitem[\protect\citeauthoryear{Krizhevsky, Hinton, et~al\mbox{.}}{Krizhevsky
  et~al\mbox{.}}{2009}]%
        {krizhevsky2009learning}
\bibfield{author}{\bibinfo{person}{Alex Krizhevsky}, \bibinfo{person}{Geoffrey
  Hinton}, {et~al\mbox{.}}} \bibinfo{year}{2009}\natexlab{}.
\newblock \showarticletitle{Learning multiple layers of features from tiny
  images}.
\newblock  (\bibinfo{year}{2009}).
\newblock


\bibitem[\protect\citeauthoryear{Krizhevsky, Sutskever, and Hinton}{Krizhevsky
  et~al\mbox{.}}{2017}]%
        {krizhevsky2017imagenet}
\bibfield{author}{\bibinfo{person}{Alex Krizhevsky}, \bibinfo{person}{Ilya
  Sutskever}, {and} \bibinfo{person}{Geoffrey~E Hinton}.}
  \bibinfo{year}{2017}\natexlab{}.
\newblock \showarticletitle{Imagenet classification with deep convolutional
  neural networks}.
\newblock \bibinfo{journal}{\emph{Commun. ACM}} \bibinfo{volume}{60},
  \bibinfo{number}{6} (\bibinfo{year}{2017}), \bibinfo{pages}{84--90}.
\newblock


\bibitem[\protect\citeauthoryear{Kwabena, Qin, Zhuang, and Qin}{Kwabena
  et~al\mbox{.}}{2019}]%
        {kwabena2019mscryptonet}
\bibfield{author}{\bibinfo{person}{Owusu-Agyemang Kwabena},
  \bibinfo{person}{Zhen Qin}, \bibinfo{person}{Tianming Zhuang}, {and}
  \bibinfo{person}{Zhiguang Qin}.} \bibinfo{year}{2019}\natexlab{}.
\newblock \showarticletitle{MSCryptoNet: Multi-scheme privacy-preserving deep
  learning in cloud computing}.
\newblock \bibinfo{journal}{\emph{IEEE Access}}  \bibinfo{volume}{7}
  (\bibinfo{year}{2019}), \bibinfo{pages}{29344--29354}.
\newblock


\bibitem[\protect\citeauthoryear{LeCun}{LeCun}{1998}]%
        {lecun1998mnist}
\bibfield{author}{\bibinfo{person}{Yann LeCun}.}
  \bibinfo{year}{1998}\natexlab{}.
\newblock \showarticletitle{The MNIST database of handwritten digits}.
\newblock \bibinfo{journal}{\emph{http://yann. lecun. com/exdb/mnist/}}
  (\bibinfo{year}{1998}).
\newblock


\bibitem[\protect\citeauthoryear{LeCun, Bottou, Bengio, and Haffner}{LeCun
  et~al\mbox{.}}{1998}]%
        {lecun1998gradient}
\bibfield{author}{\bibinfo{person}{Yann LeCun}, \bibinfo{person}{L{\'e}on
  Bottou}, \bibinfo{person}{Yoshua Bengio}, {and} \bibinfo{person}{Patrick
  Haffner}.} \bibinfo{year}{1998}\natexlab{}.
\newblock \showarticletitle{Gradient-based learning applied to document
  recognition}.
\newblock \bibinfo{journal}{\emph{Proc. IEEE}} \bibinfo{volume}{86},
  \bibinfo{number}{11} (\bibinfo{year}{1998}), \bibinfo{pages}{2278--2324}.
\newblock


\bibitem[\protect\citeauthoryear{Li, Andersen, Park, Smola, Ahmed, Josifovski,
  Long, Shekita, and Su}{Li et~al\mbox{.}}{2014}]%
        {li2014scaling}
\bibfield{author}{\bibinfo{person}{Mu Li}, \bibinfo{person}{David~G Andersen},
  \bibinfo{person}{Jun~Woo Park}, \bibinfo{person}{Alexander~J Smola},
  \bibinfo{person}{Amr Ahmed}, \bibinfo{person}{Vanja Josifovski},
  \bibinfo{person}{James Long}, \bibinfo{person}{Eugene~J Shekita}, {and}
  \bibinfo{person}{Bor-Yiing Su}.} \bibinfo{year}{2014}\natexlab{}.
\newblock \showarticletitle{Scaling distributed machine learning with the
  parameter server}. In \bibinfo{booktitle}{\emph{OSDI}}.
  \bibinfo{pages}{583--598}.
\newblock


\bibitem[\protect\citeauthoryear{Lin, Han, Mao, Wang, and Dally}{Lin
  et~al\mbox{.}}{2018}]%
        {lin2017deep}
\bibfield{author}{\bibinfo{person}{Yujun Lin}, \bibinfo{person}{Song Han},
  \bibinfo{person}{Huizi Mao}, \bibinfo{person}{Yu Wang}, {and}
  \bibinfo{person}{Bill Dally}.} \bibinfo{year}{2018}\natexlab{}.
\newblock \showarticletitle{Deep Gradient Compression: Reducing the
  Communication Bandwidth for Distributed Training}.
\newblock  (\bibinfo{year}{2018}).
\newblock


\bibitem[\protect\citeauthoryear{McMahan, Moore, Ramage, Hampson, and
  y~Arcas}{McMahan et~al\mbox{.}}{2017}]%
        {mcmahan2017communication}
\bibfield{author}{\bibinfo{person}{Brendan McMahan}, \bibinfo{person}{Eider
  Moore}, \bibinfo{person}{Daniel Ramage}, \bibinfo{person}{Seth Hampson},
  {and} \bibinfo{person}{Blaise~Aguera y Arcas}.}
  \bibinfo{year}{2017}\natexlab{}.
\newblock \showarticletitle{Communication-efficient learning of deep networks
  from decentralized data}. In \bibinfo{booktitle}{\emph{PMLR AISTATS}}.
  \bibinfo{pages}{1273--1282}.
\newblock


\bibitem[\protect\citeauthoryear{Melis, Song, De~Cristofaro, and
  Shmatikov}{Melis et~al\mbox{.}}{2019}]%
        {melis2019exploiting}
\bibfield{author}{\bibinfo{person}{Luca Melis}, \bibinfo{person}{Congzheng
  Song}, \bibinfo{person}{Emiliano De~Cristofaro}, {and}
  \bibinfo{person}{Vitaly Shmatikov}.} \bibinfo{year}{2019}\natexlab{}.
\newblock \showarticletitle{Exploiting unintended feature leakage in
  collaborative learning}. In \bibinfo{booktitle}{\emph{IEEE SP}}.
  \bibinfo{pages}{691--706}.
\newblock


\bibitem[\protect\citeauthoryear{Modas, Moosavi-Dezfooli, and Frossard}{Modas
  et~al\mbox{.}}{2019}]%
        {modas2019sparsefool}
\bibfield{author}{\bibinfo{person}{Apostolos Modas},
  \bibinfo{person}{Seyed-Mohsen Moosavi-Dezfooli}, {and}
  \bibinfo{person}{Pascal Frossard}.} \bibinfo{year}{2019}\natexlab{}.
\newblock \showarticletitle{Sparsefool: a few pixels make a big difference}. In
  \bibinfo{booktitle}{\emph{IEEE CVPR}}. \bibinfo{pages}{9087--9096}.
\newblock


\bibitem[\protect\citeauthoryear{Mohassel and Zhang}{Mohassel and
  Zhang}{2017}]%
        {mohassel2017secureml}
\bibfield{author}{\bibinfo{person}{Payman Mohassel} {and}
  \bibinfo{person}{Yupeng Zhang}.} \bibinfo{year}{2017}\natexlab{}.
\newblock \showarticletitle{Secureml: A system for scalable privacy-preserving
  machine learning}. In \bibinfo{booktitle}{\emph{IEEE SP}}.
  \bibinfo{pages}{19--38}.
\newblock


\bibitem[\protect\citeauthoryear{Moosavi-Dezfooli, Fawzi, and
  Frossard}{Moosavi-Dezfooli et~al\mbox{.}}{2016}]%
        {moosavi2016deepfool}
\bibfield{author}{\bibinfo{person}{Seyed-Mohsen Moosavi-Dezfooli},
  \bibinfo{person}{Alhussein Fawzi}, {and} \bibinfo{person}{Pascal Frossard}.}
  \bibinfo{year}{2016}\natexlab{}.
\newblock \showarticletitle{Deepfool: a simple and accurate method to fool deep
  neural networks}. In \bibinfo{booktitle}{\emph{IEEE CVPR}}.
  \bibinfo{pages}{2574--2582}.
\newblock


\bibitem[\protect\citeauthoryear{Moritz, Nishihara, Stoica, and Jordan}{Moritz
  et~al\mbox{.}}{2015}]%
        {moritz2015sparknet}
\bibfield{author}{\bibinfo{person}{Philipp Moritz}, \bibinfo{person}{Robert
  Nishihara}, \bibinfo{person}{Ion Stoica}, {and} \bibinfo{person}{Michael~I
  Jordan}.} \bibinfo{year}{2015}\natexlab{}.
\newblock \showarticletitle{Sparknet: Training deep networks in spark}.
\newblock \bibinfo{journal}{\emph{arXiv preprint arXiv:1511.06051}}
  (\bibinfo{year}{2015}).
\newblock


\bibitem[\protect\citeauthoryear{Parkhi, Vedaldi, and Zisserman}{Parkhi
  et~al\mbox{.}}{2015}]%
        {parkhi2015deep}
\bibfield{author}{\bibinfo{person}{Omkar~M Parkhi}, \bibinfo{person}{Andrea
  Vedaldi}, {and} \bibinfo{person}{Andrew Zisserman}.}
  \bibinfo{year}{2015}\natexlab{}.
\newblock \showarticletitle{Deep face recognition}.
\newblock  (\bibinfo{year}{2015}).
\newblock


\bibitem[\protect\citeauthoryear{Paszke, Gross, Massa, Lerer, Bradbury, Chanan,
  Killeen, Lin, Gimelshein, Antiga, et~al\mbox{.}}{Paszke
  et~al\mbox{.}}{2019}]%
        {paszke2019pytorch}
\bibfield{author}{\bibinfo{person}{Adam Paszke}, \bibinfo{person}{Sam Gross},
  \bibinfo{person}{Francisco Massa}, \bibinfo{person}{Adam Lerer},
  \bibinfo{person}{James Bradbury}, \bibinfo{person}{Gregory Chanan},
  \bibinfo{person}{Trevor Killeen}, \bibinfo{person}{Zeming Lin},
  \bibinfo{person}{Natalia Gimelshein}, \bibinfo{person}{Luca Antiga},
  {et~al\mbox{.}}} \bibinfo{year}{2019}\natexlab{}.
\newblock \showarticletitle{Pytorch: An imperative style, high-performance deep
  learning library}.
\newblock \bibinfo{journal}{\emph{NIPS}}  \bibinfo{volume}{32}
  (\bibinfo{year}{2019}), \bibinfo{pages}{8026--8037}.
\newblock


\bibitem[\protect\citeauthoryear{Rudin, Osher, and Fatemi}{Rudin
  et~al\mbox{.}}{1992}]%
        {rudin1992nonlinear}
\bibfield{author}{\bibinfo{person}{Leonid~I Rudin}, \bibinfo{person}{Stanley
  Osher}, {and} \bibinfo{person}{Emad Fatemi}.}
  \bibinfo{year}{1992}\natexlab{}.
\newblock \showarticletitle{Nonlinear total variation based noise removal
  algorithms}.
\newblock \bibinfo{journal}{\emph{Physica D: nonlinear phenomena}}
  \bibinfo{volume}{60}, \bibinfo{number}{1-4} (\bibinfo{year}{1992}),
  \bibinfo{pages}{259--268}.
\newblock


\bibitem[\protect\citeauthoryear{Shamsabadi, Sanchez-Matilla, and
  Cavallaro}{Shamsabadi et~al\mbox{.}}{2020}]%
        {shamsabadi2020colorfool}
\bibfield{author}{\bibinfo{person}{Ali~Shahin Shamsabadi},
  \bibinfo{person}{Ricardo Sanchez-Matilla}, {and} \bibinfo{person}{Andrea
  Cavallaro}.} \bibinfo{year}{2020}\natexlab{}.
\newblock \showarticletitle{Colorfool: Semantic adversarial colorization}. In
  \bibinfo{booktitle}{\emph{IEEE CVPR}}. \bibinfo{pages}{1151--1160}.
\newblock


\bibitem[\protect\citeauthoryear{Shokri and Shmatikov}{Shokri and
  Shmatikov}{2015}]%
        {shokri2015privacy}
\bibfield{author}{\bibinfo{person}{Reza Shokri} {and} \bibinfo{person}{Vitaly
  Shmatikov}.} \bibinfo{year}{2015}\natexlab{}.
\newblock \showarticletitle{Privacy-preserving deep learning}. In
  \bibinfo{booktitle}{\emph{ACM CCS}}. \bibinfo{pages}{1310--1321}.
\newblock


\bibitem[\protect\citeauthoryear{Shokri, Stronati, Song, and Shmatikov}{Shokri
  et~al\mbox{.}}{2017}]%
        {shokri2017membership}
\bibfield{author}{\bibinfo{person}{Reza Shokri}, \bibinfo{person}{Marco
  Stronati}, \bibinfo{person}{Congzheng Song}, {and} \bibinfo{person}{Vitaly
  Shmatikov}.} \bibinfo{year}{2017}\natexlab{}.
\newblock \showarticletitle{Membership inference attacks against machine
  learning models}. In \bibinfo{booktitle}{\emph{IEEE SP}}.
  \bibinfo{pages}{3--18}.
\newblock


\bibitem[\protect\citeauthoryear{Xing, Ho, Dai, Kim, Wei, Lee, Zheng, Xie,
  Kumar, and Yu}{Xing et~al\mbox{.}}{2015}]%
        {xing2015petuum}
\bibfield{author}{\bibinfo{person}{Eric~P Xing}, \bibinfo{person}{Qirong Ho},
  \bibinfo{person}{Wei Dai}, \bibinfo{person}{Jin~Kyu Kim},
  \bibinfo{person}{Jinliang Wei}, \bibinfo{person}{Seunghak Lee},
  \bibinfo{person}{Xun Zheng}, \bibinfo{person}{Pengtao Xie},
  \bibinfo{person}{Abhimanu Kumar}, {and} \bibinfo{person}{Yaoliang Yu}.}
  \bibinfo{year}{2015}\natexlab{}.
\newblock \showarticletitle{Petuum: A new platform for distributed machine
  learning on big data}.
\newblock \bibinfo{journal}{\emph{IEEE Trans. on Big Data}}
  \bibinfo{volume}{1}, \bibinfo{number}{2} (\bibinfo{year}{2015}),
  \bibinfo{pages}{49--67}.
\newblock


\bibitem[\protect\citeauthoryear{Zeiler, Krishnan, Taylor, and Fergus}{Zeiler
  et~al\mbox{.}}{2010}]%
        {zeiler2010deconvolutional}
\bibfield{author}{\bibinfo{person}{Matthew~D Zeiler}, \bibinfo{person}{Dilip
  Krishnan}, \bibinfo{person}{Graham~W Taylor}, {and} \bibinfo{person}{Rob
  Fergus}.} \bibinfo{year}{2010}\natexlab{}.
\newblock \showarticletitle{Deconvolutional networks}. In
  \bibinfo{booktitle}{\emph{IEEE CVPR}}. \bibinfo{pages}{2528--2535}.
\newblock


\bibitem[\protect\citeauthoryear{Zeiler, Taylor, and Fergus}{Zeiler
  et~al\mbox{.}}{2011}]%
        {zeiler2011adaptive}
\bibfield{author}{\bibinfo{person}{Matthew~D Zeiler}, \bibinfo{person}{Graham~W
  Taylor}, {and} \bibinfo{person}{Rob Fergus}.}
  \bibinfo{year}{2011}\natexlab{}.
\newblock \showarticletitle{Adaptive deconvolutional networks for mid and high
  level feature learning}. In \bibinfo{booktitle}{\emph{IEEE ICCV}}.
  \bibinfo{pages}{2018--2025}.
\newblock


\bibitem[\protect\citeauthoryear{Zhao, Mopuri, and Bilen}{Zhao
  et~al\mbox{.}}{2020}]%
        {zhao2020idlg}
\bibfield{author}{\bibinfo{person}{Bo Zhao}, \bibinfo{person}{Konda~Reddy
  Mopuri}, {and} \bibinfo{person}{Hakan Bilen}.}
  \bibinfo{year}{2020}\natexlab{}.
\newblock \showarticletitle{iDLG: Improved Deep Leakage from Gradients}.
\newblock \bibinfo{journal}{\emph{arXiv preprint arXiv:2001.02610}}
  (\bibinfo{year}{2020}).
\newblock


\bibitem[\protect\citeauthoryear{Zhu, Liu, and Han}{Zhu et~al\mbox{.}}{2019}]%
        {zhu2019deep}
\bibfield{author}{\bibinfo{person}{Ligeng Zhu}, \bibinfo{person}{Zhijian Liu},
  {and} \bibinfo{person}{Song Han}.} \bibinfo{year}{2019}\natexlab{}.
\newblock \showarticletitle{Deep leakage from gradients}. In
  \bibinfo{booktitle}{\emph{NIPS}}. \bibinfo{pages}{14774--14784}.
\newblock


\end{thebibliography}
